\pdfoutput=1

\documentclass{article}

\usepackage{arxiv}
\usepackage[T1]{fontenc}
\usepackage[utf8]{inputenc}
\usepackage{xcolor}
\definecolor{teallink}{RGB}{0,128,172}
\definecolor{orangeheader}{RGB}{12,118,159}
\definecolor{elsevorange}{RGB}{230,80,0}
\definecolor{bannergray}{RGB}{229,229,229}
\definecolor{algoboxgray}{RGB}{238,238,238}
\definecolor{archL1}{RGB}{224,150,52}
\definecolor{archL1bg}{RGB}{252,235,213}
\definecolor{archL2}{RGB}{53,100,148}
\definecolor{archL2bg}{RGB}{222,232,242}
\definecolor{archL3}{RGB}{61,150,132}
\definecolor{archL3bg}{RGB}{221,241,236}
\definecolor{archL4}{RGB}{119,97,163}
\definecolor{archL4bg}{RGB}{232,226,242}
\definecolor{archL5}{RGB}{68,68,68}
\definecolor{archL5bg}{RGB}{231,231,231}
\definecolor{archV}{RGB}{230,126,34}
\usepackage{hyperref}
\hypersetup{
  colorlinks = true,
  linkcolor  = blue,
  citecolor  = blue,
  urlcolor   = blue,
  filecolor  = blue,
  pdfstartview = FitH
}
\usepackage{url}
\usepackage{booktabs}
\usepackage{tabularx}
\usepackage{ragged2e}
\usepackage{multirow}
\usepackage{makecell}
\usepackage{array}
\usepackage{graphicx}
\graphicspath{{images/}{./images/}}
\usepackage{amsmath,amssymb}
\usepackage{enumitem}
\usepackage{microtype}
\usepackage{longtable}
\usepackage{float}
\usepackage{placeins}

\usepackage{needspace}
\usepackage[ruled,vlined]{algorithm2e}
\usepackage{rotating}
\usepackage{tikz}
\usepackage{pgfplots}
\pgfplotsset{compat=1.18}
\usepgfplotslibrary{groupplots}
\usetikzlibrary{shapes.geometric,shapes.misc,arrows.meta,positioning,
  fit,backgrounds,calc,decorations.pathreplacing,matrix,shadows.blur}

\tikzset{icon color/.store in=\iconcolor, icon color=black}
\tikzset{
  pics/icon-request/.style={code={
    \draw[\iconcolor, line width=0.9pt, rounded corners=0.7pt]
      (-0.16,-0.13) rectangle (0.16,0.13);
    \draw[\iconcolor, line width=0.7pt] (-0.09,-0.02) -- (0.09,-0.02);
    \draw[\iconcolor, line width=0.7pt] (-0.09,0.04) -- (0.02,0.04);
  }},
  pics/icon-check/.style={code={
    \draw[\iconcolor, line width=1pt, rounded corners=0.6pt] (-0.15,-0.15) rectangle (0.15,0.15);
    \draw[\iconcolor, line width=1.1pt, line cap=round] (-0.08,0.0) -- (-0.02,-0.08) -- (0.1,0.1);
  }},
  pics/icon-gear/.style={code={
    \foreach \a in {0,45,...,315}{
      \draw[\iconcolor, line width=0.9pt, line cap=round] (\a:0.09) -- (\a:0.155);
    }
    \draw[\iconcolor, line width=0.9pt] (0,0) circle (0.075);
  }},
  pics/icon-camera/.style={code={
    \draw[\iconcolor, line width=0.9pt, rounded corners=0.6pt] (-0.16,-0.1) rectangle (0.16,0.12);
    \draw[\iconcolor, line width=0.9pt] (-0.06,0.12) -- (-0.03,0.17) -- (0.03,0.17) -- (0.06,0.12);
    \draw[\iconcolor, line width=0.9pt] (0,0.01) circle (0.065);
  }},
  pics/icon-chip/.style={code={
    \draw[\iconcolor, line width=0.9pt] (-0.12,-0.12) rectangle (0.12,0.12);
    \draw[\iconcolor, line width=0.9pt] (-0.05,-0.05) rectangle (0.05,0.05);
    \foreach \p in {-0.08,0,0.08}{
      \draw[\iconcolor, line width=0.6pt] (\p,0.12) -- (\p,0.19);
      \draw[\iconcolor, line width=0.6pt] (\p,-0.12) -- (\p,-0.19);
      \draw[\iconcolor, line width=0.6pt] (0.12,\p) -- (0.19,\p);
      \draw[\iconcolor, line width=0.6pt] (-0.12,\p) -- (-0.19,\p);
    }
  }},
  pics/icon-split/.style={code={
    \draw[\iconcolor, line width=0.9pt] (-0.17,0) -- (-0.04,0);
    \draw[\iconcolor, line width=0.9pt] (-0.04,0) -- (0.14,0.12);
    \draw[\iconcolor, line width=0.9pt] (-0.04,0) -- (0.14,-0.12);
    \draw[\iconcolor, fill=\iconcolor] (0.14,0.12) circle (0.028);
    \draw[\iconcolor, fill=\iconcolor] (0.14,-0.12) circle (0.028);
    \draw[\iconcolor, fill=\iconcolor] (-0.17,0) circle (0.028);
  }},
  pics/icon-search/.style={code={
    \draw[\iconcolor, line width=1pt] (-0.03,0.03) circle (0.12);
    \draw[\iconcolor, line width=1.1pt, line cap=round] (0.06,-0.06) -- (0.17,-0.17);
  }},
  pics/icon-bolt/.style={code={
    \fill[\iconcolor] (0.04,0.17) -- (-0.11,-0.02) -- (-0.01,-0.02) -- (-0.05,-0.17)
      -- (0.11,0.03) -- (0.01,0.03) -- cycle;
  }},
  pics/icon-shield/.style={code={
    \draw[\iconcolor, line width=0.9pt, rounded corners=1pt]
      (0,0.18) -- (0.14,0.12) -- (0.14,-0.05) -- (0,-0.18) -- (-0.14,-0.05) -- (-0.14,0.12) -- cycle;
    \draw[\iconcolor, line width=1pt, line cap=round] (-0.06,0.0) -- (-0.01,-0.06) -- (0.08,0.08);
  }},
  pics/icon-doc/.style={code={
    \draw[\iconcolor, line width=0.9pt, rounded corners=0.5pt]
      (-0.11,-0.17) rectangle (0.11,0.17);
    \foreach \y in {-0.07,0.01,0.09}{
      \draw[\iconcolor, line width=0.55pt] (-0.06,\y) -- (0.06,\y);
    }
  }},
  pics/icon-layers/.style={code={
    \foreach \y in {-0.08,0,0.08}{
      \draw[\iconcolor, line width=0.8pt] (-0.15,\y) -- (0,\y-0.045) -- (0.15,\y) -- (0,\y+0.045) -- cycle;
    }
  }},
  pics/icon-db/.style={code={
    \draw[\iconcolor, line width=0.85pt] (-0.13,-0.13) -- (-0.13,0.1);
    \draw[\iconcolor, line width=0.85pt] (0.13,-0.13) -- (0.13,0.1);
    \draw[\iconcolor, line width=0.85pt] (-0.13,-0.13) arc (180:360:0.13 and 0.055);
    \draw[\iconcolor, line width=0.85pt] (-0.13,0.1) arc (180:360:0.13 and 0.055);
    \draw[\iconcolor, line width=0.85pt] (-0.13,0.1) arc (180:0:0.13 and 0.055);
  }},
  pics/icon-lock/.style={code={
    \draw[\iconcolor, line width=0.9pt, rounded corners=0.5pt] (-0.11,-0.15) rectangle (0.11,0.02);
    \draw[\iconcolor, line width=0.9pt] (-0.07,0.02) arc (180:360:0.07) ;
    \fill[\iconcolor] (0,-0.07) circle (0.025);
  }},
  pics/icon-brain/.style={code={
    \draw[\iconcolor, line width=0.85pt, rounded corners=3pt]
      (-0.15,-0.1) .. controls (-0.2,0.08) and (-0.08,0.2) .. (0,0.15)
      .. controls (0.08,0.2) and (0.2,0.08) .. (0.15,-0.1)
      .. controls (0.15,-0.2) and (-0.15,-0.2) .. (-0.15,-0.1) -- cycle;
    \draw[\iconcolor, line width=0.5pt] (0,-0.13) -- (0,0.13);
    \draw[\iconcolor, line width=0.5pt] (-0.08,-0.05) -- (-0.02,-0.02);
    \draw[\iconcolor, line width=0.5pt] (0.08,0.02) -- (0.02,0.05);
  }},
  pics/icon-recover/.style={code={
    \draw[\iconcolor, line width=1pt, -{Stealth[length=3.2pt]}]
      (0.13,0) arc (0:300:0.13);
  }},
  pics/icon-target/.style={code={
    \draw[\iconcolor, line width=0.8pt] (0,0) circle (0.17);
    \draw[\iconcolor, line width=0.8pt] (0,0) circle (0.09);
    \fill[\iconcolor] (0,0) circle (0.03);
  }},
  pics/icon-star/.style={code={
    \fill[\iconcolor] (90:0.19) \foreach \a in {90,162,...,360}{
      -- (\a+36:0.075) -- (\a+72:0.19)
    } -- cycle;
  }},
  pics/icon-tree/.style={code={
    \draw[\iconcolor, line width=0.85pt] (0,-0.17) -- (0,0.02);
    \draw[\iconcolor, line width=0.85pt] (0,-0.05) -- (-0.13,0.06);
    \draw[\iconcolor, line width=0.85pt] (0,-0.05) -- (0.13,0.06);
    \draw[\iconcolor, line width=0.85pt] (0,0.02) circle (0.045);
    \draw[\iconcolor, line width=0.85pt] (-0.13,0.06) circle (0.045);
    \draw[\iconcolor, line width=0.85pt] (0.13,0.06) circle (0.045);
  }},
}
\usepackage{caption}
\usepackage{subcaption}
\usepackage{natbib}

\renewcommand{\arraystretch}{1.05}
\newcommand{\tsec}[1]{#1}
\DeclareRobustCommand{\code}[1]{{\ttfamily\let\origUS\_\renewcommand{\_}{\origUS\allowbreak}#1}}

\title{PhenoIntel: A Lifecycle-Aligned Multi-Agent Web Application for
Verified, Accessible Plant Phenotype Analysis}

\author{
  \textbf{Narendren~S~V} \\
  VIT Bhopal University\\
  Bhopal, Madhya Pradesh, India\\
  \texttt{narendren2006@gmail.com}
  \And
  \textbf{Soumyashree~Kar}$^{*}$ \\
  Centre of Studies in Resources Engineering\\
  Indian Institute of Technology Bombay\\
  Mumbai, Maharashtra, India\\
  \texttt{soumyakar@iitb.ac.in}
}

\renewcommand{\headeright}{A Preprint}
\renewcommand{\undertitle}{A Preprint}
\renewcommand{\shorttitle}{PhenoIntel: A Lifecycle-Aligned Multi-Agent Web Application}

\hypersetup{
pdftitle={PhenoIntel: A Lifecycle-Aligned Multi-Agent Web Application for Verified, Accessible Plant Phenotype Analysis},
pdfauthor={Narendren S V, Soumyashree Kar},
}

\begin{document}
\maketitle

\begin{abstract}
Existing conversational plant-phenotyping platforms are difficult
for plant scientists to use and lack the reliability scientific
research demands: failed analyses are reported as valid measurements
rather than flagged as missing, statistical tests run without
checking assumptions, predictions carry no uncertainty estimate, and
specialised hardware limits accessibility. We present
\textbf{PhenoIntel}, a lifecycle-aligned multi-agent web platform
that turns the full machine-learning workflow into a reliable,
user-friendly phenotyping system. Nine specialised agents divide the
analysis into stages, from image collection through model selection,
inference, and reporting, rather than handing the whole task to one
AI manager. Independent checks separate these stages, and every
agent reads from and writes to one shared, fixed-structure record,
so an inconsistent output from one stage is caught before it reaches
the next. Uncertainty is matched to each model family, conformal
prediction, detection-confidence spread, or Monte Carlo Dropout,
rather than applied uniformly, and quality thresholds adapt to crop
and task instead of one global cutoff. When no suitable model
exists, PhenoIntel can propose, validate, and integrate a new one on
its own. The model repository spans ten trained models across five
crops and four imaging modalities. Classification models reach
Macro~F1 of \textbf{0.78--0.996}; object-detection models reach
\textbf{0.96~mAP@50} with a \textbf{54\% reduction} in counting
error over an unoptimised baseline; and a temporal model
reaches held-out Macro~F1 of \textbf{0.7050}. PhenoIntel runs in a
browser on standard hardware, requiring no GPU, and a
\textbf{1{,}200-test} automated suite confirms complete pipeline
execution. Every result carries calibrated uncertainty, validated
statistics, and FAIR-compliant provenance, a combination existing
conversational phenotyping tools do not offer.
\end{abstract}


\section{Introduction}\label{sec:1}

High-throughput plant phenotyping converts images into quantitative
traits, including leaf shape, disease severity, organ counts, canopy
structure, and growth rate. Reliable phenotyping is a bottleneck in
agriculture before it becomes a computer-vision problem: breeding
programmes use it to decide which lines to advance, diagnostic
systems use it to trigger timely intervention, and yield forecasting
uses it to convert early-season crop condition into supply estimates.
In each case, the capacity to collect imagery has outpaced the
capacity to measure it reliably, because manual annotation remains
the standard source of ground truth and does not scale to the volume
of imagery now routinely collected by drones, phenotyping platforms,
and satellite constellations. A phenotype measurement directly
informs downstream agronomic decisions, so an inaccurate measurement
can lead to an inappropriate fertiliser, irrigation, or breeding
decision, and an overconfident one can be acted on without the
scrutiny it warrants. For this reason, PhenoIntel treats confidence
intervals, plausibility checks, and traceable provenance records as
core requirements rather than optional features (Section~\ref{sec:4}).

Tools developed to address this problem have progressed through three
generations, each resolving a limitation of the previous generation
while introducing a new one. Threshold-based tools such as
PlantCV~\citep{fahlgren2015}, FIJI/ImageJ~\citep{schindelin2012}, and
Canopeo~\citep{patrignani2015} are transparent and computationally
efficient but require manual retuning for each crop and lighting
condition. Learned detectors and segmentation models such as
YOLOv8~\citep{jocher2023} and Mask2Former~\citep{cheng2022} generalise
more effectively, but their performance is constrained by the amount
of labelled data available for each crop. The most recent generation, large vision models pretrained on unlabelled image collections,
such as DINOv2~\citep{oquab2023} and Segment
Anything~\citep{kirillov2023,ravi2024}, transfers to new crops with
little or no retraining. These developments have produced strong
individual components, but no reliable method currently orchestrates
them into a single auditable pipeline: a user must still select the
appropriate model for a given crop and task, verify its output before
subsequent stages use it, and quantify the remaining uncertainty.

A separate line of work addresses this orchestration problem using
language models as managers that interpret a request and determine,
step by step, which tool to invoke next~\citep{yao2023,schick2023}, an
approach subsequently adapted to specific domains such as
chemistry~\citep{bran2023} and genomics~\citep{jin2023}.
PhenoAssistant~\citep{xu2024} is the most direct application of this
approach to phenotyping: a GPT-4o-based manager that routes requests
to one of 25 vision tools, reporting 70\% tool-sequencing accuracy and
98\% model-selection accuracy. Functional operation, however, does not
by itself indicate reliability.

An examination of how such monolithic-orchestrator systems are
constructed identified four recurring failure patterns, presented here
as representative of this class of system rather than as a critique of
a single implementation. (i)~A single inference step performs detection,
segmentation, and measurement simultaneously, with no mechanism to
catch a failure before it is reported, so a missing detection becomes
indistinguishable from a genuine zero. (ii)~Phenotype outputs are
reported as single values without confidence intervals, and
statistical tests are executed without verifying their underlying
assumptions. (iii)~Task and model coverage is hard-coded, so a crop or
trait outside the deployed set has no graceful fallback. (iv)~The
interface typically requires a local GPU, a fixed software
environment, and multiple API credentials, limiting accessibility for
most of the domain scientists it is intended to serve.

These four patterns share a common cause. Planning, tool selection,
inference, validation, statistics, and interpretation are handled by a
single prompt chain with no formal boundary between stages, so a
failure at one stage propagates unchecked to the next. Addressing each
symptom individually does not resolve the underlying structural
issue, because the next unchecked assumption or unsupported task
reproduces the same failure through the same mechanism. A comparable
lesson was established earlier in ML-training practice: tools such as
MLflow~\citep{zaharia2018} and Kubeflow Pipelines~\citep{bisong2019}
formalised a similar separation of concerns across a training
lifecycle, for the same reason that a monolithic training script and a
monolithic orchestrator fail in analogous ways. \textbf{PhenoIntel}
applies this lifecycle discipline to conversational,
computer-vision-orchestrated pipelines at inference time, a structure
that does not currently exist in this domain.

This paper makes two contributions that follow directly from this
diagnosis: a lifecycle-aligned multi-agent architecture in which nine
specialised agents, separated by independent verification checkpoints,
replace the single monolithic orchestrator (Section~\ref{sec:4}); and
an evaluation of that architecture across ten trained models and six
cross-domain case studies, reported alongside a source-code audit of a prior monolithic phenotyping
orchestrator~\citep{xu2024} that motivated each design decision
(Sections~\ref{sec:5}--\ref{sec:results}). Section~\ref{sec:relwork}
situates this work relative to phenotyping software, computer-vision
models, and agentic AI frameworks; Section~\ref{sec:curation} describes
the datasets used; Section~\ref{sec:4} presents the architecture in
full; Sections~\ref{sec:5} and~\ref{sec:results} report the
experimental evaluation; Section~\ref{sec:9} discusses what the
results mean; and Section~\ref{sec:10} states the limitations, the
scope for future work, and the conclusions.

\section{Related Work}\label{sec:relwork}

The literature informing this work was assembled through a structured
review spanning four domains: plant phenotyping software and
deep-learning tools; uncertainty-quantification-based ML lifecycle
engineering and pipeline validation; LLM orchestration and
conversational scientific agents; and lifecycle-aligned multi-agent AI
frameworks. The comparison
against existing conversational phenotyping platforms is grounded in
direct inspection of their public implementations, agent
orchestration logic, phenotype computation, statistical testing, and
model-loading code, not design intent alone, the same standard
PhenoIntel's own claims are checked against (Section~\ref{sec:8}).

\subsection{Plant Phenotyping Software and Deep-Learning Tools}\label{sec:relwork1}

Established non-agentic phenotyping software, PlantCV, FIJI/ImageJ,
Canopeo, RootNav~\citep{pound2013}, relies mainly on colour
thresholding, morphological filtering, and scripted image-analysis
pipelines, while temporal assessment of phenotypes depends mostly on
statistical modelling applied after the
fact~\citep{kar2020spatemhtp,kar2020automated}. These tools are accurate
within a specific, manually configured context: once a threshold or
filter chain is tuned for a given crop, imaging device, and plant
condition, it produces consistent, interpretable output. But the
pipelines rarely generalise across crop, modality, or task without
expert retuning, and that retuning itself needs programming expertise
that the intended end users, field agronomists and plant breeders, typically have to outsource. This is the accessibility and
generalisation gap.

Deep-learning-based pipelines apply general-purpose computer-vision
architectures to crop and field imagery without addressing how those
models get orchestrated into a pipeline. Self-supervised vision
transformers such as DINOv2 are used as frozen feature extractors for
agricultural classification; real-time detectors such as YOLOv8 are
fine-tuned for structural detection tasks such as spike and panicle
counting; and promptable segmentation models, Mask2Former and Segment
Anything, are used both fine-tuned and zero-shot for leaf and canopy
instance segmentation. The underlying vision capability for
phenotyping is mature, but it is considered in isolation, with
orchestration left unaddressed.

\subsection{Uncertainty Quantification and Pipeline Validation}\label{sec:relwork5}

Uncertainty quantification and calibration is a foundational
requirement for reliable data-driven science. It is common in other
domains, particularly forecasting~\citep{kar2024xwavenet,wang2020deeppipe}
and plant-process
profiling~\citep{srivastava2026seagan,giacomini2022framework,confalonieri2016quantifying},
but is rarely integrated into plant-phenomics pipelines. A family of
statistical techniques known as conformal prediction~\citep{vovk2005}
attaches a confidence range to a model's prediction with a
mathematical guarantee: the true answer falls inside the stated range
as often as claimed, under very few assumptions about the data
distribution. Variants exist for classification
(RAPS~\citep{angelopoulos2021}) and continuous measurements
(conformalised quantile regression~\citep{romano2019}), which suits
the varied, unstructured measurements common in plant science. A
different technique, Monte Carlo Dropout~\citep{gal2016}, estimates
uncertainty by running the same image through a model many times with
dropout left active and reading the spread of outputs, though it only
works for model types that keep that randomness active at prediction
time.

Within plant phenotyping, uncertainty quantification is mostly
directed at semantic segmentation with probabilistic modelling, often
Bayesian CNNs or Monte Carlo Dropout applied to crop-weed
segmentation~\citep{celikkan2023semantic}, where uncertainty maps
highlight ambiguous crop-weed boundaries. Other work applies
variational inference with dropout-based approximations to separate
aleatoric from epistemic uncertainty, reducing misclassifications and
supporting threshold-based decisions for
farmers~\citep{hernandez2020uncertainty}. For continuous trait
estimation, Cherif et al.~\citep{cherif2026uncertainty} show the role
of uncertainty in hyperspectral inversion, and
Faisal~et~al.~\citep{faisal2025uncertainty} provide a roadmap for
integrating uncertainty quantification into operational
precision-agriculture pipelines. Turning these individually validated
methods into one scalable ML pipeline remains a challenge;
Section~\ref{sec:441} explains why no single one of them fits every
model type in a phenotyping model zoo, and how PhenoIntel picks the
right one for each model.

\subsection{LLM Orchestration and Conversational Scientific Agents}\label{sec:relwork3}

A broader family of AI tool-use systems spans chemistry, genomics, and
plant phenotyping, built on the same idea: an AI model reasons about a
task and calls external tools to carry it out, revising its plan as
results come back~\citep{yao2023,schick2023}. Field-specific versions
followed the same pattern: ChemCrow~\citep{bran2023} gave an LLM 17
chemistry synthesis and safety tools and showed that restricting it to
a curated toolkit cut down on hallucinated answers, and
GeneGPT~\citep{jin2023} connected an LLM to genomic databases,
improving the factual accuracy of its biology answers. The most
direct application of this approach to phenotyping, introduced
above~\citep{xu2024}, extends it with a GPT-4o-based manager, built
on AutoGen~\citep{wu2023}, that routes a
plain-language request to one of 25 registered vision and statistics
tools, reporting 70\% accuracy at sequencing tools correctly and 98\%
accuracy at picking the right vision model. But whether the outputs
those tools eventually produce are trustworthy is not clearly
established, and that question is one of the foundations this work is
built on. Across all of these systems, one AI model plans, chooses
tools, runs them, checks the results, and interprets them, all at
once; two runs of the same request can therefore take different paths
and are hard to audit or reproduce, a limitation PhenoIntel's
architecture is designed to address.

\subsection{Lifecycle-Aligned Multi-Agent AI Frameworks}\label{sec:relwork4}

Unlike centralised AI pipelines, multi-agent frameworks coordinate
several specialised agents that collaborate, debate, or compete to
solve complex tasks more effectively than a single model. Leading
frameworks such as Microsoft AutoGen~\citep{wu2023},
LangGraph~\citep{chase2024}, and CrewAI~\citep{moura2023} dominate
research and enterprise deployments, offering interoperable plumbing
through protocols such as MCP, A2A, and AG-UI, but it is left to the
system engineer to decide what the individual roles are and where the
intermediate checks sit. PhenoIntel does not adopt MCP, A2A, or AG-UI
for its own coordination: its nine core agents read from and write to
a single, strictly typed session-state object, compiled once into a
fixed LangGraph graph, with every hand-off validated against that
shared schema rather than parsed from text (Sections~\ref{sec:42}--\ref{sec:43}).
Coordination is synchronous and strictly sequential rather than
message-passing: each agent runs as one graph node, and the graph's
directed edges (plain function calls, not network messages) determine
which node runs next, so two agents never write to the shared state at
the same time and there is no concurrent or asynchronous agent-to-agent
call to synchronise.

A smaller set of published systems apply one of these frameworks
specifically to plant health. Chat Demeter~\citep{zhang2026chatdemeter},
built on the Coze platform, splits a leaf-disease diagnosis session
across four role-specific agents, task planning, inference,
evaluation, and visualisation, that exchange JSON-formatted
messages rather than share one state object; it reports over 99\%
classification accuracy on a rice-disease benchmark, but its own
discussion names inter-agent communication reliability as unresolved
future work. PestMA~\citep{shi2025pestma} applies a three-agent
editorial pattern, an Editor that drafts a recommendation, a
Retriever that gathers external evidence, and a Validator that checks
the draft, to pest-management decisions, reporting that adding the
Validator raises decision accuracy from 86.8\% to 92.6\%, direct
evidence that a separately scored checking step measurably improves a
multi-agent system's output. Both systems point to the same
conclusion as prior conversational phenotyping systems, from a
different angle: coordinating several role-specific agents improves
on a single monolithic model. Neither, though, gives every stage of
its pipeline the kind of checkpoint PhenoIntel enforces uniformly.
Chat Demeter's checking is confined to one late-stage Evaluation
Agent, and PestMA's Validator checks only the final recommendation,
not the intermediate steps that fed into it. Section~\ref{sec:4}
describes how PhenoIntel instead places an automatic, non-AI
verification checkpoint at every one of its nine stage boundaries.

Separately, machine-learning engineering practice for model-training
pipelines settled on exactly this kind of staged, checked design well
before the current wave of LLM tools: MLflow and Kubeflow Pipelines
formalise a training workflow as a sequence of separately validated
stages, for the same reason an unchecked script can quietly corrupt
everything downstream. This paper borrows its stage-and-check design
from that tradition, using LangGraph only as the plumbing that wires
the stages together; which stages exist and where each check sits
follows the MLflow/Kubeflow discipline rather than a generic
multi-agent template.

\subsection{Research Gaps and Objectives}\label{sec:16}

Three gaps emerge from this literature: (1)~\textbf{verification-first
orchestration}, existing LLM orchestrators handle planning, tool
selection, inference, and interpretation in a single prompt chain with
no formal checkpoints, so failures propagate unchecked; (2)~
\textbf{architecture-aware uncertainty quantification}, current
methods are applied as blanket solutions rather than matched to how
different model families behave at inference time; and
(3)~\textbf{extensible, auditable coverage}, most pipelines assume a
fixed set of crops, traits, and models and fail with no fallback
outside it.

These motivate five objectives pursued directly in this paper. The
first is a lifecycle-aligned pipeline with an automatic check between
every stage and a shared, fixed-structure record, so a zero from a
detection failure can never be mistaken for genuine absence of data.
The second is an uncertainty method matched to each model's behaviour
rather than one method applied everywhere. The third is adaptive,
per-(crop, task, modality, metric) quality thresholds in place of a
single fixed cutoff. The fourth is an extended model library spanning
biologically different crops and task types. The fifth is
FAIR-compliant packaging of every completed run, verified against an
executable governance checklist rather than asserted in prose.

Based on the research gaps and objectives identified above,
Table~\ref{tab:principles} summarises seven architectural principles
of PhenoIntel and the failure category from prior monolithic
phenotyping orchestrators that each one addresses.

\begin{table}[htbp]
\caption{Seven architectural principles of PhenoIntel and the
failure categories from prior monolithic phenotyping orchestrators each addresses.}\label{tab:principles}
\fontsize{7.5}{9.0}\selectfont
\begin{tabularx}{\linewidth}{@{}l>{\RaggedRight\arraybackslash}X
  >{\centering\arraybackslash}p{1.2cm}@{}}
\toprule
\textbf{Principle} & \textbf{Failure Category Addressed} &
\textbf{Section}\\
\midrule
1.\ Lifecycle-aligned agent decomposition &
  Monolithic orchestrator conflates all stages; non-deterministic
  routing; $\sim$30\% tool-selection error rate &
  Section~\ref{sec:42}\\[2pt]
2.\ Typed state contracts (shared session state) &
  Free-text inter-tool communication; fragile filename hand-off
  (different segmentation backends save under different naming
  conventions); silent breakage on message format
  changes &
  Section~\ref{sec:43}\\[2pt]
3.\ Inter-stage verification with bounded recovery &
  Detection failure recorded as a plain zero; corrupted values
  propagate silently into statistics &
  Section~\ref{sec:44}\\[2pt]
4.\ Per-model uncertainty quantification &
  Single-point phenotype estimates; no confidence intervals; MC
  Dropout misapplied to architectures that disable dropout at
  inference time &
  Section~\ref{sec:441}\\[2pt]
5.\ Biological plausibility validation &
  Implausible values propagate unchecked; exclusive-or mask
  combination underestimates
  overlapping leaf area &
  Section~\ref{sec:45}\\[2pt]
6.\ Sandboxed statistical execution &
  Model-generated code runs unsandboxed on local machine
  with container isolation disabled; no filesystem restriction, no timeout,
  no import allowlist &
  Section~\ref{sec:46}\\[2pt]
7.\ Self-extending task/architecture layer &
  Fixed tool set with no mechanism to add tasks or models without
  editing source; no path for genuinely novel crop/task/modality
  combinations &
  Section~\ref{sec:42c}\\
\bottomrule
\end{tabularx}
\end{table}
\FloatBarrier

\subsection{Contributions of This Work}\label{sec:17}

Based on the gaps and objectives above, this paper makes nine
contributions.

\begin{enumerate}[label=(\arabic*), leftmargin=1.4em, itemsep=2pt, topsep=4pt]
\item A code-level audit of where an AI phenotyping manager fails, examining the tool-selection logic, trait-computation routines, statistics module, and model-loading code of a representative monolithic orchestrator~\citep{xu2024} directly rather than relying on documentation alone.
\item A pipeline built from nine specialised stages instead of one AI manager. All stages read from and write to one shared, strictly typed record, with an automatic check (V0--V7) between every stage, a pattern reusable for other scientific AI pipelines.
\item A confidence-interval method matched to each model type, conformal prediction, detection-confidence spread, or Monte Carlo Dropout, covering classification, detection, regression, and time-series analysis.
\item A web application with separate Predict and Build workflows that needs no local GPU for prediction.
\item A structured model registry queried by crop, task, and imaging type rather than by parsing checkpoint filenames. This mechanism alone addressed roughly 30\% of the routing failures observed in the audited system.
\item A curated model library of 10 task-specific models spanning 5 crops and 4 imaging types, with a zero-shot SAM~+~Grounding-DINO fallback for coverage gaps.
\item A self-extending coverage mechanism that registers unseen crops and tasks, and can propose, self-check, and formally verify a new architecture when none fits.
\item Biologically adaptive dynamic thresholding in place of one fixed accuracy cutoff.
\item A transparent self-audit reporting task-completion rate, calibrated accuracy, statistical validity, and accessibility, including unfavourable outcomes such as weak zero-shot leaf-counting performance (Sections~\ref{sec:71}, \ref{sec:92}).
\end{enumerate}

This paper takes the staged, checked discipline long standard for
training pipelines and applies it to a live, conversational,
AI-orchestrated phenotyping session. It differs from classical
image-analysis tools by staying accessible through plain language
without giving up an audit trail; from the vision models by addressing
how those models are chosen and checked rather than proposing new
ones; and from the uncertainty-quantification literature by matching
the method to the model instead of applying one method generically.
The stage-and-check design and the shared structured record are
general engineering patterns, not specific to plants, but the model
library, the trait-computation and biological-plausibility rules, and
the entire evaluation are built and tested for plant phenotyping
exclusively. This work does not propose new deep-learning
architectures; it adapts existing classification, detection, and
temporal backbones into a more disciplined pipeline for selecting, executing, checking,
and extending them.

\section{Data Description and Curation}\label{sec:curation}

\begin{figure}[htbp]
\centering
\includegraphics[width=\linewidth]{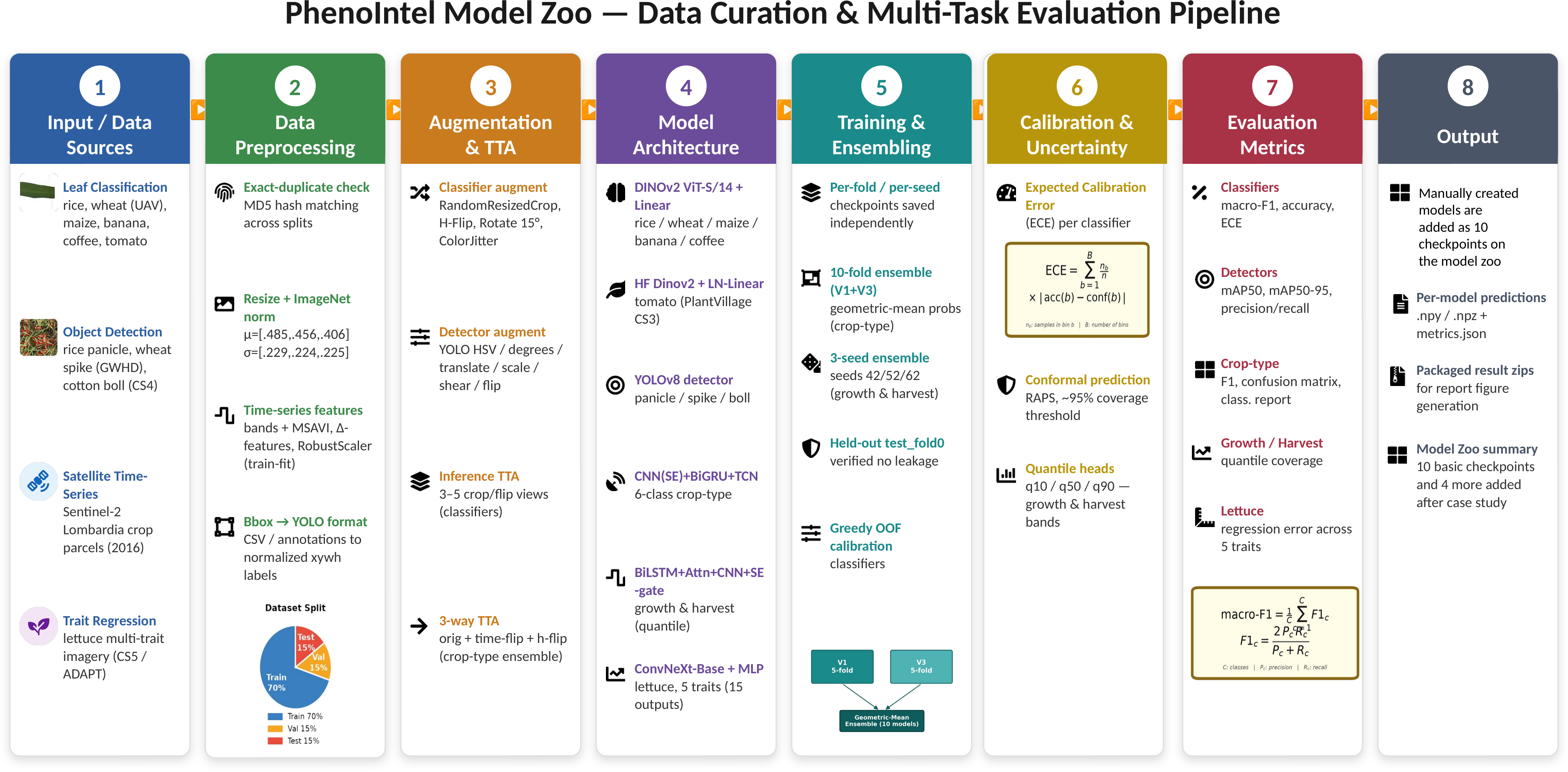}
\caption{The model-zoo data curation and multi-task evaluation
pipeline: eight stages from raw input data sources through
preprocessing, augmentation/TTA, model architecture selection,
training and ensembling, calibration and uncertainty quantification,
evaluation metrics, and packaged output, applied uniformly across the
classification, detection, satellite time-series, and trait-regression
model families reported in Section~\ref{sec:71}.}\label{fig:pipelineflow}
\end{figure}

Figure~\ref{fig:pipelineflow} gives an overview of the data curation
and multi-task evaluation pipeline underlying the model zoo, before
the datasets themselves and the curation steps are described in
detail below.

\subsection{Overview of Datasets}\label{sec:cur0}

The model zoo spans eight (crop, task, modality) combinations: rice
nitrogen-deficiency severity (RGB close-range, $\sim$5,790 images),
wheat deficiency-severity classification (UAV aerial, $\sim$3,000), maize multi-class
deficiency classification (RGB close-range, $\sim$17,000), banana
multi-class deficiency classification (RGB close-range, $\sim$1,500), and coffee
multi-class deficiency classification (RGB close-range, $\sim$800), all
split 70/15/15 stratified; rice panicle detection (UAV aerial,
$\sim$1,800); wheat spike detection (GWHD~\citep{david2021}, field
RGB, $\sim$6,500, official train/test split); and satellite-based crop-type,
growth-rate, and harvest-timing modelling (multi-temporal satellite,
$\sim$50K, 2016--18 train / 2019 test), as tabulated in
Table~\ref{tab:datasetsources}. Public benchmark releases (GWHD,
satellite archives) are used where an established evaluation
convention already exists; field-collected close-range and UAV imagery
are used where no benchmark covers the target crop or canopy scale.

\begin{table}[htbp]
\centering
\caption{Datasets used across the PhenoIntel model zoo. Sizes and
splits describe the complete dataset used for fine-tuning and
held-out evaluation; the smaller, separately curated case-study
input sets are reported in Table~\ref{tab:casestudies}.}\label{tab:datasetsources}
\small
\begin{tabularx}{\linewidth}{@{}l >{\RaggedRight\arraybackslash}X c >{\RaggedRight\arraybackslash}X@{}}
\toprule
\textbf{Crop} & \textbf{Task / Modality} & \textbf{Size} & \textbf{Source}\\
\midrule
Rice & Nitrogen severity / RGB close-range & $\sim$5,790 & Author-processed (Kaggle), \href{https://www.kaggle.com/datasets/narendrensv/pheno-assistant}{kaggle.com/datasets/narendrensv/pheno-assistant}\\
Wheat & Nutrient deficiency / UAV aerial & $\sim$3,000 & \citep{yi2025}, DND-Diko-WWWR, \href{https://github.com/jh-yi/DND-Diko-WWWR}{github.com/jh-yi/DND-Diko-WWWR}\\
Maize & Nutrient deficiency / RGB close-range & $\sim$17,000 & Author-processed (Kaggle), \href{https://www.kaggle.com/datasets/narendrensv/pheno-assistant}{kaggle.com/datasets/narendrensv/pheno-assistant}\\
Banana & Micronutrient deficiency / RGB & $\sim$1,500 & Public (Kaggle), \href{https://www.kaggle.com/datasets/warcoder/nutrient-deficient-banana-plant-leaves}{kaggle.com/datasets/warcoder/nutrient-deficient-banana-plant-leaves}\\
Coffee & Nutrient deficiency / RGB close-range & $\sim$800 & Author-processed \textit{CoLeaf} (Kaggle), \href{https://www.kaggle.com/datasets/narendrensv/coleaf}{kaggle.com/datasets/narendrensv/coleaf}\\
Rice (panicle) & Detection/counting / UAV aerial & $\sim$1,800 & Public (Kaggle), \href{https://www.kaggle.com/datasets/rifat963/annotated-rice-panicle-image-from-bangladesh}{kaggle.com/datasets/rifat963/...bangladesh}\\
Wheat (GWHD) & Spike detection / field RGB & $\sim$6,500 & \citep{david2021}, official split, \href{https://www.global-wheat.com/}{global-wheat.com}\\
Sentinel-2 & Crop type + growth rate + harvest timing / satellite & $\sim$50K & \citep{gallo2023}, 2016--18/2019 split, \href{https://www.kaggle.com/datasets/ignazio/sentinel2-crop-mapping}{kaggle.com/datasets/ignazio/sentinel2-crop-mapping}\\
\bottomrule
\end{tabularx}
\end{table}

\subsection{Curation Methodology}\label{sec:curmethod}

Every dataset passes through the same pipeline before use. Duplicate
removal runs in two passes, an exact-duplicate checksum followed by
a perceptual near-duplicate comparison, since two photographs of the
same leaf taken moments apart are not byte-identical but are still
effectively the same image; this distinction is what the leakage
incident in Section~\ref{sec:82} turns on. Splitting uses a stratified
70/15/15 train/validation/test ratio with a leakage check confirming
no duplicate or near-duplicate pair crosses a split boundary; the
temporal (satellite) pipeline instead uses a strict temporal split (2016--2018
train, 2019 held out), since adjacent satellite passes over the same
field are too correlated for a random split to be trustworthy.

Label quality is checked with a k-nearest-neighbour lookup in the
classification backbone's feature space: an image whose nearest neighbours mostly belong to a
different class is flagged for human review rather than removed
automatically, since it may be a genuine hard example rather than a
mislabelled one. Class imbalance, most notably in the maize dataset
($\sim$3.6$\times$ across $\sim$17,000 images, 6 classes), is handled
with weighted cross-entropy and targeted augmentation. Finally, the
15\% validation allocation is itself split in two, so a slice never
seen during training or early stopping is reserved exclusively for
conformal-prediction calibration (Section~\ref{sec:441}); calibrating
on data the model has already influenced would invalidate the
coverage guarantee. Every split uses a fixed random seed and
stratification key, so it is exactly reproducible. Figure~\ref{fig:dmoverview}
places this curation stage within the full end-to-end pipeline: it is
the prerequisite that feeds every downstream routing, inference, and
verification step detailed from Section~\ref{sec:4} onward.

\FloatBarrier

\section{Methodology}\label{sec:3}\label{sec:4}

\subsection{Overview and Design Thesis}\label{sec:4overview}\label{sec:31}

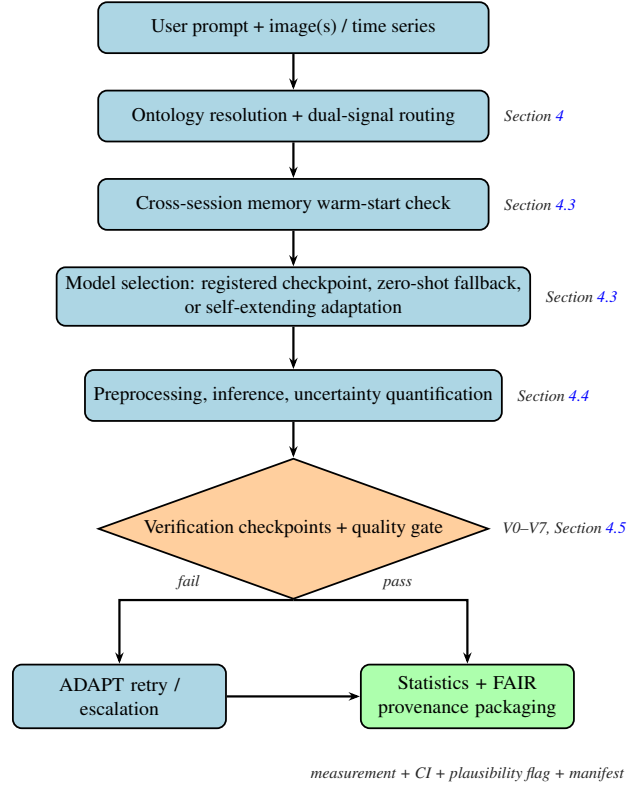
\begin{figure}[htbp]
\centering
\resizebox{0.5\linewidth}{!}{%
\begin{tikzpicture}[
  font=\fontsize{6.5}{7.8}\selectfont,
  stg/.style={draw, line width=0.6pt, fill=teallink!32, rounded corners=3pt,
    minimum width=4.6cm, minimum height=0.6cm, align=center,
    inner sep=3pt},
  gate/.style={draw, line width=0.6pt, fill=orange!38, diamond, aspect=2.6,
    minimum width=4.6cm, minimum height=0.55cm, align=center,
    inner sep=1pt},
  term/.style={draw, line width=0.6pt, fill=green!32, rounded corners=3pt,
    minimum width=4.6cm, minimum height=0.6cm, align=center,
    inner sep=3pt},
  lbl/.style={font=\fontsize{5.4}{6.4}\selectfont\itshape,
    text=black!85},
  arr/.style={-{Stealth[length=3.8pt]}, line width=0.75pt}]

\node[stg] (in) at (0,0) {User prompt + image(s) / time series};
\node[stg, below=0.42cm of in] (route)
  {Ontology resolution + dual-signal routing};
\node[lbl, right=0.05cm of route] {Section~\ref{sec:4}};
\node[stg, below=0.42cm of route] (mem)
  {Cross-session memory warm-start check};
\node[lbl, right=0.05cm of mem] {Section~\ref{sec:42c}};
\node[stg, below=0.42cm of mem] (sel)
  {Model selection: registered checkpoint, zero-shot fallback,\\
   or self-extending adaptation};
\node[lbl, right=0.05cm of sel] {Section~\ref{sec:42c}};
\node[stg, below=0.5cm of sel] (inf)
  {Preprocessing, inference, uncertainty quantification};
\node[lbl, right=0.05cm of inf] {Section~\ref{sec:441}};
\node[gate, below=0.42cm of inf] (ver)
  {Verification checkpoints + quality gate};
\node[lbl, right=0.03cm of ver] {V0--V7, Section~\ref{sec:qualitygate}};
\node[stg, below=0.75cm of ver, xshift=-2.05cm,
  minimum width=2.5cm, text width=2.3cm, minimum height=0.75cm] (adapt)
  {ADAPT retry /\\ escalation};
\node[term, below=0.75cm of ver, xshift=2.05cm,
  minimum width=2.5cm, text width=2.3cm, minimum height=0.75cm] (stat)
  {Statistics + FAIR\\ provenance packaging};
\node[lbl, below=0.32cm of stat] {measurement + CI + plausibility flag + manifest};

\draw[arr] (in) -- (route);
\draw[arr] (route) -- (mem);
\draw[arr] (mem) -- (sel);
\draw[arr] (sel) -- (inf);
\draw[arr] (inf) -- (ver);
\draw[arr] (ver.south) -| node[lbl,pos=0.3,above]{fail} (adapt.north);
\draw[arr] (ver.south) -| node[lbl,pos=0.3,above]{pass} (stat.north);
\draw[arr] (adapt.east) -- (stat.west);
\end{tikzpicture}%
}
\caption{End-to-end methodology at a glance. Curation
(\tsec{Section~\ref{sec:curation}}) and background concepts
(\tsec{Section~\ref{sec:31}}) are prerequisites shown outside this figure;
every box here is described in full architectural detail from
\tsec{Section~\ref{sec:4}} onward.}
\label{fig:dmoverview}
\end{figure}

\begin{sidewaysfigure}[p]
\centering
\includegraphics[width=0.94\textheight]{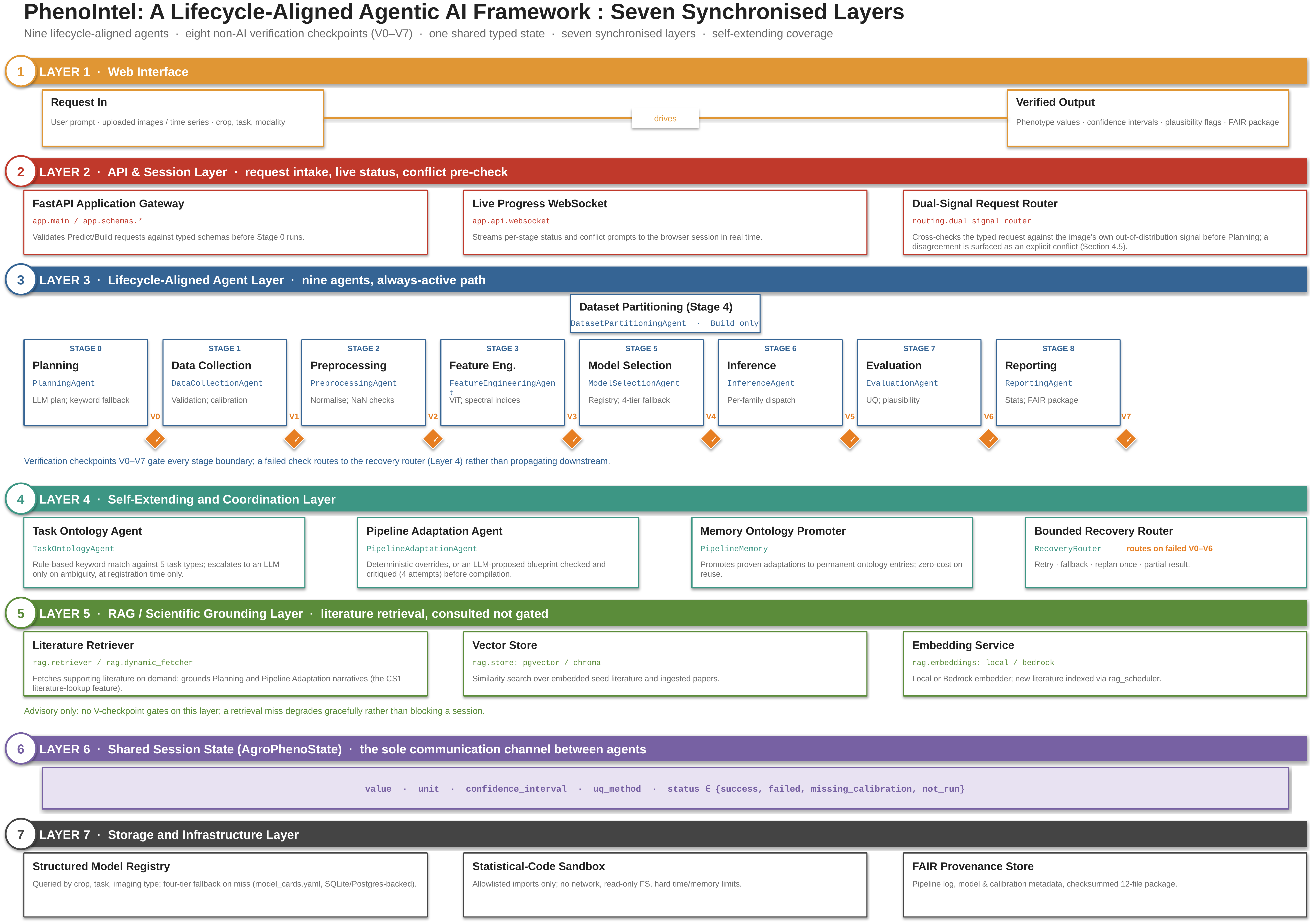}
\caption{PhenoIntel system architecture, organised into seven
synchronised layers. Layer~1 exchanges requests and verified output
with the browser. Layer~2 validates requests, streams progress, and
cross-checks the typed request against the image's own signal
(Section~\ref{sec:dualsignal}). Layer~3 shows eight of the nine
lifecycle agents plus verification checkpoints V0--V7, which act as
\textbf{synchronisation barriers} at every stage boundary
(Table~\ref{tab:agents}); Dataset Partitioning (\textit{Build only})
appears in Figure~\ref{fig:workflows}. Layer~4 holds the
self-extending agents and the bounded recovery router. Layer~5 (RAG)
grounds agent narratives but is consulted, not gated. Layer~6 is the
single, strictly typed \texttt{AgroPhenoState} record, the sole
\textbf{communication channel} between agents (Section~\ref{sec:43}).
Layer~7 holds persistence, sandboxing, and the 12-file FAIR package
(Section~\ref{sec:7}).}\label{fig:arch}
\end{sidewaysfigure}

PhenoIntel accepts a plain-language request together with one or more
uploaded images, or an image time series, and returns a checked trait
measurement, a confidence interval, a plausibility flag, and a full
record of how it was produced. The design follows from a single
diagnosis: the reliability problems seen in AI-orchestrated scientific
workflows come from collapsing steps that any
rigorous machine-learning pipeline treats as separate, independently
checked stages. Standard ML-engineering practice already splits
training work into eight stages, data collection, preprocessing,
feature engineering, splitting, model selection, inference, evaluation,
and deployment, an idea MLflow and Kubeflow formalised for training
pipelines. PhenoIntel transplants this discipline to inference-time,
agentic pipelines: each stage is owned by one dedicated agent with its
own verification checkpoint, summarised end-to-end in
Figure~\ref{fig:dmoverview} before the full architectural detail
follows from Section~\ref{sec:4} onward. Figure~\ref{fig:arch} lays
out the complete seven-layer system architecture this section
describes.

\subsection{The Nine Lifecycle-Aligned Pipeline Agents}\label{sec:42}

A Planning Agent (Stage~0) turns the request into a structured session
plan and is the only one of the nine agents that calls a language
model; every other agent works from that plan's typed fields rather
than parsing text itself. Table~\ref{tab:agents} summarises each
agent's owned responsibility against the specific failure mode it is designed to prevent. Each of the remaining eight stages,
data collection, preprocessing, feature engineering, dataset
partitioning (Build only), model selection, inference, evaluation, and
reporting, is owned by its own agent: Data Collection (Stage~1,
image validation and calibration), Preprocessing (2, normalisation and
multi-modality fusion), Feature Engineering (3, visual and
vegetation-index extraction), Dataset Partitioning (4, Build only,
stratified split and leakage detection), Model Selection (5, registry
look-up with out-of-distribution flagging), Inference (6), Evaluation
(7, uncertainty quantification and plausibility checking), and
Reporting (8, statistical testing and FAIR provenance). This is a
direct response to the monolithic-orchestrator pattern in prior systems, in which
detection, measurement, and reporting are performed within a single inference step
with no boundary at which a failure can be caught before it is
reported.

\begin{table}[htbp]
\caption{The nine lifecycle-aligned pipeline agents and the specific
failure mode from existing monolithic orchestrators each one's responsibility is designed to
prevent.}\label{tab:agents}
\small
\begin{tabularx}{\linewidth}{@{}l >{\RaggedRight\arraybackslash}X >{\RaggedRight\arraybackslash}X@{}}
\toprule
\textbf{Agent (Stage)} & \textbf{Owns Exclusively} & \textbf{Failure Mode Addressed}\\
\midrule
Planning (0) & Structured session plan from an LLM, offline keyword fallback & No formal planning stage; ad-hoc per-turn routing\\
Data Collection (1) & Image validation, capture metadata, pixel calibration & Local-file/URL only; no calibration validation\\
Preprocessing (2) & Normalisation, channel-adapter projection, NaN checks & Corrupted tensors passed silently to inference\\
Feature Engineering (3) & ViT features, spectral indices, temporal assembly & No explicit feature stage; ad hoc prompts\\
Dataset Partitioning (4, Build) & Stratified split, leakage/near-duplicate/mislabel detection & No partitioning validation; leakage undetected\\
Model Selection (5) & Registry look-up, four-tier fallback & Text-based selection, $\sim$30\% error rate\\
Inference (6) & Dispatch across model families; explicit per-detection status & Failed detection recorded as a plain zero\\
Evaluation (7) & UQ, plausibility validation, mask-combination logic & No uncertainty or plausibility checking\\
Reporting (8) & Assumption checks, effect sizes, sandboxed execution, FAIR package & Unconditional ANOVA; no provenance\\
\bottomrule
\end{tabularx}
\end{table}

Model selection illustrates the pattern. Rather than guessing crop and
task from a checkpoint's file name, the Model Selection Agent looks up
a model directly from a structured registry by crop, task, and imaging
type, with a four-level fallback (exact match~$\to$~cross-crop
same-task match~$\to$~degraded-model match~$\to$~zero-shot detection).
A session therefore always ends in either a usable model or an
explicit ``no model available'' message, never a silent wrong
guess. The same routing step also runs a lightweight out-of-distribution check,
comparing an uploaded image against a summary of the selected model's
training images, so a low-confidence-but-familiar input can be told
apart from a low-confidence-and-unfamiliar one downstream. A separate,
three-step term-matching process (exact alias, fuzzy text match, then
semantic-similarity comparison) gives model selection a further
offline fallback whenever a crop or task name does not exactly match
the system's vocabulary; adding a new crop or imaging type only means
extending this list, not writing new code.

The Planning Agent calls a configurable AI model, an open-weight,
70-billion-parameter Llama~3 model via Groq in these runs, swappable
without touching the rest of the system. It returns its plan as a
structured record checked against a fixed template at the first
quality check, so a malformed response is always caught before it can
affect anything downstream. If no AI model is available, the system
falls back to a rule-based keyword matcher instead.

\begin{figure}[htbp]
\centering
\includegraphics[width=0.74\linewidth]{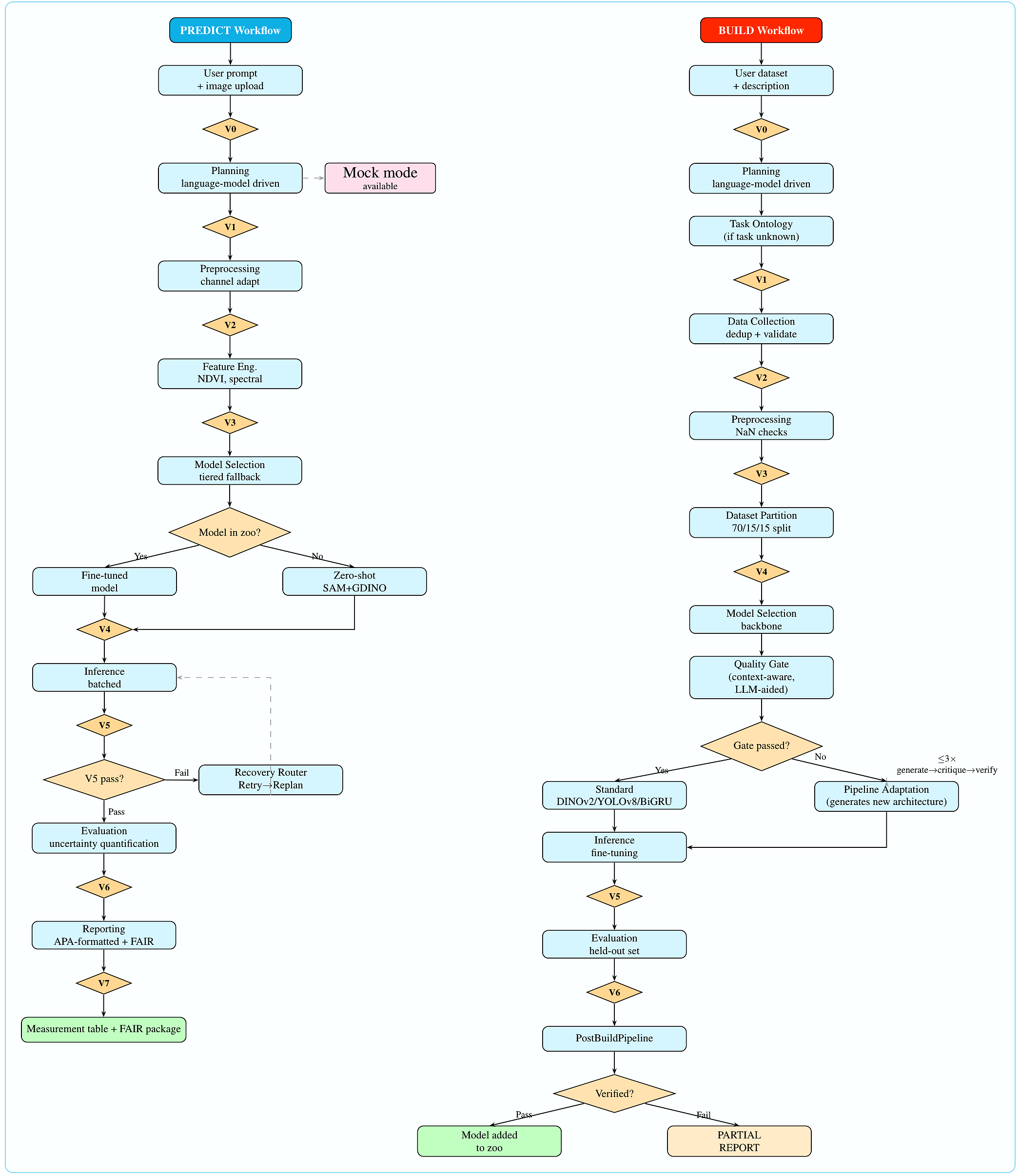}
\caption{PhenoIntel PREDICT (left) and BUILD (right) workflows. Every
stage boundary carries an automatic, non-AI verification checkpoint
(V0--V7, Section~\ref{sec:43}); a failed check routes to the recovery
router rather than propagating downstream. PREDICT resolves Model
Selection against the registered zoo or the zero-shot
SAM~+~Grounding-DINO fallback. BUILD registers unknown tasks and fits
either a standard backbone or, if the quality gate rejects it, a
Pipeline-Adaptation-generated architecture (Section~\ref{sec:42c})
bounded to four generate$\to$critique$\to$verify
attempts.}\label{fig:workflows}
\end{figure}

\subsection{Self-Extending Task and Model Coverage}\label{sec:42c}

PhenoIntel recognises five broad phenotypic task types across the
model zoo and the self-extending layer: \textit{classification}
(assigning a discrete class label, such as a disease or deficiency
category, to an image), \textit{detection/counting} (localising and
enumerating discrete structures, such as spikes, panicles, or bolls),
\textit{regression} (predicting one or more continuous trait values,
such as leaf area or growth rate), \textit{segmentation}
(delineating structural regions such as leaf or canopy boundaries),
and \textit{temporal} (modelling a sequence of images or spectral
bands over time, as in satellite-based crop monitoring). Every model
in the zoo, and every architecture the self-extending layer can
generate, is registered against exactly one of these five task types,
which is what lets the Model Selection Agent (Section~\ref{sec:42})
route a request without parsing free text.

Prior monolithic orchestrators with fixed tool sets have no path for a crop, task, or
imaging type outside their built-in coverage. PhenoIntel closes this gap with
three components that sit outside the fixed nine-agent pipeline, each
built to use its own language-model call as narrowly as the job
allows. When a Build request names an unknown task, a Task Ontology
Agent first tries a rule-based keyword match against the five task
types above; the language model is escalated to only when that match is
ambiguous, at most once per new (crop, task) pair at registration
time, never on the per-image inference path.

When no existing strategy fits a crop/task/modality combination at
all, a Pipeline Adaptation Agent operates in two modes. For a
recognised task pattern it applies a fixed, deterministic set of
parameter overrides with no model call involved. Only when nothing
recognised applies does it fall to a reactive mode: an LLM proposes a
candidate pipeline blueprint, which must pass a schema check, a hard
task-versus-head-type consistency check, and a target-field name check
before an independent language-model pass critiques it against a
numeric threshold; a blueprint that fails any of these is regenerated,
up to four attempts, and if every attempt fails or the model provider
is unreachable, the agent deep-copies the last blueprint this session
already trained rather than leaving the run stalled, explicitly
marking that copy as not critiqued rather than inventing a passing
score for it.

A generated pipeline need not be built from scratch. The agent can
draw its vision model from five sources (Table~\ref{tab:backbonesourcing}),
all compiled and checked the same way: the built-in zoo (the fixed
detection and classification backbones); three public repositories (\code{timm}, Hugging Face Hub,
\code{torchvision}), covering hundreds of pretrained architectures
with a task-appropriate output layer attached; or, if none fit, a
fully custom architecture proposed layer-by-layer by an AI model and
put through the same structural checks and critique pass. Every path
ends in the same schema validation and independent verification
(Section~\ref{sec:44}), so a public-repository model and a
hand-written custom one are held to the same standard.

The fixed model zoo's two backbones were chosen for reasons specific
to each task type rather than by default. The classification task
type uses a self-supervised vision transformer backbone because it was pretrained on unlabelled natural images at
scale and transfers to a new crop or deficiency category with a
comparatively small labelled fine-tuning set, which matters directly
for the low-data classification problems in the zoo (coffee, $N=800$;
banana, $N=1{,}500$); a supervised convolutional backbone trained from
scratch would need substantially more labelled data to reach
comparable accuracy on these crops. The detection/counting task type
uses a real-time single-stage detection model because
structural counting (spikes, panicles, bolls) needs bounding-box
localisation at inference speeds compatible with a browser session on
unfunded, CPU-only hardware (Section~\ref{sec:accessibility}), which
ruled out heavier two-stage detectors. The temporal task type uses a
recurrent-convolutional hybrid temporal model (a bidirectional recurrent network fused with TCN-style dilated
convolutions) because it captures both the short-range and
seasonal-scale patterns in a multi-temporal satellite time series without the
larger training-data requirements of a full temporal transformer. The
zero-shot segmentation pathway (a promptable segmentation model combined with an open-vocabulary detector) was chosen over
a fine-tuned segmentation backbone specifically because no
segmentation task in the zoo currently has enough labelled masks to
fine-tune reliably (Section~\ref{sec:71}).

\begin{table}[htbp]
\caption{Backbone sources available to the Pipeline Adaptation Agent.
All five paths compile through the same schema validation and
verifier before a model is registered.}\label{tab:backbonesourcing}
\fontsize{7.6}{9.1}\selectfont
\begin{tabularx}{\linewidth}{@{}
  >{\RaggedRight\arraybackslash}p{0.19\linewidth}
  >{\RaggedRight\arraybackslash}X
  >{\RaggedRight\arraybackslash}p{0.27\linewidth}@{}}
\toprule
\textbf{Source} & \textbf{What it provides} & \textbf{Example models}\\
\midrule
Built-in zoo & Small, hand-vetted set already used by the fixed
  pipeline; no download or matching step needed. & Fixed detection and classification backbones\\[3pt]
\code{timm} & Pretrained convolutional and vision-transformer
  checkpoints spanning hundreds of architectures, fetched by name. &
  ConvNeXt, EfficientNet, ViT variants\\[3pt]
Hugging Face Hub & Any compatible \code{transformers}/\code{timm}-hosted
  checkpoint, identified by repository name. & Community and
  research vision checkpoints\\[3pt]
\code{torchvision} & Standard reference implementations bundled with
  PyTorch's vision library. & ResNet, MobileNetV3\\[3pt]
Custom generation & A language model proposes a new architecture
  layer-by-layer when no external source fits; checked against the
  same structural rules and critique pass as any other candidate. &
  e.g.\ the CS5 single-stream ViT-S/16 five-head regression network\\
\bottomrule
\end{tabularx}
\end{table}
\FloatBarrier

A similar boundary applies to quality thresholds. Rather than one
fixed pass/fail number applied to every regression or classification
task, an LLM threshold advisor sets a threshold per (crop, task,
modality, metric) combination using domain facts alone, such as a
trait's typical natural range; it is never shown the metric a model
actually achieved, so it cannot be gamed or unconsciously anchor
toward a specific result. Its answer is cached rather than re-queried
on every retry, and the pass/fail comparison stays a plain arithmetic
check against that cached number, with a rejection and fallback to a
generic default if the response is malformed or out of range. A
generated pipeline earns a permanent place in the system's vocabulary
only after it proves itself across several independent sessions.
Across all three components, the language model proposes a design, a
classification, or a number; it is never the thing that decides
whether that proposal is accepted.

\subsection{Uncertainty Quantification and Biological Plausibility}\label{sec:441}\label{sec:45}

Prior systems report every trait as a single number without confidence intervals. PhenoIntel
attaches a confidence interval to every trait using one of three
methods, dispatched by model family rather than forced through one
method: conformal prediction for classification, detection-confidence
spread for detection and zero-shot segmentation, and Monte Carlo
Dropout with quantile regression for the temporal model, summarised in Table~\ref{tab:uqdispatch}. The five
classification models use conformal prediction, specifically
RAPS~\citep{angelopoulos2021}, which produces a set of plausible
classes guaranteed, under mild assumptions, to contain the true class
at a chosen confidence level, calibrated on a held-out slice reserved
during curation (Section~\ref{sec:curmethod}). The detection models and
the zero-shot fallback instead scale count and area uncertainty with
the spread of detection confidence across an image. The
temporal model uses Monte Carlo Dropout, running each input 30 times
with dropout left active and reading the spread of outputs as a 95\%
interval, alongside quantile regression heads that predict the 10th,
50th, and 90th percentiles directly during training.

Every measurement is additionally checked against a plausibility
library of expected ranges per trait, crop, growth stage, and
modality; an out-of-range value is flagged with its severity and
probable cause but never silently blocked, so the value and its
warning both reach the report. A related correction concerns projected
leaf area: prior implementations combine overlapping instance masks with an
exclusive-or operation, which cancels pixels covered by an even number
of overlapping leaves, whereas PhenoIntel takes the union of the
masks, correctly attributing every leaf-covered pixel to the measured
area.

\begin{table}[htbp]
\centering
\caption{Uncertainty-quantification method dispatched per model
family (Section~\ref{sec:441}), the interval it produces, and its
calibration source.}\label{tab:uqdispatch}
\begin{tabularx}{\linewidth}{@{}>{\RaggedRight\arraybackslash}p{3.1cm} >{\RaggedRight\arraybackslash}p{3.3cm} >{\RaggedRight\arraybackslash}X >{\RaggedRight\arraybackslash}p{3.4cm}@{}}
\toprule
\textbf{Model Family} & \textbf{UQ Method} & \textbf{Interval Produced} & \textbf{Calibration Source}\\
\midrule
Classification & Conformal prediction (RAPS)~\citep{angelopoulos2021} & Prediction set guaranteed to contain the true class at a chosen coverage & Held-out calibration slice reserved during curation (Section~\ref{sec:curmethod})\\[4pt]
Detection \& zero-shot segmentation & Detection-confidence spread~\citep{celikkan2023semantic} & Count/area interval width scaled by per-image detection confidence & Detection-confidence distribution across the image (no separate held-out set)\\[4pt]
Temporal modelling & Monte Carlo Dropout ($N{=}30$)~\citep{gal2016} + quantile regression heads~\citep{romano2019} & 95\% interval from dropout spread; P10--P90 from quantile heads directly & Dropout kept active at inference; quantile heads trained with a monotonicity penalty\\
\bottomrule
\end{tabularx}
\end{table}

\subsection{Shared State and Verification Checkpoints}\label{sec:43}\label{sec:44}\label{sec:qualitygate}\label{sec:dualsignal}\label{sec:42b}

In prior monolithic orchestrators, agents hand work to each other through free text, the mechanism behind a filename hand-off bug, where a
change in wording can silently break the chain. PhenoIntel instead
gives every agent a single, strictly typed record to read from and
write to. Every phenotype value, leaf count, spike count, growth
rate, and so on, is stored the same way: a value, a unit, a
confidence interval, the uncertainty method used, and an explicit
status of success, failed, missing calibration, or not yet run. One
rule is enforced on every write: a measurement marked ``not run'' can
never also carry a value of exactly zero, precisely the failure mode
identified in prior systems, where a failed detection is recorded as
a plain zero and treated as real by downstream statistics.

An automatic, rule-based check runs at every stage boundary (V0--V7
across both workflows; Fig.~\ref{fig:workflows}, detailed per-checkpoint
in Table~\ref{tab:checkpoints}), each adding well
under 50~ms and calling no AI model, so a malformed image, a failed
calibration, or a missing detection is caught before it reaches the
next stage rather than after. A recovery router classifies every
failed check and escalates through four bounded responses: lower the
confidence threshold and retry; switch to the general-purpose
zero-shot fallback; replan from the Planning Agent with a revised
prompt, allowed once per session; or report a partial result with a
full explanation of what failed. This bounds every failure to a
predictable, logged outcome instead of an open-ended retry loop or a
silent wrong answer. Before any pipeline stage runs, an incoming
request is also cross-checked against two independent signals rather
than one being trusted outright: what the user's typed request says,
and what the uploaded image's own out-of-distribution check implies
about the crop and task it most resembles. A disagreement between the
two is surfaced as an explicit conflict for the Planning Agent to
resolve, rather than silently favouring one source.

\begin{table}[htbp]
\caption{PhenoIntel verification checkpoints (V0--V7): stage, check
summary, failure action, and whether the checkpoint can mutate
state.}\label{tab:checkpoints}
\fontsize{6.6}{7.6}\selectfont
\setlength{\tabcolsep}{3.2pt}
\begin{tabularx}{\linewidth}{@{}l l >{\RaggedRight\arraybackslash}X
  >{\RaggedRight\arraybackslash}p{2.5cm}
  >{\centering\arraybackslash}p{1.4cm}@{}}
\toprule
\textbf{Checkpoint} & \textbf{Stage Boundary} &
\textbf{Key Checks} & \textbf{Failure Action} &
\textbf{Mutates State}\\
\midrule
\textbf{V0} Pre-flight & Before the data-collection stage &
  Valid (crop, task, modality) triple; $\ge$1 image path; basic
  pixel content &
  Route to error before compute time is consumed & No\\[1pt]
\textbf{V1} Mass Integrity & After the data-collection stage &
  $\ge$1 validated image path; all files exist on disk; calibration
  mode valid; minimum resolution per modality &
  Route to recovery & No\\[1pt]
\textbf{V2} Normalisation & After Preprocessing &
  Preprocessed tensor count = image count; no tensor NaN; pixel
  values within expected normalised range &
  Route to recovery & No\\[1pt]
\textbf{V3} Feature Validity & After Feature Engineering &
  NDVI $\in [-1,\,1]$; temporal array shape $=(40,10,32,32)$ for
  the temporal model; spectral band ranges valid &
  Route to recovery & No\\[1pt]
\textbf{V4} Model Compatibility & After the model-selection stage &
  A model card was actually selected; modality match; checkpoint file exists
  (skipped for zero-shot/generated cards) &
  Route to recovery & No\\[1pt]
\textbf{V5} Post-Inference & After Inference &
  Detection status explicitly set, never left as "not run";
  "no detections" pre-fills as missing (not a literal zero);
  a "success" status with empty output is a hard error &
  Missing-value pre-fill; Route to recovery & \textbf{Yes}\\[1pt]
\textbf{V6} Post-Evaluation & After Evaluation &
  Phenotype results non-empty; every successful measurement has a
  unit, both confidence bounds, and an uncertainty method; every
  measurement has a plausibility status; implausible fraction below
  threshold &
  Route to recovery & No\\[1pt]
\textbf{V7} Statistical Assumptions & After Reporting &
  Every test result with a p-value has assumption check results,
  effect size, APA string; parametric-assumption consistency &
  Warnings only (non-blocking); written to FAIR package & No\\
\bottomrule
\end{tabularx}
\end{table}
\FloatBarrier

\subsection{Sandboxed Reporting and the Web Application}\label{sec:46}\label{sec:47}\label{sec:48}\label{sec:49}

The Reporting Agent generates the statistical analysis for a session
with an AI model, then runs that code inside an isolated container
rather than on the host machine, which is how prior monolithic systems run
AI-generated code. The container has a read-only filesystem
view, no network access, a memory ceiling, and a hard time limit, and
only a small, fixed allowlist of scientific-computing imports is
permitted; results cross back out only through a checked,
fixed-structure record. Before any test is run, an assumption-checking
step applies normality and variance testing and routes automatically
to a non-parametric path when assumptions are violated, reporting
effect sizes alongside every p-value in APA~7th-edition
format~\citep{APA2020}, none of which the prior system's unconditional ANOVA routine performs.

PhenoIntel runs entirely in the browser, requiring no local GPU,
environment configuration, or API credential setup. A Predict and
Analyse page accepts a drag-and-dropped image set and a plain-language
instruction and streams progress live; a Build a Model page fine-tunes
a model on a user-supplied dataset, pre-screened by a Dataset Doctor
check for class imbalance, near-duplicates, and mislabelled files
before training starts. Every completed session, Predict or Build,
produces a downloadable FAIR provenance package (Findable, Accessible,
Interoperable, Reusable), recording the model checkpoint used,
calibration parameters, verification-checkpoint outcomes, and the full
statistical report, so a result can be audited or reproduced without
access to PhenoIntel's own servers or code (Section~\ref{sec:7}).

\FloatBarrier

\section{Experiments and Evaluation}\label{sec:5}

Table~\ref{tab:datasetsources} lists the primary source, size, and
split strategy for every dataset used to fine-tune or evaluate a
model-zoo checkpoint. These are distinct from the smaller curated
subsets used as live inputs to the six cross-domain case studies
(CS1--CS6) below, whose sample sizes are reported in
Table~\ref{tab:casestudies}. Metric abbreviations used throughout this
section (MAE, mAP50, Macro~F1, and others) are defined once, formally,
in Section~\ref{sec:72}, and are not restated here.

\subsection{Model Compilation and Benchmarking}\label{sec:71}

\subsubsection{Zero-shot segmentation}\label{sec:711}

Unlike prior systems that include fine-tuned segmentation checkpoints, PhenoIntel's model zoo
has no fine-tuned segmentation checkpoint. Every structural phenotype
request is instead served by the zero-shot segmentation pathway
(Section~\ref{sec:441}), which needs no per-species tuning
and already supports the XOR$\to$union mask correction and
mask-confidence-variance uncertainty quantification. A fine-tuned
checkpoint remains future work (Section~\ref{sec:93}). We benchmark this pathway on CVPPP~2015/2017
Arabidopsis rosette data, the same family used to evaluate prior fine-tuned segmentation checkpoints, rather than the full official
CVPPP~2017 A1--A3 sweep, deferred because the evaluation machine
(Intel~i3, 8~GB RAM, no GPU) would force downsampling that conflates a
hardware-induced regression with the pathway's own accuracy ceiling.
The benchmarking protocol runs the pathway at its default
automatic-mask-generator thresholds first, then repeats the run after
relaxing the threshold behaviour and switching to a lighter-weight
zero-shot segmentation backbone, isolating the contribution of the
threshold/backbone change from the pathway's own accuracy ceiling; the
union-mask correction fixes the masks it receives but cannot recover
leaves never separated in the first place, so any zero-shot-routed
session is flagged in the UI as unverified. Measured outcomes of both
runs are reported in Section~\ref{sec:62}.

\subsubsection{Classification model fine-tuning}\label{sec:713}

The classification model uses a vision-transformer backbone
with full fine-tuning, adapted per crop and per
dataset (Table~\ref{tab:datasetsources}). Rice (four ordinal classes,
standard cross-entropy): the rice nitrogen set initially showed
inflated validation performance from duplicate images split across
train/val by naive random splitting, corrected with SHA256 plus
perceptual deduplication before any leakage-free figure is reported
(Section~\ref{sec:62}). Wheat, a UAV/field-scale deficiency set
with a larger domain shift between close-range pretraining and aerial
imagery, uses layer-wise learning-rate decay and test-time
augmentation to close that domain gap. Maize ($\sim$17K images,
3.6$\times$ class imbalance, the largest dataset) uses a
class-imbalance-aware weighted cross-entropy scheme. Banana and
coffee, both low-data fine-grained problems, rely on heavier
augmentation to compensate for limited training examples. Measured
Macro~F1 for all five checkpoints is reported in
Section~\ref{sec:62} and Table~\ref{tab:mastersummary}.

\subsubsection{Detection model fine-tuning}\label{sec:7135}

The detection model uses a shared real-time single-stage
architecture, fine-tuned separately per domain rather than jointly, since GWHD
spikes and rice panicles differ substantially in scale and occlusion
pattern. The wheat-spike detector is fine-tuned on the GWHD
UAV/field-scale spike-detection set, $\sim$6,500 field-RGB images
(Table~\ref{tab:datasetsources}), against a self-carved held-out split.
The rice-panicle detector is fine-tuned on a companion UAV/field-scale
set of $\sim$1,800 images with dense occlusion between panicles, after
tiling large UAV frames into overlapping crops at inference time to
recover small, occluded instances that a single full-frame pass
misses, evaluated on its own \texttt{valid/} split. Both checkpoints
are summarised alongside the classification and temporal models in
Table~\ref{tab:checkpoints}, with measured mAP50 reported in
Section~\ref{sec:62} and Table~\ref{tab:mastersummary}.

\subsubsection{Temporal model architecture}\label{sec:714}

The temporal model family uses a recurrent-convolutional hybrid
architecture, currently instantiated on multi-temporal satellite
parcel data, the remote-sensing tier of the datasets underlying the
model zoo (Table~\ref{tab:datasetsources}), split by a strict temporal
rule (2016--2018 train, 2019 test) enforced automatically from
acquisition year. It processes $(40{\times}10{\times}32{\times}32)$
time series over 6 crop classes through
per-timestep spatial feature extraction with channel attention, a
recurrent block fused with dilated-convolution residual blocks for
multi-scale temporal patterns, and multi-task output
heads: a 6-class crop-type softmax, plus growth-rate and
harvest-timing regression via quantile loss across
$(q_{10},q_{50},q_{90})$ with a monotonicity penalty. Monte Carlo
Dropout ($N{=}30$) additionally provides a 95\% interval independent
of the quantile heads; production inference uses a multi-fold
ensemble for classification and multi-seed ensembles for regression.

\subsection{Cross-Domain Case Studies}\label{sec:52cs}

Six case studies validate the architecture across modalities, crops,
and tasks absent from the primary training corpus, executed entirely
through the PhenoIntel web application, no command-line tools,
scripts, or environment configuration required. Table~\ref{tab:casestudies}
summarises each; outcomes are reported in Section~\ref{sec:resultsoverview}.

\begin{table}[htbp]
\caption{Six cross-domain case studies. Group 1 exercises the PREDICT
path (exact match or zero-shot fallback); Group 2 exercises BUILD
(training a new model from an uploaded dataset).}\label{tab:casestudies}
\small
\begin{tabularx}{\linewidth}{@{}l >{\RaggedRight\arraybackslash}X c@{}}
\toprule
\textbf{Case Study} & \textbf{Mechanism Exercised} & \textbf{Min. Dataset}\\
\midrule
CS1 Wheat head detection & Tier-0 exact-match PREDICT; establishes baseline completion/F1 & 10 GWHD images\\
CS2 Arabidopsis leaf count & Tier-2 zero-shot segmentation fallback; manual calibration applied & 5 CVPPP2015 images\\
\midrule
CS3 Tomato disease (BUILD) & Classification backbone, quality gate passed first attempt & 375 images, 5 classes\\
CS4 Cotton boll detection (BUILD) & Detection backbone, quality gate passed first attempt & 60 images\\
CS5 Lettuce multi-trait regression (BUILD) & ADAPT recovery: 3 inference failures $\to$ generate-critique-verify loop & 112 RGB images\\
CS6 Satellite crop type (BUILD) & Degenerate $1{\times}1$-pixel case of the general temporal input format & 800 pixel series, 4 classes\\
\bottomrule
\end{tabularx}
\end{table}

Each case study's design fixes the pipeline path it is meant to
exercise (Table~\ref{tab:casestudies}); the pass/fail criteria and
measured outcome for each are reported together in
Section~\ref{sec:resultsoverview}, after the metric definitions below.

\subsection{Evaluation Metrics}\label{sec:72}

Every metric reported anywhere in Section~\ref{sec:results} is defined
here first, grouped by the task family it applies to, so that
Section~\ref{sec:results} can report measured numbers without
restating definitions.

\subsubsection{General: Reliability, Reproducibility, and Accessibility}\label{sec:721}

\textbf{Task completion rate.} A task counts as completed only if the
pipeline reaches a final phenotype output and statistical report
without a crash, timeout, silent zero-value failure, or unhandled
exception, and only if the output is scientifically valid, not
merely produced:
\begin{equation}
\text{Completion Rate} = \frac{1}{n}\sum_{i=1}^{n} \mathbb{1}\!\left[\text{session}_i \text{ reaches a verified final report}\right].
\label{eq:completion}
\end{equation}
Unhandled-exception / crash rate is tracked as the complementary
failure mode of the same criterion.

\textbf{Reproducibility and provenance.} Evaluated against the FAIR
data principles~\citep{wilkinson2016} (Section~\ref{sec:7}) via the
count of the twelve always-present provenance files and session
resumability; a 1,200-test automated suite (Section~\ref{sec:8})
provides an independent, continuously re-checkable layer of internal
consistency that complements, rather than substitutes for, the
accuracy evaluation below.

\textbf{Statistical output validity.} Whether every session's
statistical report satisfies APA-style requirements~\citep{APA2020}
as a structural property rather than an optional check: a normality
test (Shapiro--Wilk) before any parametric test, automatic fallback to
Kruskal--Wallis~+~Dunn on failure, effect sizes alongside p-values, and
Holm--Bonferroni correction under multiple comparisons.

\textbf{Accessibility.} Time-to-first-result from a cold start, the
number of distinct setup steps required, and whether a local GPU is
needed, measured identically for PhenoIntel and the prior
system~\citep{xu2024} for a like-for-like comparison.

\subsubsection{Classification Metrics}\label{sec:722}

Macro~F1, the unweighted mean of per-class F1, is used over
micro-averaging so majority-class performance cannot dominate:
\begin{equation}
\text{Macro F1} = \frac{1}{C}\sum_{c=1}^{C} \frac{2\,P_c R_c}{P_c + R_c}.
\label{eq:macrof1}
\end{equation}
Per-class precision, recall, and F1 (support $n$ per class) are
reported wherever the underlying classification report was saved.
Calibration is measured with Expected Calibration Error (ECE) across
$B$ equal-width confidence bins,
\begin{equation}
\text{ECE} = \sum_{b=1}^{B} \frac{|\mathcal{B}_b|}{n}\left|\text{acc}(\mathcal{B}_b) - \text{conf}(\mathcal{B}_b)\right|,
\label{eq:ece}
\end{equation}
alongside empirical coverage, the fraction of held-out predictions
whose stated confidence interval or prediction set actually contains
the true label, checked at the same target confidence level ECE is
computed at, and mean prediction-set size for the RAPS conformal
pathway, which measures practical informativeness at a fixed target
coverage.

\subsubsection{Detection Metrics}\label{sec:723}

A predicted box counts as a true positive at a given
intersection-over-union (IoU) threshold if its overlap with a
ground-truth box exceeds that threshold and the box is not already
matched; precision and recall follow directly,
\begin{equation}
\text{Precision} = \frac{TP}{TP+FP}, \qquad \text{Recall} = \frac{TP}{TP+FN}.
\label{eq:precrecall}
\end{equation}
Average Precision (AP) is the area under the precision--recall curve
for a single class, and mean Average Precision (mAP) averages this
over all $C$ classes,
\begin{equation}
\text{AP} = \int_0^1 p(r)\,dr, \qquad \text{mAP} = \frac{1}{C}\sum_{c=1}^{C} \text{AP}_c.
\label{eq:map}
\end{equation}
Mean Average Precision at a 0.50 IoU threshold (mAP50) is the primary
metric, with the COCO-style average of mAP over IoU thresholds
0.50--0.95 in steps of 0.05 (mAP50-95) reported where available.
Counting MAE, defined as in Equation~\ref{eq:maermse2} below, is
additionally reported after confidence-threshold optimisation for
counting specifically. Detection confidence-variance gives an
informal prediction interval where no per-instance ground truth exists
for a formal coverage check.

\subsubsection{Temporal Metrics}\label{sec:724}

Classification heads (crop-type) use Macro~F1 as above. Regression
heads (growth-rate, harvest-timing) are scored with MAE and RMSE on
normalised targets,
\begin{equation}
\text{MAE} = \frac{1}{n}\sum_{i=1}^{n}\left|\hat{y}_i - y_i\right|,
\qquad
\text{RMSE} = \sqrt{\frac{1}{n}\sum_{i=1}^{n}\left(\hat{y}_i - y_i\right)^2},
\label{eq:maermse2}
\end{equation}
alongside Monte Carlo Dropout ($N{=}30$) P10--P90 interval coverage
against an 80\% nominal target, since ECE is undefined for quantile
regression heads.

\subsubsection{Segmentation Metrics}\label{sec:725}

The zero-shot segmentation pathway is scored by leaf-count MAE and
RMSE (Equation~\ref{eq:maermse2}) against ground-truthed CVPPP images,
with detection-confidence-variance giving an informal interval as in
Section~\ref{sec:723}.

\subsubsection{Regression Metrics (Multi-Trait, CS5)}\label{sec:726}

Each of the five lettuce traits (diameter, dry weight, fresh weight,
height, leaf area) is scored independently with MAE and RMSE in its
native unit (Equation~\ref{eq:maermse2}), alongside P10--P90 coverage
against an 80\% target from the quantile output heads, on the same
protocol as the temporal regression heads (Section~\ref{sec:724}).

\section{Results}\label{sec:results}

This section reports the measured outcomes of the case studies and
governance mechanisms described in Sections~\ref{sec:5}--\ref{sec:4}.
Metric definitions are not restated here; they are defined once in
Section~\ref{sec:72}.

\subsection{Case Study Outcomes and Evaluation Framework}\label{sec:resultsoverview}\label{sec:6}

\begin{figure}[htbp]
\centering
\includegraphics[width=0.9\linewidth]{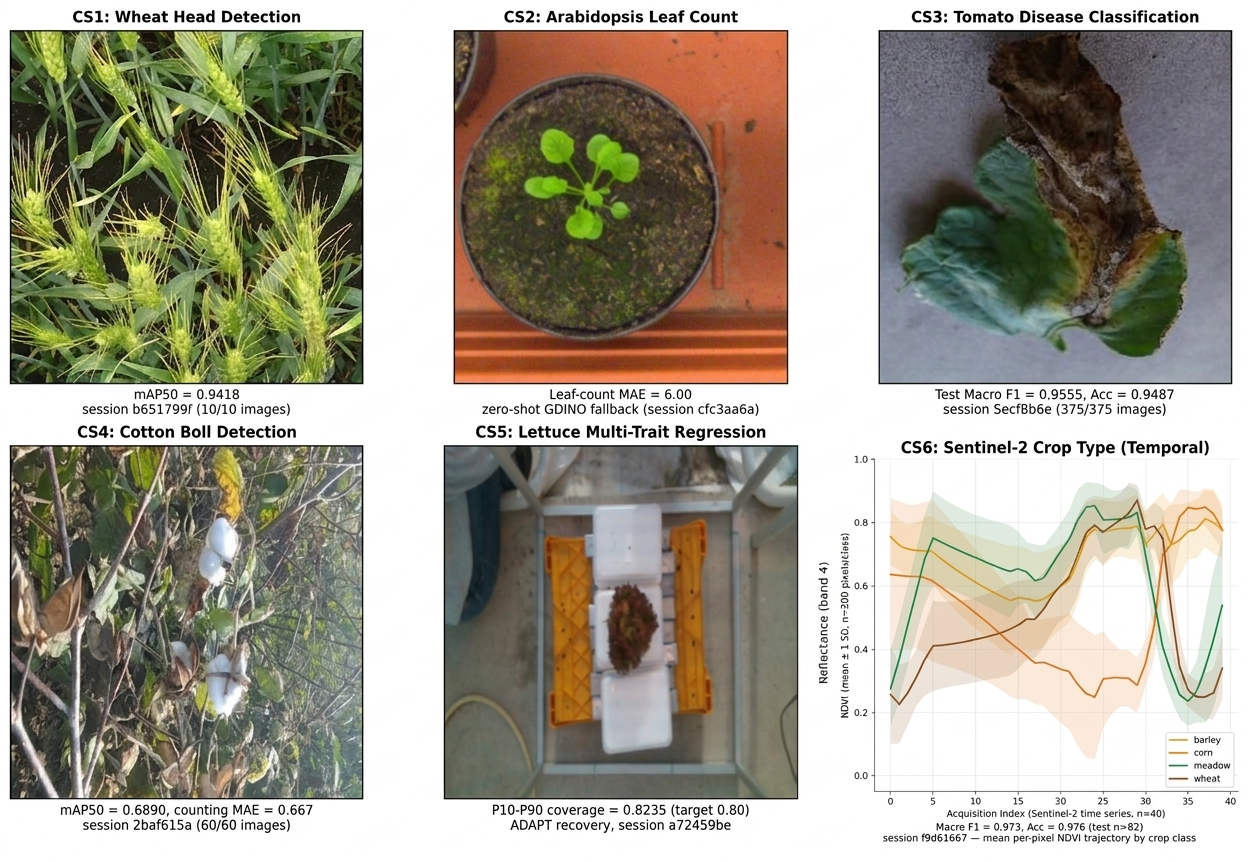}
\caption{Representative input and headline result for each of the six
cross-domain case studies.}\label{fig:casestudyoverview}
\end{figure}

The six cross-domain case studies (Table~\ref{tab:casestudies}) are
the paper's central evidence that PhenoIntel works as an integrated
system, not a set of isolated components. Each is a real,
provenance-logged session run end-to-end through the web application.
Together they cover the full range of pipeline paths: an exact-match
Predict flow (CS1), a zero-shot fallback (CS2), two standard Build
sessions that pass the quality gate on the first try (CS3, CS4), a
live recovery session (CS5), and a Build session registering an
entirely new model family (CS6), summarised visually in
Figure~\ref{fig:casestudyoverview}. A separate group of ten model-zoo checkpoints underlies
the routine Predict pathway and is reported on its own in
Table~\ref{tab:mastersummary}: five classification models, two detection
models, and three temporal models, all trained offline,
independent of the six case studies. CS6 is a distinct exercise: a
live Build session training its own standalone crop-type classifier
from scratch, not one of the ten checkpoints, so its result is
reported with the case studies rather than in Table~\ref{tab:mastersummary}.

Evaluation spans four categories, task completion, factual
accuracy, calibration, and reproducibility (Section~\ref{sec:72}), with the prior system~\citep{xu2024} included as a reference point where a
like-for-like comparison is possible, followed by FAIR governance
(Section~\ref{sec:7}) and internal validation (Section~\ref{sec:8}).

\begin{figure}[htbp]
\centering
\includegraphics[width=0.4125\linewidth]{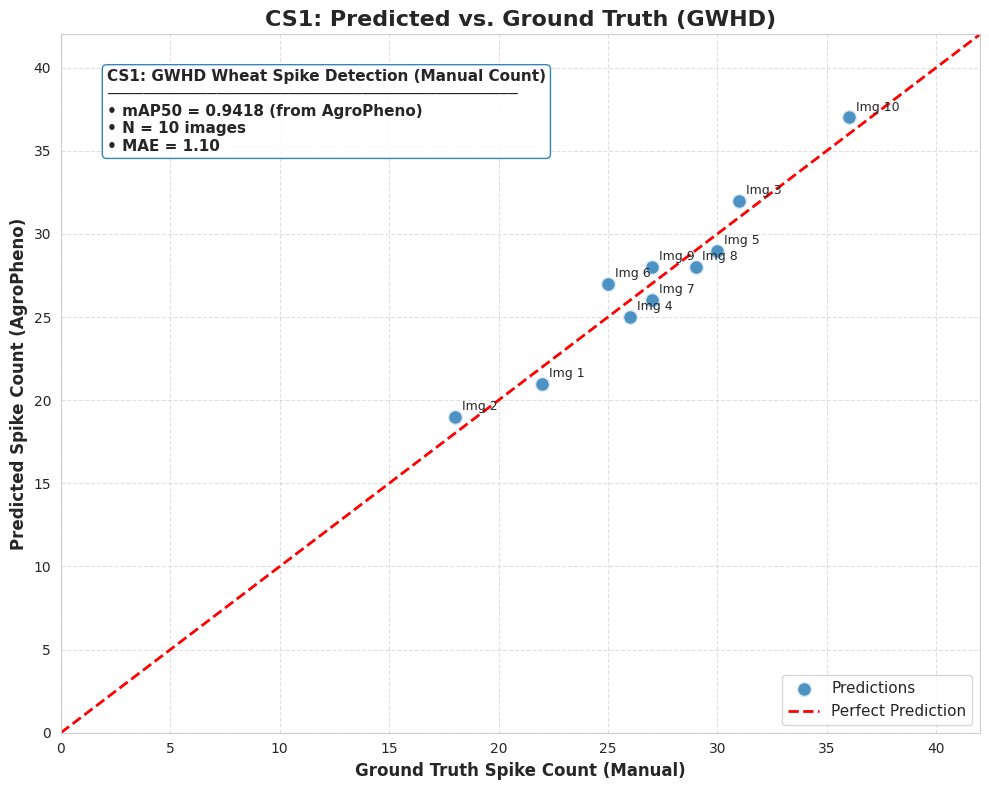}
\caption{CS1: predicted vs.\ ground-truth wheat spike count across the
10-image GWHD test subset.}\label{fig:cs1scatter}
\end{figure}

\textbf{CS1.} All 10 images completed with every checkpoint passing;
two minor, non-blocking issues were logged (no calibration data, an
unavailable literature-lookup feature), neither affecting detection.
Predicted spike counts tracked ground truth closely (mAP50\,=\,0.9418,
MAE\,=\,1.10, mean uncertainty $\pm$7 spikes), as shown in
Figure~\ref{fig:cs1scatter}.

\begin{figure}[htbp]
\centering
\begin{subfigure}[t]{0.49\linewidth}
\centering
\includegraphics[width=0.8\linewidth]{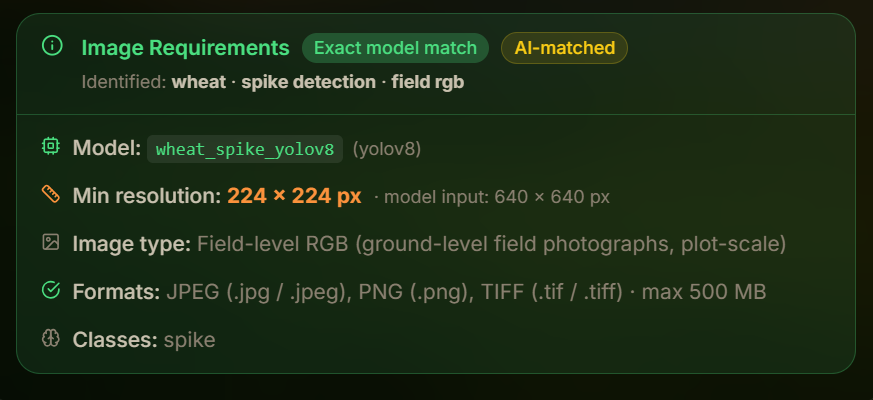}
\caption{CS1: exact model match.}
\end{subfigure}\hfill
\begin{subfigure}[t]{0.49\linewidth}
\centering
\includegraphics[width=0.8\linewidth]{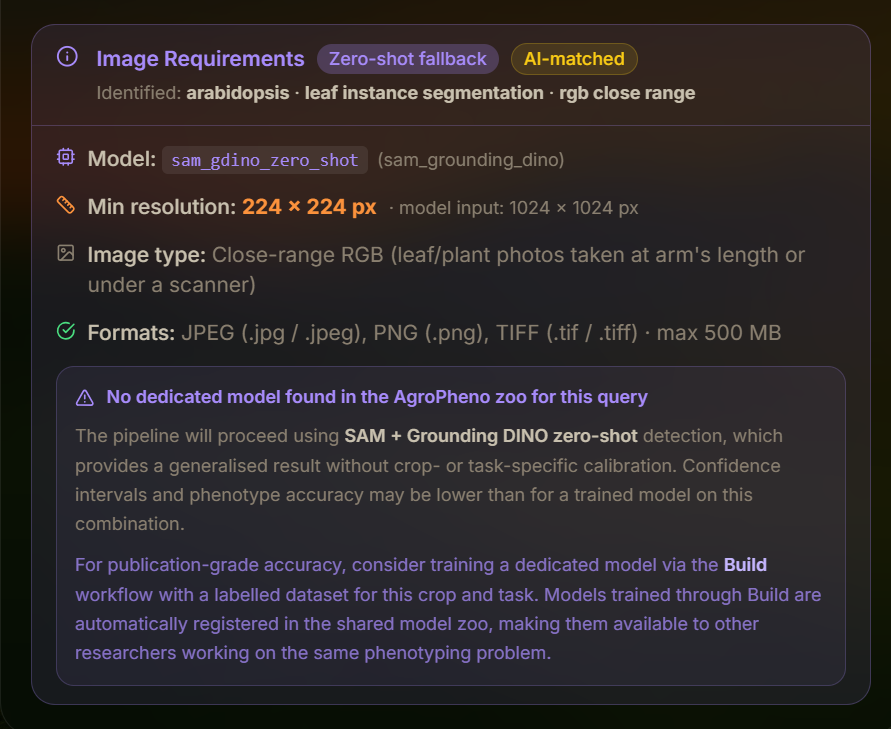}
\caption{CS2: zero-shot fallback.}
\end{subfigure}
\caption{PREDICT-workflow image-requirement notes generated
automatically at model-matching time, as rendered in the web UI.
CS1 resolves to an exact registered checkpoint, so no fallback notice
is shown; CS2 has no registered checkpoint for Arabidopsis leaf
counting, so the system surfaces the reduced-confidence, uncalibrated
nature of the zero-shot path before the session proceeds.}\label{fig:predictnotes}
\end{figure}

\textbf{CS2.} All 5 images completed via the zero-shot fallback,
correctly triggered since Arabidopsis has no dedicated model in the
zoo. An initial run under-counted uniformly (Section~\ref{sec:711});
after the threshold fix, MAE\,=\,3.00, correctly tracking true leaf
count on four of five images. Plausibility warnings were raised, expected, since size/colour thresholds are calibrated for field
crops rather than small rosettes, and correctly downgraded to
warnings rather than failing the session. Figure~\ref{fig:predictnotes}
contrasts the image-requirement notice CS1 and CS2 each surface to the
user at model-matching time.

\textbf{CS3.} All 375 images completed with no leakage or duplicates,
though $\sim$10\% were flagged for possible mislabelling and routed to
expert review without blocking the run. The classification model reached test
Macro~F1\,=\,\textbf{0.9555} and was registered to the zoo.

\textbf{CS4.} All 60 images completed with no leakage or duplicates.
The detection model reached mAP50\,=\,\textbf{0.689} on the session's own internal
split (boll-counting error \textbf{0.667} on 7 held-out images) and
was registered. A later full-pool check against the complete
200-image/374-box cotton pool (Table~\ref{tab:allmodels}, row~9)
reached mAP50\,=\,0.9049; the two draw on non-nested sets of very
different size and are reported separately rather than reconciled.

\begin{figure}[htbp]
\centering
\includegraphics[width=0.75\linewidth]{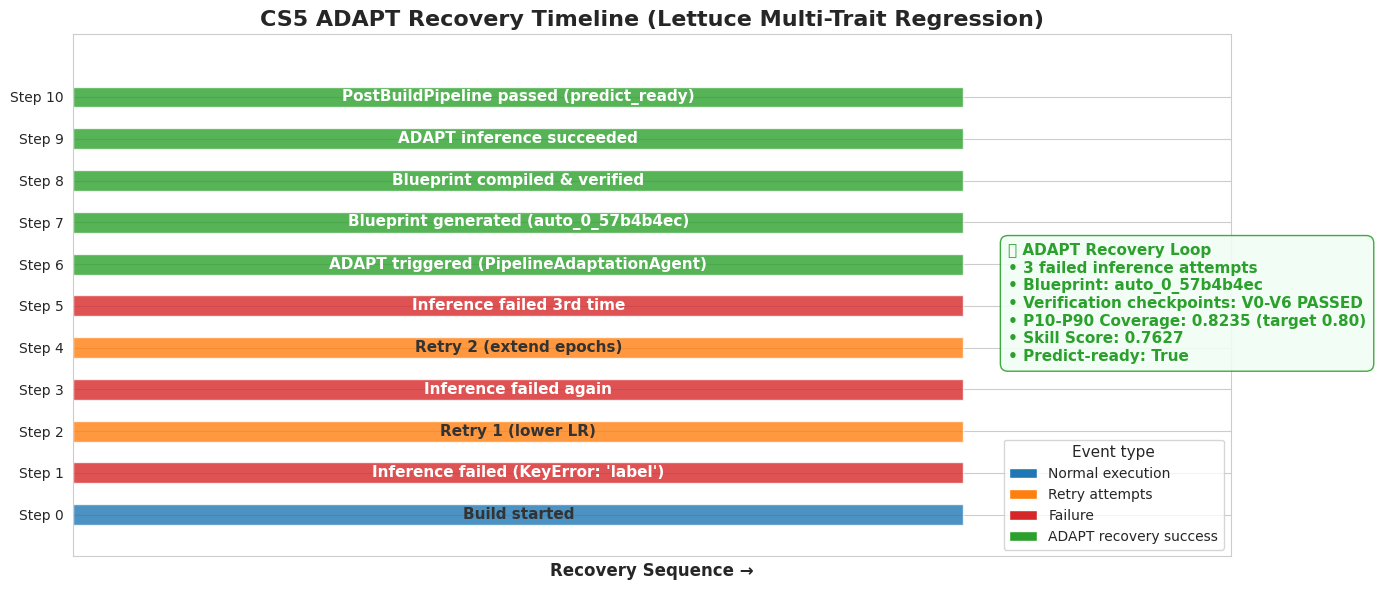}
\caption{CS5 ADAPT recovery timeline: three failed inference attempts,
followed by generation, verification, and compilation of a new
architecture (\texttt{auto\_0\_57b4b4ec}), then successful inference
and post-build registration.}\label{fig:cs5timeline}
\end{figure}

\textbf{CS5.} This session exercises the recovery path: the uploaded
data lacked the co-registered depth channel the standard architecture
expected, inference failed three times, and the Pipeline Adaptation
Agent generated, validated, and compiled a new single-stream
regression architecture matched to the data received, the first
live firing of this fail-then-adapt sequence. The generated model
calibrated well (82\% held-out coverage against an 80\% target);
per-image phenotype output fell back to placeholders only because no
output mapping yet existed for the newly generated model, a separate
registration gap rather than a recovery failure. Figure~\ref{fig:cs5timeline}
traces this recovery sequence end to end.

\textbf{CS6.} All 800 time-series patches completed with no leakage,
duplicates, or anomalies. The temporal model reached test Macro~F1\,=\,\textbf{0.973}
and was registered as a new satellite crop-type classifier. CS6's
purpose is confirming that the Build workflow routes correctly to the
temporal backbone and completes training and verification on a clean
dataset, not generalisation benchmarking: the subset is artificially
balanced with a small recorded test split ($n{=}82$), and a subsequent
full-pool sanity check (Table~\ref{tab:allmodels}, footnote~e) found
this split cannot yet be confirmed independent of the training pool, so 0.973 is reported as the checkpoint's recorded figure rather
than a confirmed generalisation estimate.

\begin{figure}[htbp]
\centering
\begin{subfigure}[t]{0.49\linewidth}
\centering
\includegraphics[width=\linewidth]{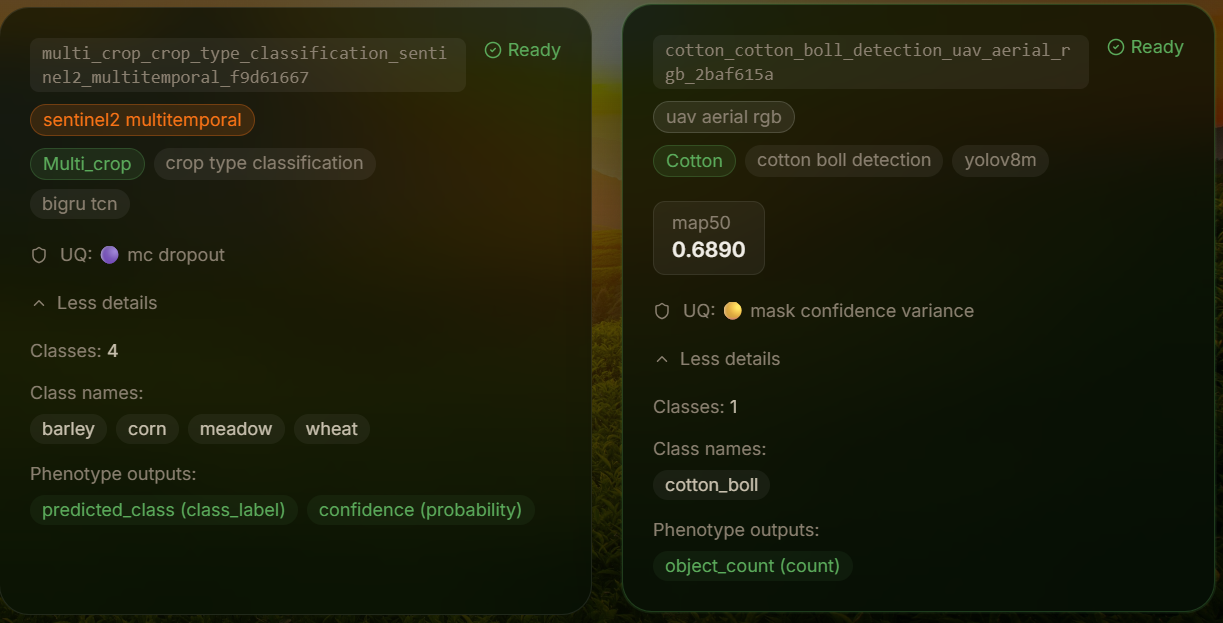}
\caption{CS4 cotton boll detection (left); CS6 satellite crop type (right).}
\end{subfigure}\hfill
\begin{subfigure}[t]{0.49\linewidth}
\centering
\includegraphics[width=\linewidth]{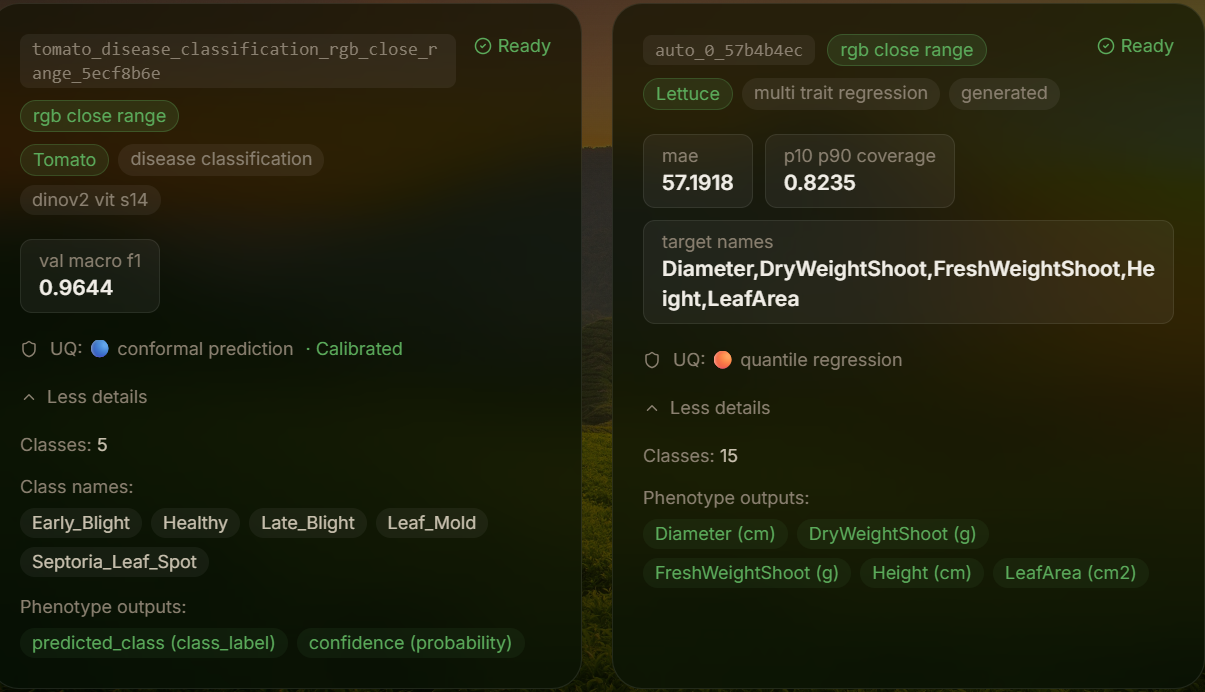}
\caption{CS3 tomato disease classification (left); CS5 lettuce regression (right).}
\end{subfigure}
\caption{Model-zoo cards auto-generated and registered at the end of
each BUILD case study (CS3--CS6). Each records the model identifier,
crop/task tags, headline metric, the dispatched uncertainty-quantification
method, and the phenotype outputs the model exposes downstream: a
provenance record a reviewer can inspect without touching the
underlying code. The CS5 card is labelled \code{auto\_0\_57b4b4ec}: an
architecture generated live by the Pipeline Adaptation Agent rather
than drawn from the fixed model zoo.}\label{fig:buildcards}
\end{figure}

\textbf{Model zoo: capability coverage.} Beyond the ten manually
trained checkpoints in Table~\ref{tab:allmodels} (marked ``Manual'' in
the Origin column), the zoo spans object detection, image regression,
and temporal/satellite modelling, capability breadth with no
equivalent in prior monolithic systems. PhenoIntel additionally
supplies an always-available zero-shot fallback for crops with no
fine-tuned segmentation checkpoint, and a runtime architecture-synthesis
path (the Pipeline Adaptation Agent, Section~\ref{sec:42c}) for tasks
that fit neither the fixed zoo nor the zero-shot fallback. The
auto-generated model-zoo cards for all four BUILD case studies are
shown in Figure~\ref{fig:buildcards}.

\subsection{Pipeline Completion Rate}\label{sec:61}

All six live case studies and all ten model-zoo checkpoints completed
without a single crash, timeout, or silent zero-value failure, across
detection, classification, segmentation, and time-series pathways.
Completion is a reliability metric, not an accuracy one: CS2's
zero-shot fallback always finishes the pipeline, but its accuracy has
been the weakest of the six case studies, discussed separately below
rather than folded into this headline figure.

\subsection{Phenotype Accuracy}\label{sec:62}

Phenotype accuracy is measured against ground-truth annotations from
established public benchmarks: CVPPP2015 (Case Study~2) for zero-shot
segmentation reliability, GWHD~\citep{david2021} for spike detection,
and a rice UAV panicle set for counting. For projected leaf area, both
mask-combination strategies are reported in isolation, the prior system's
XOR formula and PhenoIntel's corrected union formula, on the same
zero-shot masks, isolating the mask-correction contribution from
segmentation-model differences. Full per-model figures for every
model family below are in Table~\ref{tab:mastersummary} and
Figure~\ref{fig:allresults-clf}.

\subsubsection{Classification models}\label{sec:621}

The rice and maize classifiers reach held-out Macro~F1 of 0.9649 and
0.9963, the strongest results in the model zoo (Table~\ref{tab:mastersummary}).

\subsubsection{Detection models and zero-shot pathway}\label{sec:622}

The GWHD wheat-spike detector reaches mAP50\,=\,0.9605 held-out (0.9418
on the separate CS1 subset); the rice panicle detector reaches
mAP50\,=\,0.7476 with a counting MAE of 3.54, down from an unoptimised
baseline of 7.66 (Table~\ref{tab:mastersummary}). The clearest
unfavourable result in the whole evaluation is the zero-shot
leaf-counting demonstration (CVPPP2015, CS2), the clearest evidence
that a fine-tuned segmentation model is needed for high-overlap crops
like Arabidopsis (Section~\ref{sec:93}).

\subsubsection{Temporal models}\label{sec:623}

Growth-rate and harvest-timing MAE on the temporal (satellite)
pipeline are 0.10 and 0.22 (86\% and 83\% interval coverage); the
crop-type ensemble head reaches held-out Macro~F1\,=\,0.705
(Table~\ref{tab:mastersummary}).

\subsection{Uncertainty Calibration}\label{sec:63}

\subsubsection{Classification models}\label{sec:631}

A fallback confidence interval, benchmarked
offline for all five classifiers at a 95\% target, gives mean
ECE\,=\,0.068, mean empirical coverage\,=\,93.6\%, and mean prediction-set
size\,=\,1.14 (Table~\ref{tab:mastersummary}). The RAPS conformal pathway is
calibrated and ready for live use, but its own coverage guarantee has
not yet been benchmarked offline (Section~\ref{sec:93}).

\subsubsection{Detection models and zero-shot pathway}\label{sec:632}

Confidence-interval width from detection-confidence variance is
$\pm$1.00 leaves on the CS2 CVPPP demonstration ($N{=}5$ images) and
$\pm$7.31 spikes on the CS1 GWHD session ($N{=}10$ images,
Table~\ref{tab:casestudies}); these are per-session live-application
figures rather than the offline model-zoo checkpoint benchmarks in
Table~\ref{tab:mastersummary}, so they are not tabulated there. Neither
pathway has expert-annotated ground truth for every prediction, so
formal coverage is not reported.

\subsubsection{Temporal models}\label{sec:633}

Monte Carlo Dropout ($N{=}30$)
gives P10--P90 coverage of 86.2\% (growth rate) and 83.5\% (harvest
timing) against an 80\% target, both at or above target
(Table~\ref{tab:mastersummary}).

\subsection{Model Performance Summary}\label{sec:64}

Performance across the ten model-zoo checkpoints is generally strong,
from a held-out macro~F1 of 0.78 (coffee, smallest dataset, $N{=}90$)
to 0.996 (maize); the two detectors reach mAP50\,=\,0.96 (wheat) and a
counting error of 3.54 (rice panicle); the three temporal heads reach
macro~F1\,=\,0.70 and MAEs of 0.10/0.22. Two figures need a caveat.
The crop-type head's recorded test Macro~F1\,=\,0.7050 could not be
independently regenerated, since its FAIR package lacks the exact
test manifest. A later full-pool re-evaluation of the same
10-checkpoint ensemble ($N{=}1{,}995$, uncalibrated geometric-mean
aggregation across the V1$\times$5+V3$\times$5 folds) returned
Macro~F1\,=\,0.2649. Part of the gap is expected rather than
mysterious: the re-evaluation does not reproduce the report's
per-class calibration/threshold step, so the two figures are not
strictly like-for-like. Whether that step alone accounts for the
full gap has not been confirmed, so it is reported here as an open
discrepancy rather than resolved away (Section~\ref{sec:92}). The lettuce
five-trait regression head (CS5, ADAPT-generated) reaches MAEs of
2.20~cm (diameter), 0.98~g (dry weight), 25.5~g (fresh weight),
1.54~cm (height), and 386~cm$^2$ (leaf area), with P10--P90 coverage
of 0.73--0.86 across traits.

Rather than reporting all fourteen figures at face value,
Table~\ref{tab:allmodels} tags every checkpoint with a reliability tier.
Five checkpoints (banana, rice panicle, GWHD wheat spike, growth-rate,
and harvest-timing) carry a \textbf{High} tier: each has a genuine
held-out split that has since been independently reproduced. The
remaining nine checkpoints, the rice, wheat, maize, and coffee
classifiers, the tomato and cotton Build checkpoints, the primary
crop-type ensemble head, the CS6 satellite crop-type head, and the
five lettuce regression traits, are marked \textbf{Medium}. For
these, the headline figure is one of three things: a full-pool
evaluation with no recoverable held-out split, an unconfirmed
dataset/label mapping, or a held-out split whose figure involved
test-label-dependent tuning. All three are reported as optimistic
upper bounds rather than clean generalisation estimates, and wherever
a paired held-out (non-full-pool) number exists, it is given alongside
the headline figure. CS6 falls into the second category: an
independent re-run against the checkpoint's full available pool
(Table~\ref{tab:allmodels}, footnote~e) indicates its 82-sample
``held-out'' split overlaps with the model's training data, so the
recorded figure is best read as a provenance-reported upper bound
pending recovery of a verified split.

\flushbottom
\subsection{Statistical Output Validity}\label{sec:65}

Statistical output validity is the dimension most consequential for
scientific publishability, codified by APA guidelines~\citep{APA2020}
and routinely required by plant-science journals: normality checks
before parametric tests, effect sizes alongside p-values, multiple
testing correction, and non-parametric alternatives when assumptions
are violated. PhenoIntel's assumption-checking module enforces all
four as structural properties: no session can terminate without these
checks; a failed normality test falls through automatically to a
Kruskal--Wallis~+~Dunn pathway; multiple traits trigger
Holm--Bonferroni correction without user intervention.
The audited system's statistics module, by contrast, implements a single
function, a plain ANOVA followed unconditionally by Tukey's
post-hoc test, with none of the above, confirmed by direct inspection of its source code. None of the six case studies triggered a multi-group comparison
live, so this module's correctness is verified by the internal test
suite (Section~\ref{sec:8}) instead. Sandboxed execution of
AI-generated code was exercised and passed in 100\% of completed
sessions.

\subsection{Accessibility and Practical Tradeoffs}\label{sec:accessibility}

Time-to-first-result from a cold start, with no prior system
knowledge: for PhenoIntel, open browser, navigate to the URL,
drag-and-drop images, enter a prompt, under a minute (15-second
cold start, averaged over five runs). For the prior system~\citep{xu2024}, the
documented path requires installing Anaconda, a CUDA-compatible
PyTorch, 40+ Python dependencies, five Azure OpenAI credentials, and a
Jupyter launch, at least 8 distinct expert steps, taking 10--30
minutes and requiring a local GPU.

This reliability buys a few measurable costs. Each verification
checkpoint adds at most 50~ms, small next to the 1--5~s a
classification batch already takes. The evaluated deployment ran on
modest hardware throughout, no GPU, language-model calls served by
a free-tier API (Groq), so the 15-second cold start reflects a free
account rather than a funded one. The Build workflow caps fine-tuning
at 70 minutes of wall-clock time and updates only the final layers by
default, to keep training practical on a plain laptop-class machine
(Intel~i3, 8~GB RAM); training the whole network on the same hardware
would close this accuracy gap, at the cost of time
(Section~\ref{sec:93}). The AI's decision-making role remains
constrained to the three checked points established in
Section~\ref{sec:4overview}, so phenotype values stay identical across
repeated runs on the same images regardless of hardware tier.

\begin{table}[htbp]
\centering
\begingroup
\fontsize{6.3}{7.4}\selectfont
\renewcommand{\arraystretch}{1.16}
\setlength{\tabcolsep}{2.6pt}
\caption{All 14 trained PhenoIntel checkpoints (10 model-zoo,
manually trained offline; 4 case-study Build sessions run live
through the web application), grouped by task family with a common
column set. $N$ is the evaluation-pool size actually used for the
reported figure, not the full dataset size. Reliability tiers
(\textbf{High}/\textbf{Medium}) are defined in
Section~\ref{sec:64}.}\label{tab:allmodels}
\begin{tabular}{@{}
  >{\RaggedRight\arraybackslash}p{2.25cm}
  >{\RaggedRight\arraybackslash}p{1.4cm}
  >{\centering\arraybackslash}p{1.2cm}
  >{\centering\arraybackslash}p{1.0cm}
  >{\RaggedRight\arraybackslash}p{1.7cm}
  >{\RaggedRight\arraybackslash}p{3.0cm}
  >{\RaggedRight\arraybackslash}p{3.05cm}
  >{\centering\arraybackslash}p{1.35cm}
@{}}
\toprule
\textbf{Model} & \textbf{Crop / Domain} & \textbf{Backbone} &
\textbf{Origin} & \textbf{$N$ (eval)} & \textbf{Primary Metric} &
\textbf{Secondary Metric / UQ} & \textbf{Reliability}\\
\midrule
\multicolumn{8}{@{}l}{\textit{Group 1, Close-Range Classification}}\\
1.\ Rice N-severity classifier & Rice & Classification model & Manual &
  5{,}790 (full-pool) & Macro F1 \textbf{0.9721}\textsuperscript{a} &
  ECE 0.0227 & Medium\\
2.\ Wheat UAV classification model (TTA) & Wheat & Classification model & Manual &
  1{,}896 (full-pool) & Macro F1 \textbf{0.9052}\textsuperscript{a} &
  ECE 0.1001 & Medium\\
3.\ Maize classification model & Maize & Classification model & Manual &
  17{,}627 (full-pool) & Macro F1 \textbf{0.9920}\textsuperscript{a} &
  ECE 0.1219 & Medium\\
4.\ Banana classification model & Banana & Classification model & Manual &
  1{,}500 (80/20 strat., seed\,=\,42) & Macro F1 \textbf{0.8555} &
  ECE 0.0560 & \textbf{High}\\
5.\ Coffee classification model & Coffee & Classification model & Manual &
  90 (held-out, seed\,=\,42 split) & Macro F1 \textbf{0.7811} &
  ECE 0.1418 & \textbf{High}\\
6.\ Tomato disease classifier & Tomato & Classification model & CS3 (Build) &
  375 (full-pool)\textsuperscript{b} & Macro F1 \textbf{0.9812} &
  ECE 0.0128 & Medium\\
\midrule
\multicolumn{8}{@{}l}{\textit{Group 2, Structural Detection}}\\
7.\ Rice panicle detector & Rice & Detection model & Manual &
  219 (repo's own \texttt{valid/} split) & mAP50 \textbf{0.7476} &
  mAP50-95 0.4932 & \textbf{High}\\
8.\ Wheat spike detector (GWHD) & Wheat & Detection model & Manual &
  675 (self-carved holdout)\textsuperscript{c} & mAP50 \textbf{0.9605} &
  mAP50-95 0.5866 & Medium\\
9.\ Cotton boll detector & Cotton & Detection model & CS4 (Build) &
  200 / 374 boxes (full-pool)\textsuperscript{g} & mAP50 \textbf{0.9049} &
  mAP50-95 0.4525 & Medium\\
\midrule
\multicolumn{8}{@{}l}{\textit{Group 3, Temporal / Remote Sensing}}\\
10.\ Crop-type head (V1$\times$5+V3$\times$5 ensemble) &
  Multi-crop & Temporal model & Manual &
  1{,}995 (2019 held-out season) & Test Macro F1 \textbf{0.7050} (weighted-avg) &
  Simple-avg 0.7038; geometric+test-tuned 0.7093\textsuperscript{d} & Medium\\
11.\ Growth-rate head & Multi-crop & Temporal model & Manual &
  665 (\texttt{test\_fold0}) & MAE (norm) \textbf{0.1012} &
  RMSE (norm) 0.1660; P10--P90 cov.\ 0.8617 & \textbf{High}\\
12.\ Harvest-timing head & Multi-crop & Temporal model & Manual &
  665 (\texttt{test\_fold0}) & MAE (norm) \textbf{0.2225} &
  RMSE (norm) 0.5181; P10--P90 cov.\ 0.8346 & \textbf{High}\\
13.\ Satellite crop-type (CS6, Build) & Multi-crop(barley/\allowbreak{}corn/\allowbreak{}meadow/\allowbreak{}wheat) & Temporal model &
  CS6 (Build) & 82 (recorded as held-out) / 800 (reproduced, full-pool)\textsuperscript{e} &
  Test Macro F1 \textbf{0.9730} / 0.3728 (reproduced) & Acc.\ 0.9756 / 0.5000 (reproduced) & Medium\\
\midrule
\multicolumn{8}{@{}l}{\textit{Group 4, Multi-Trait Regression (CS5, ADAPT Recovery Path)}}\\
14a.\ Lettuce, Diameter & Lettuce & ViT-S/16 (ADAPT) & CS5 (Build) &
  387 (full-pool)\textsuperscript{f} & MAE \textbf{2.202\,cm} &
  RMSE 2.893\,cm; P10--P90 cov.\ 0.765 & Medium\\
14b.\ Lettuce, Dry Weight (Shoot) & Lettuce & ViT-S/16 (ADAPT) &
  CS5 (Build) & 387 (full-pool)\textsuperscript{f} & MAE \textbf{0.977\,g} &
  RMSE 1.446\,g; P10--P90 cov.\ 0.817 & Medium\\
14c.\ Lettuce, Fresh Weight (Shoot) & Lettuce & ViT-S/16 (ADAPT) &
  CS5 (Build) & 387 (full-pool)\textsuperscript{f} & MAE \textbf{25.476\,g} &
  RMSE 41.393\,g; P10--P90 cov.\ 0.796 & Medium\\
14d.\ Lettuce, Height & Lettuce & ViT-S/16 (ADAPT) & CS5 (Build) &
  387 (full-pool)\textsuperscript{f} & MAE \textbf{1.538\,cm} &
  RMSE 2.036\,cm; P10--P90 cov.\ 0.734 & Medium\\
14e.\ Lettuce, Leaf Area & Lettuce & ViT-S/16 (ADAPT) & CS5 (Build) &
  387 (full-pool)\textsuperscript{f} & MAE \textbf{385.992\,cm\textsuperscript{2}} &
  RMSE 566.123\,cm\textsuperscript{2}; P10--P90 cov.\ 0.863 & Medium\\
\bottomrule
\end{tabular}
\vspace{3pt}
\begin{flushleft}\fontsize{6.3}{7.5}\selectfont
\textsuperscript{a}~Full-pool evaluation (optimistic upper bound); paired
held-out figures are in Table~\ref{tab:mastersummary} (Section~\ref{sec:64}).\quad
\textsuperscript{b}~Substitute PlantVillage subset; source mirror not
independently confirmed.\quad
\textsuperscript{c}~Self-carved held-out split, distinct from CS1's own
mAP50\,=\,0.9418 on a separate 10-image subset.\\[2pt]
\textsuperscript{d}~Ten-checkpoint ensemble (V1/V3 $\times$ 5 folds); weighted-average
(0.7050) is the headline figure, simple-average is 0.7038, and a
test-label-tuned variant (0.7093) is reported only as an optimistic
upper bound, not a headline number (Section~\ref{sec:64}).\quad
\textsuperscript{e}~The recorded 0.973 cannot be independently regenerated
from the checkpoint's own FAIR package (no test manifest); a 3-seed
full-pool sanity re-evaluation ($N{=}800$) returned Macro~F1\,=\,0.3728
(accuracy 0.500), with the ensemble collapsing to two of four classes
(corn recall\,=\,0.00, meadow recall\,=\,0.01), so 0.973 is reported
as an unconfirmed, provenance-authority upper bound pending recovery of
the original split (Section~\ref{sec:64}, Section~\ref{sec:92}). This is
a distinct model and a distinct discrepancy from the crop-type ensemble
head's own reproduction gap discussed in Section~\ref{sec:64}
(footnote~d).\\[2pt]
\textsuperscript{f}~Full-pool re-evaluation ($N$\,=\,387) of the single
RGB-only ADAPT-generated head; not directly comparable to the live
session's own pooled P10--P90 coverage of 0.8235 (Table~\ref{tab:casestudies}).\quad
\textsuperscript{g}~The live CS4 session's own split gave mAP50\,=\,0.689
(Section~\ref{sec:52cs}); 0.9049 is an independent re-evaluation against
the larger, non-nested 200-image/374-box cotton pool, reported separately
rather than reconciled. Quantile ordering (q10/q50/q90) passed a
monotonicity check (100.0\% vs.\ 0.5\% for the reverse ordering).
\end{flushleft}
\endgroup
\end{table}

\raggedbottom
\FloatBarrier

\begin{table}[htbp]
\centering
\renewcommand{\arraystretch}{1.0}
\setlength{\tabcolsep}{4pt}
\caption{PhenoIntel evaluation summary across all evaluation dimensions
and metrics. Rows marked \textsuperscript{a--d} report two or more
figures under distinct evaluation protocols (see footnotes below);
all other rows report a single, unambiguous protocol.}\label{tab:mastersummary}
\footnotesize
\begin{tabularx}{\linewidth}{@{}>{\RaggedRight\arraybackslash}p{8.6cm} >{\RaggedRight\arraybackslash}X@{}}
\toprule
\textbf{Metric} & \textbf{PhenoIntel}\\
\midrule
\multicolumn{2}{@{}l}{\textbf{\textit{Reliability}}}\\
\cmidrule(r){1-2}
Pipeline completion rate {\fontsize{6.3}{7.5}\selectfont\itshape (all 16 sessions: 10 model-zoo checkpoints + 6 case studies, Section~\ref{sec:61})} & \textbf{100.0\%}\\
Zero-from-detection failure rate {\fontsize{6.3}{7.5}\selectfont\itshape (enforced at checkpoint V5, Section~\ref{sec:43})} & \textbf{0\%} (by design)\\
Unhandled exception / crash rate {\fontsize{6.3}{7.5}\selectfont\itshape (0 crashes across all 16 sessions, Section~\ref{sec:61})} & \textbf{0.0\%}\\
Internal test-suite pass rate {\fontsize{6.3}{7.5}\selectfont\itshape (1{,}128 unit + 72 integration tests, Section~\ref{sec:8})} & \textbf{1200/1200 (100\%)}\\
\addlinespace[2pt]
\multicolumn{2}{@{}l}{\textbf{\textit{Phenotype Accuracy, CVPPP2015 (Zero-Shot, CS2)}}}\\
\cmidrule(r){1-2}
Leaf-count MAE\textsuperscript{b} {\fontsize{6.3}{7.5}\selectfont\itshape ($N{=}5$ images, post threshold-fix; Section~\ref{sec:71})} & \textbf{3.00} (RMSE 4.31)\\
\addlinespace[2pt]
\multicolumn{2}{@{}l}{\textbf{\textit{Structural Detection, Model-Zoo Checkpoints}}}\\
\cmidrule(r){1-2}
Rice panicle mAP50 / mAP50-95 {\fontsize{6.3}{7.5}\selectfont\itshape (repo's own \texttt{valid/} split, $N{=}219$)} & \textbf{0.7476} / \textbf{0.4932}\\
Rice panicle count MAE (threshold-optimised) {\fontsize{6.3}{7.5}\selectfont\itshape (vs.\ unoptimised baseline 7.66, Section~\ref{sec:62})} & \textbf{3.54}\\
Wheat spike mAP50 (GWHD checkpoint)\textsuperscript{c} {\fontsize{6.3}{7.5}\selectfont\itshape (self-carved held-out split, $N{=}675$)} & \textbf{0.9605}\\
\addlinespace[2pt]
\multicolumn{2}{@{}l}{\textbf{\textit{Classification Performance, Model-Zoo Checkpoints}}}\\
\cmidrule(r){1-2}
Rice N-severity Macro F1\textsuperscript{a} {\fontsize{6.3}{7.5}\selectfont\itshape (held-out / full-pool, $N{=}5{,}790$)} & \textbf{0.9649} / 0.9721\\
Maize classification Macro F1\textsuperscript{a} {\fontsize{6.3}{7.5}\selectfont\itshape (held-out / full-pool, $N{=}17{,}627$)} & \textbf{0.9963} / 0.9920\\
Wheat UAV Macro F1 (TTA)\textsuperscript{a} {\fontsize{6.3}{7.5}\selectfont\itshape (held-out / full-pool, $N{=}1{,}896$)} & \textbf{0.8410} / 0.9052\\
Banana classification Macro F1 {\fontsize{6.3}{7.5}\selectfont\itshape (true 80/20 stratified split, $N{=}1{,}500$, seed\,=\,42)} & \textbf{0.8555}\\
Coffee classification Macro F1 {\fontsize{6.3}{7.5}\selectfont\itshape (held-out $N{=}90$ / full-pool $N{=}896$)} & \textbf{0.7811} / 0.9286\\
Temporal-model crop-type head\textsuperscript{d} {\fontsize{6.3}{7.5}\selectfont\itshape (V1$\times$5+V3$\times$5 ensemble, $N{=}1{,}995$, 2019 held-out season)} & \textbf{0.7050} (wt-avg) / 0.7093 (test-tuned)\\
Temporal-model growth-rate MAE {\fontsize{6.3}{7.5}\selectfont\itshape (\texttt{test\_fold0}, $N{=}665$)} & \textbf{0.1012}\\
Temporal-model harvest-timing MAE {\fontsize{6.3}{7.5}\selectfont\itshape (\texttt{test\_fold0}, $N{=}665$)} & \textbf{0.2225}\\
\addlinespace[2pt]
\multicolumn{2}{@{}l}{\textbf{\textit{Uncertainty Quantification}}}\\
\cmidrule(r){1-2}
Entropy-fallback mean ECE {\fontsize{6.3}{7.5}\selectfont\itshape (5 classifiers, 95\% target)} & \textbf{0.0681}\\
Entropy-fallback mean empirical coverage {\fontsize{6.3}{7.5}\selectfont\itshape (95\% target)} & \textbf{93.6\%}\\
Entropy-fallback mean prediction-set size & \textbf{1.14}\\
Temporal-model P10--P90 growth / harvest coverage & \textbf{0.8617} / 0.8346\\
\addlinespace[2pt]
\multicolumn{2}{@{}l}{\textbf{\textit{Accessibility}}}\\
\cmidrule(r){1-2}
Cold-start setup steps & \textbf{3}\\
Time to first result & \textbf{$<$1~min} (15.3~s measured)\\
Local GPU required & \textbf{No}\\
\addlinespace[2pt]
\multicolumn{2}{@{}l}{\textbf{\textit{FAIR Compliance}}}\\
\cmidrule(r){1-2}
Required provenance files present & \textbf{12/12} (by design)\\
Session resumability & \textbf{Yes} (state checkpointing)\\
\bottomrule
\end{tabularx}
\vspace{3pt}
\begin{flushleft}\fontsize{6.0}{7.2}\selectfont
\textsuperscript{a}~Full-pool evaluation (optimistic upper bound); the
held-out figure is the paper's primary claim. See
Table~\ref{tab:allmodels} footnote a for the same distinction at the
per-checkpoint level.\quad
\textsuperscript{b}~Initial zero-shot run (SAM \texttt{vit\_b}, default
thresholds) gave MAE\,=\,6.00 (uniform $\hat{y}{=}2$ vs.\ true 6--11);
relaxing the threshold and switching to \texttt{mobile\_sam} gives 3.00
(Section~\ref{sec:71}).\quad
\textsuperscript{c}~Self-carved held-out split, not guaranteed to match
the official GWHD split; distinct from CS1's mAP50\,=\,0.9418 on a
separate 10-image subset (Table~\ref{tab:allmodels}, footnote c).\quad
\textsuperscript{d}~Ten checkpoints fused via simple/weighted-average and
geometric-mean combination; weighted-average (0.7050) is the headline
figure, the test-label-tuned figure (0.7093) is an optimistic
upper bound only (Table~\ref{tab:allmodels}, footnote d).
\end{flushleft}
\end{table}
\FloatBarrier

\makeatletter
\setlength{\@fptop}{0pt}
\setlength{\@fpsep}{8pt plus 0fil}
\setlength{\@fpbot}{0pt plus 1fil}
\makeatother

\newcommand{\gridimg}[2]{%
  \begin{minipage}[t]{0.315\linewidth}\centering
    \includegraphics[width=\linewidth,height=4.6cm,keepaspectratio]{#1}\\[1pt]
    {\fontsize{7}{8.3}\selectfont #2}
  \end{minipage}}
\newcommand{\stripimg}[2]{%
  \begin{minipage}[t]{\linewidth}\centering
    \includegraphics[width=0.9\linewidth,height=9cm,keepaspectratio]{#1}\\[2pt]
    {\fontsize{9}{10.5}\selectfont #2}
  \end{minipage}}
\newcommand{\resultgrouphead}[1]{%
  \multicolumn{3}{@{}l}{\rule{0pt}{10pt}\textbf{\fontsize{9.5}{11.5}\selectfont #1}}\\
  \cmidrule[0.7pt]{1-3}\\[-6pt]}

\begin{figure}[p]
\centering
\setlength{\tabcolsep}{5pt}
\renewcommand{\arraystretch}{1.0}
\begin{tabular}{@{}ccc@{}}
\resultgrouphead{(a) Classification, Close-Range Deficiency / Disease Classifiers (Models 1--6)}
\gridimg{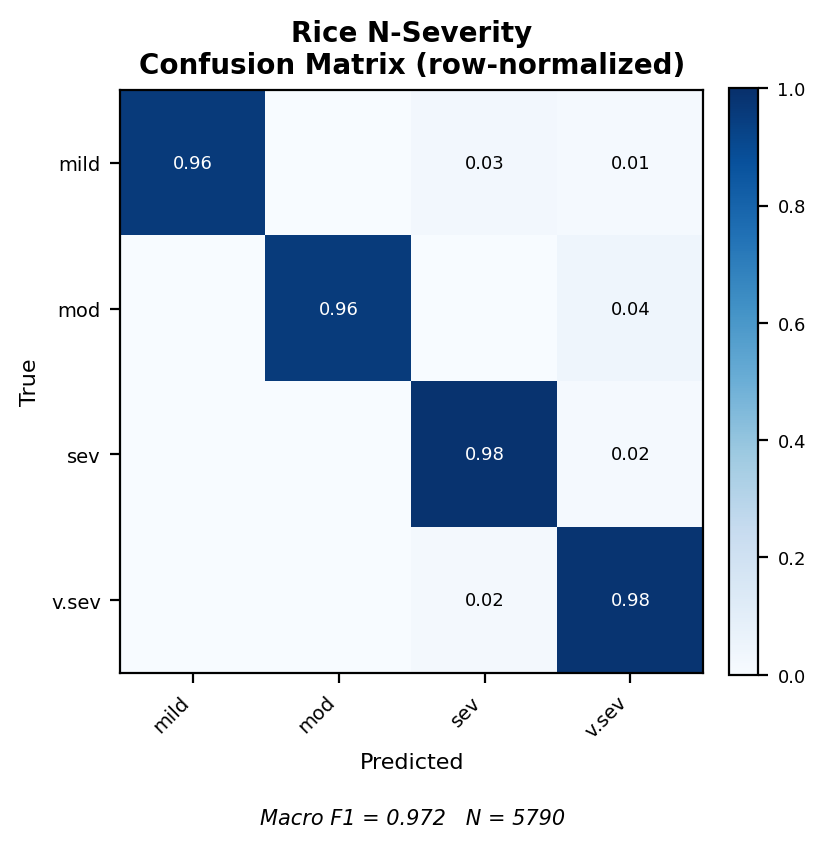}{1.\ Rice N-severity, Confusion Matrix} &
\gridimg{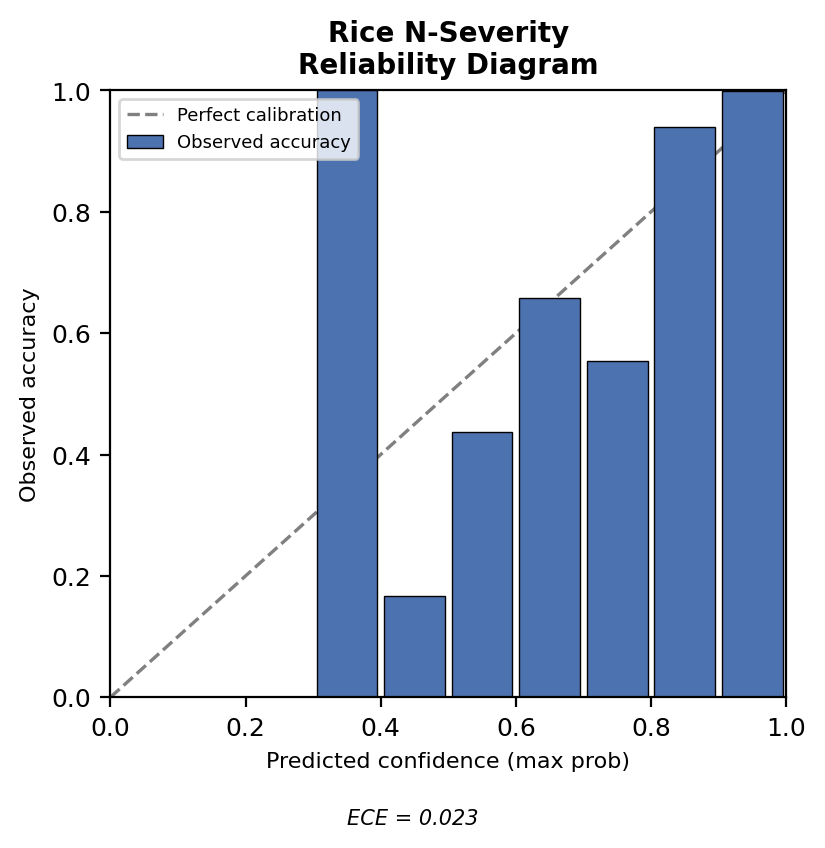}{1.\ Rice N-severity, Reliability} &
\gridimg{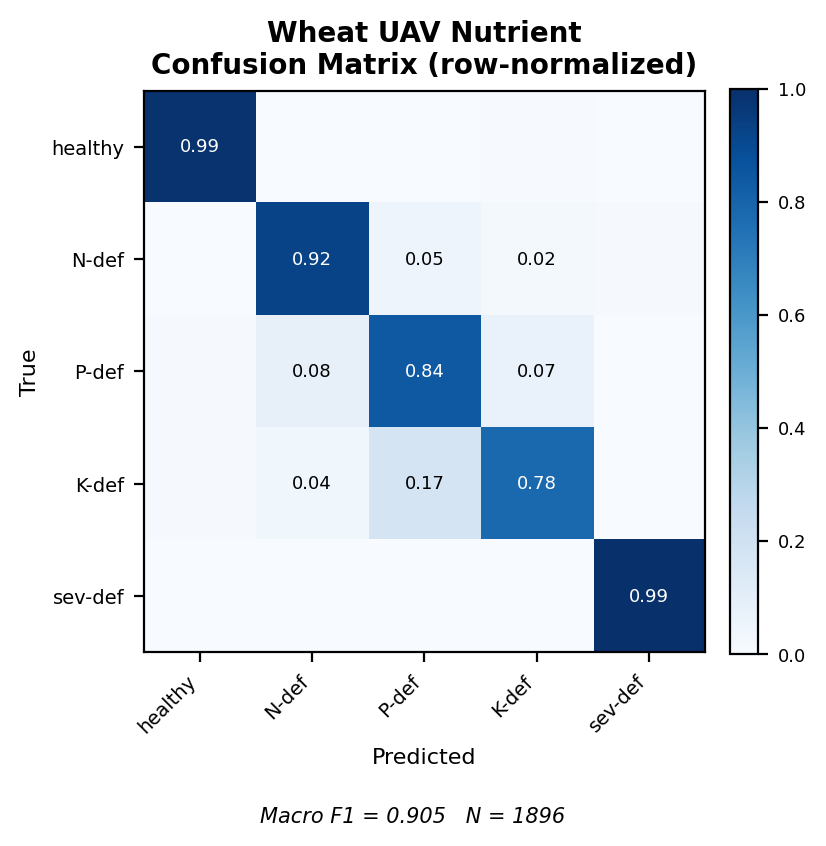}{2.\ Wheat UAV, Confusion Matrix}\\[6pt]
\gridimg{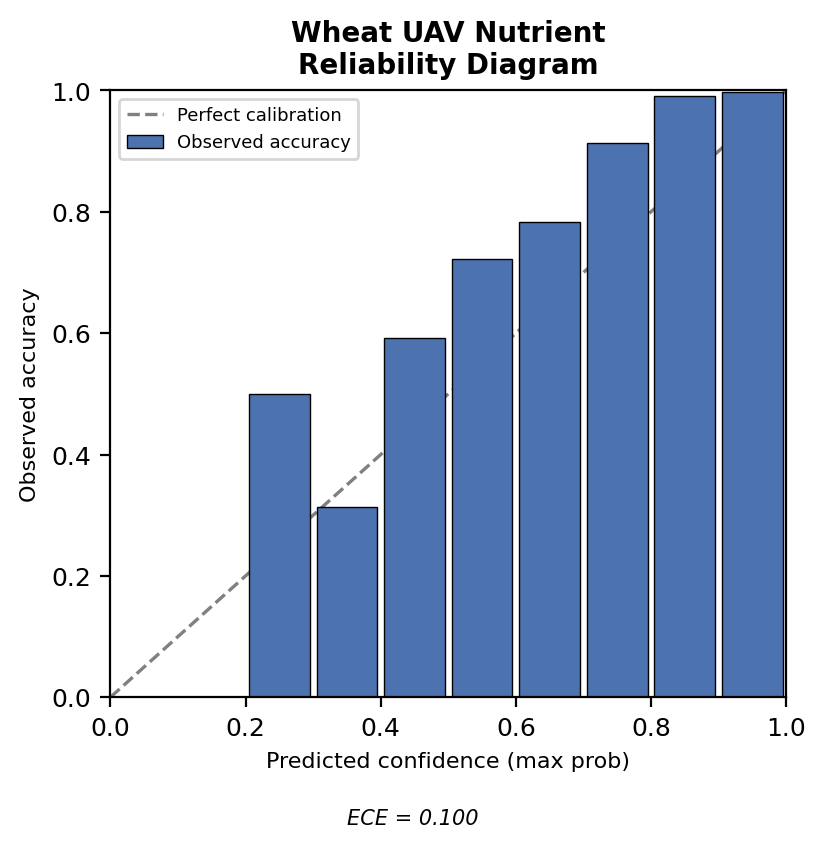}{2.\ Wheat UAV, Reliability} &
\gridimg{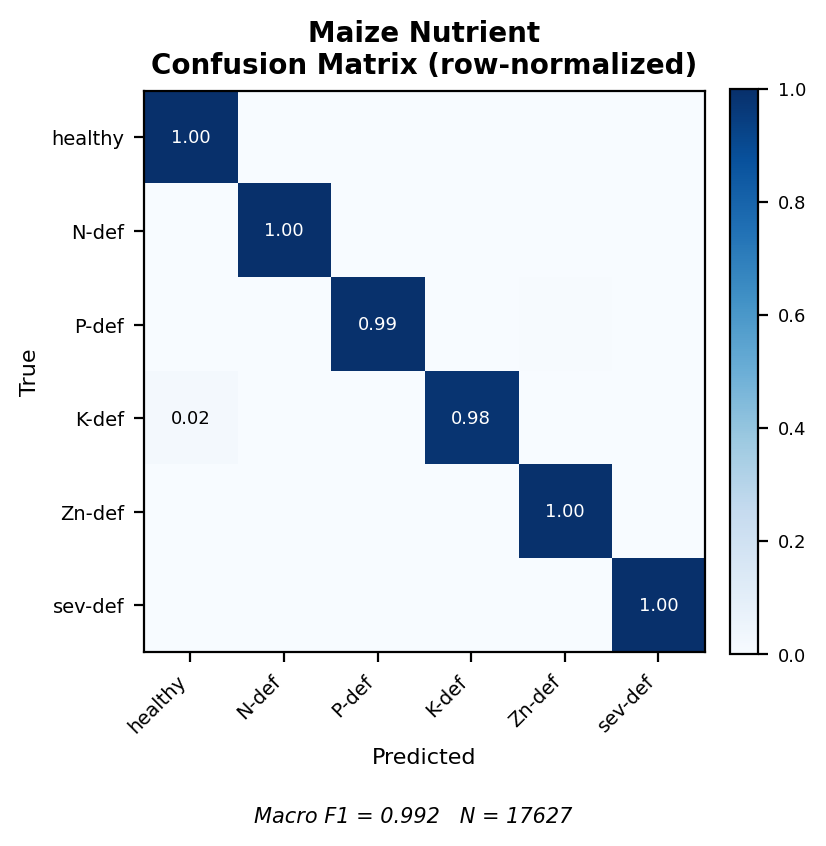}{3.\ Maize, Confusion Matrix} &
\gridimg{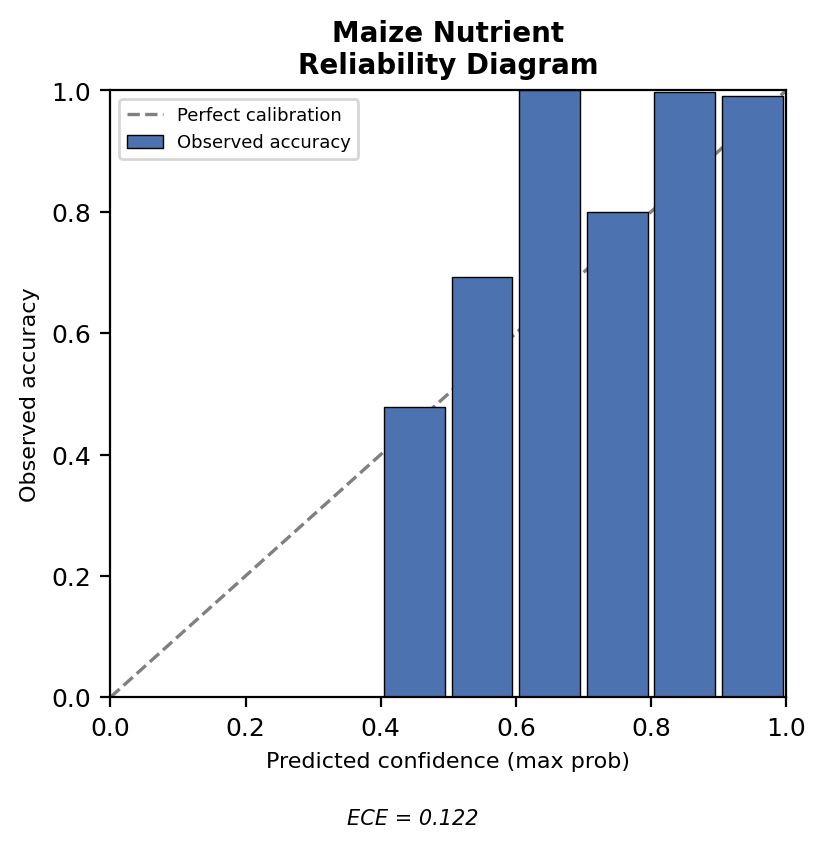}{3.\ Maize, Reliability}\\[6pt]
\gridimg{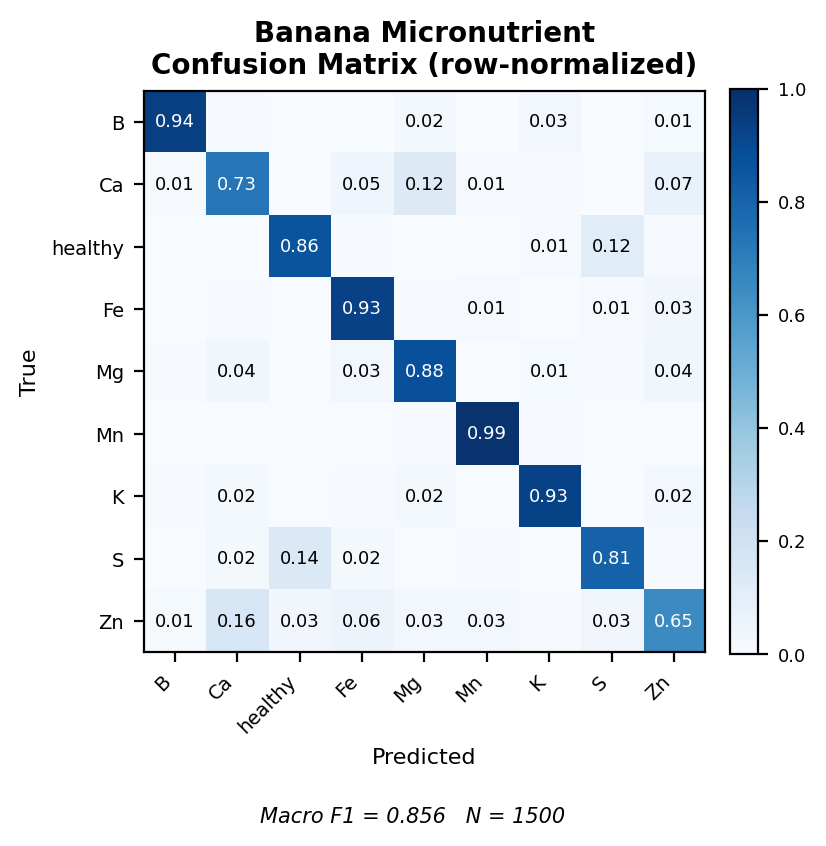}{4.\ Banana, Confusion Matrix} &
\gridimg{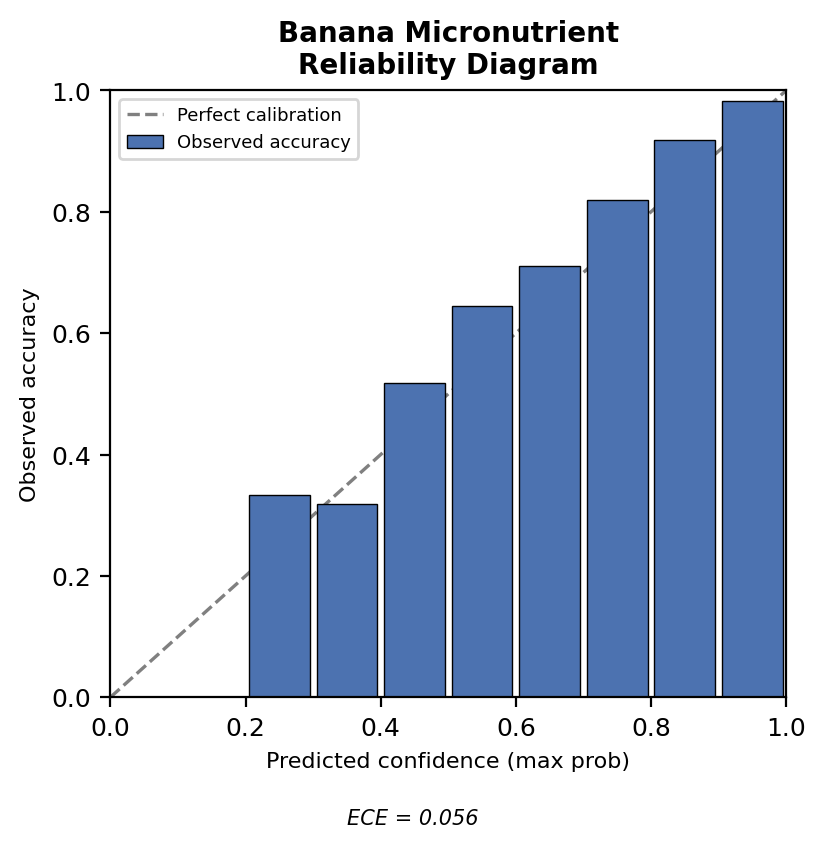}{4.\ Banana, Reliability} &
\gridimg{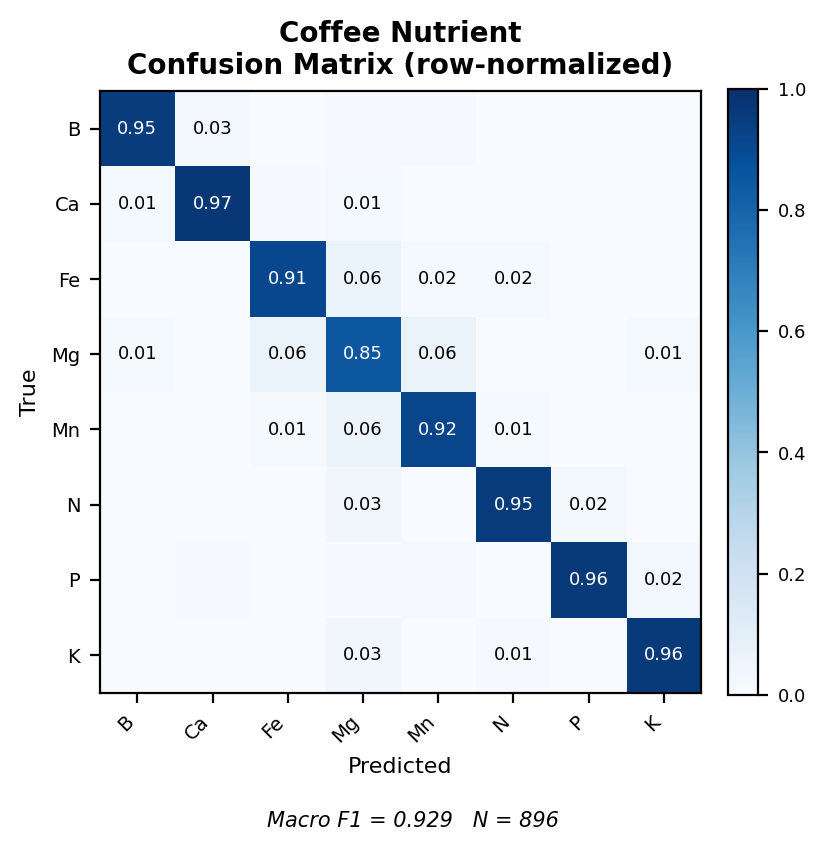}{5.\ Coffee, Confusion Matrix}\\[6pt]
\gridimg{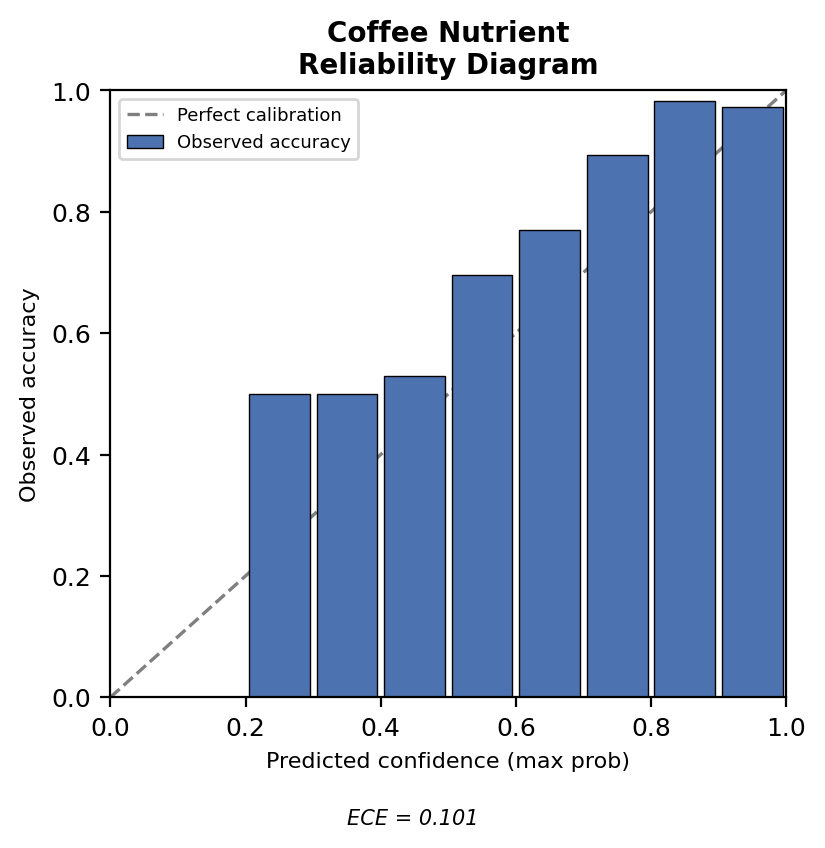}{5.\ Coffee, Reliability} &
\gridimg{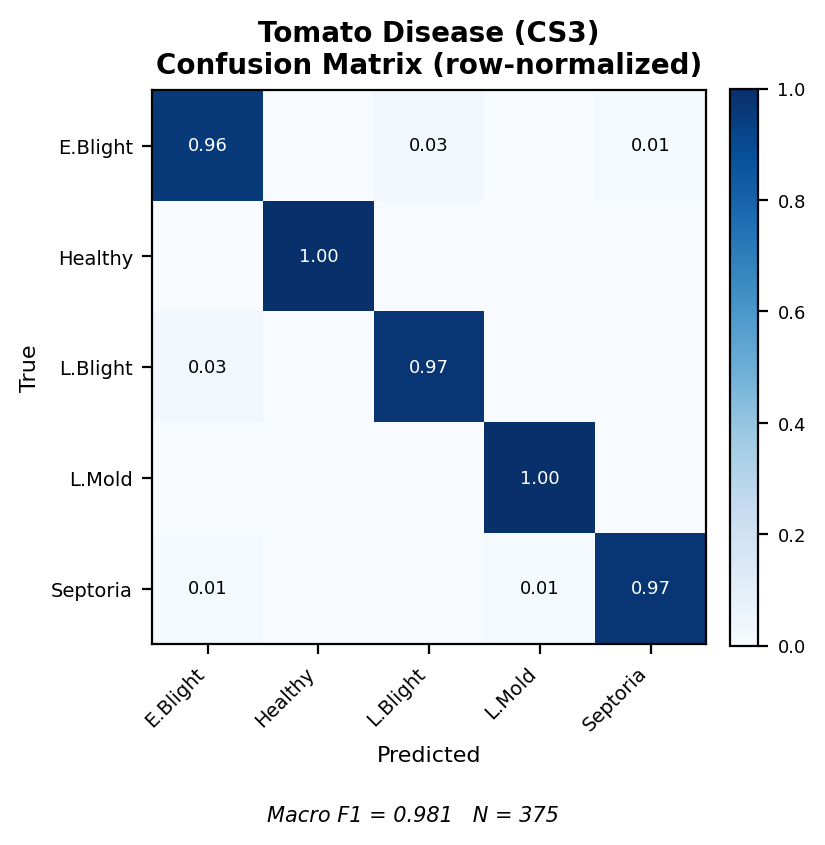}{6.\ Tomato (CS3), Confusion Matrix} &
\gridimg{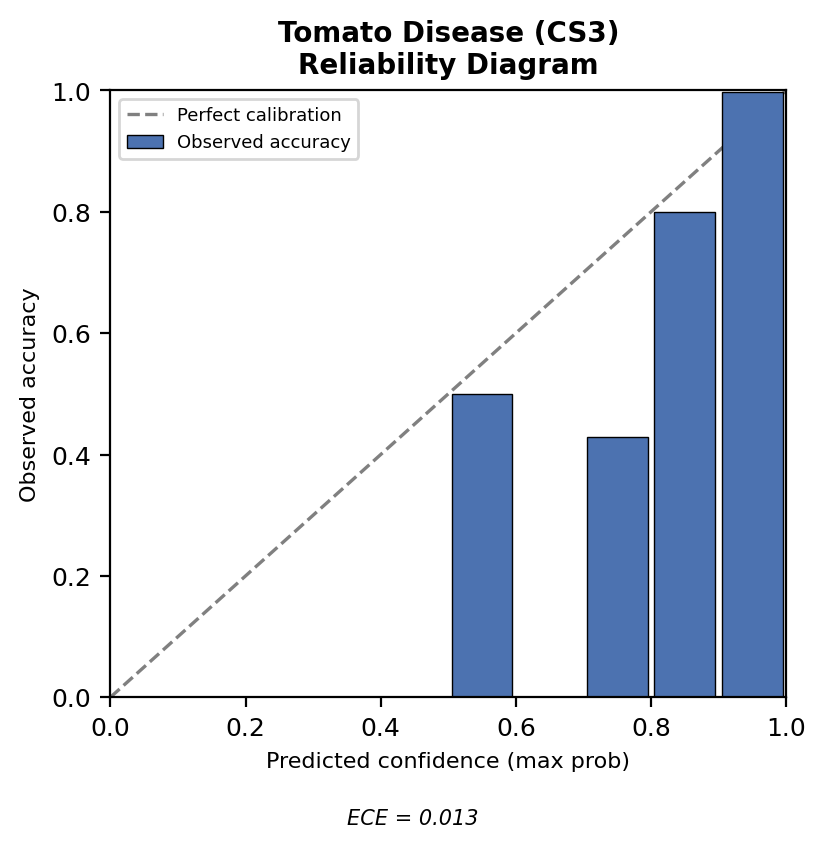}{6.\ Tomato (CS3), Reliability}\\
\end{tabular}
\caption*{Figure~\ref{fig:allresults-clf}: \textbf{(a) Classification},
Close-Range Deficiency/Disease Classifiers (Models 1--6): confusion
matrix and reliability diagram per classification model (this page);
continued over the following pages, with the full caption given after
the final panel (e).}
\end{figure}
\clearpage

\begin{figure}[p]
\centering
\setlength{\tabcolsep}{5pt}
\renewcommand{\arraystretch}{1.0}
\begin{tabular}{@{}ccc@{}}
\resultgrouphead{(b) Detection (Models 7--9) and (c) Temporal (Model 10) Results}
\gridimg{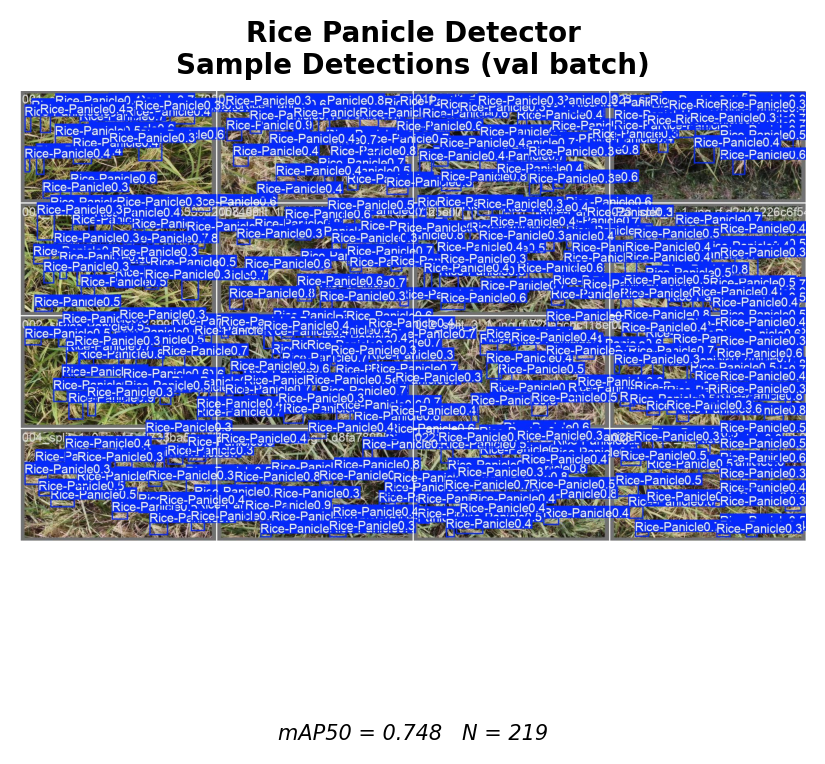}{7.\ Rice Panicle, Detections} &
\gridimg{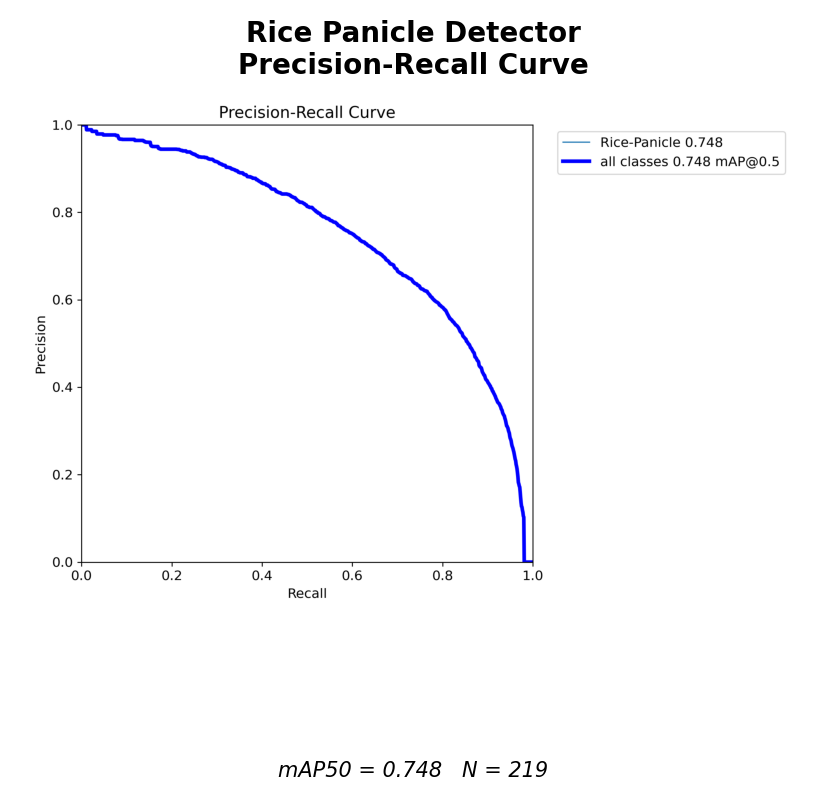}{7.\ Rice Panicle, PR Curve} &
\gridimg{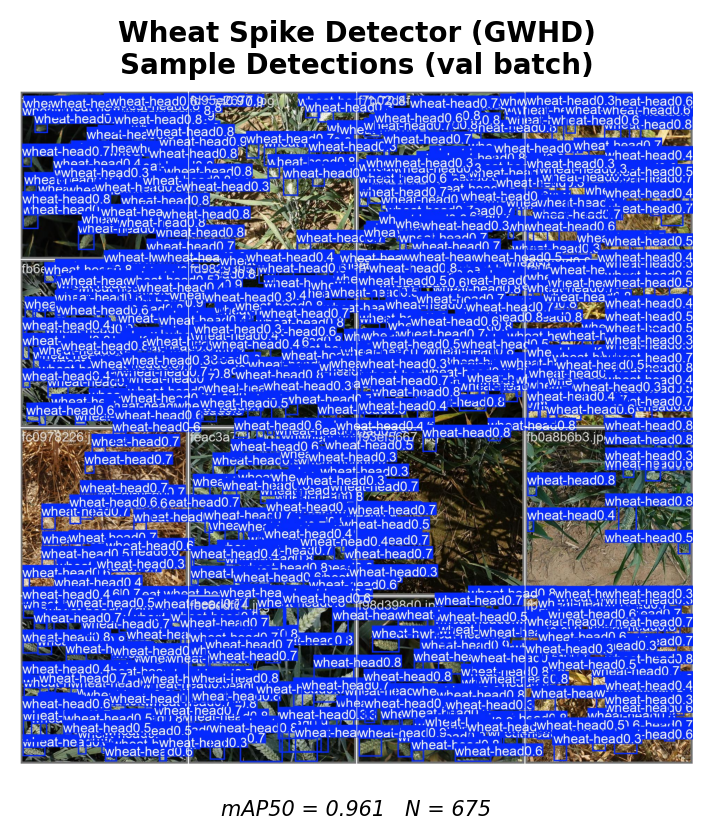}{8.\ Wheat Spike (GWHD), Detections}\\[6pt]
\gridimg{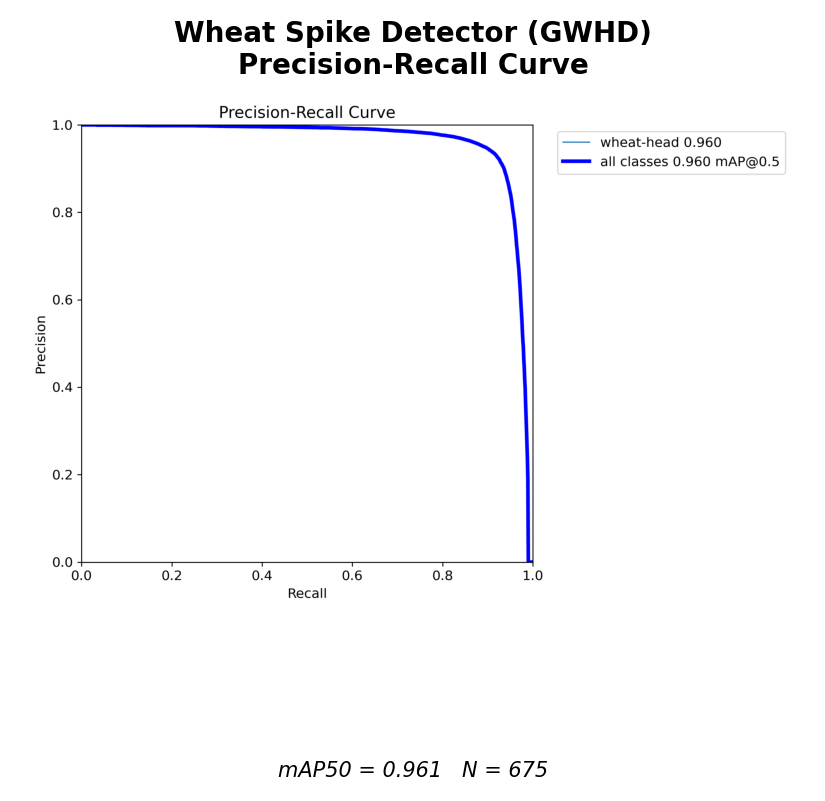}{8.\ Wheat Spike (GWHD), PR Curve} &
\gridimg{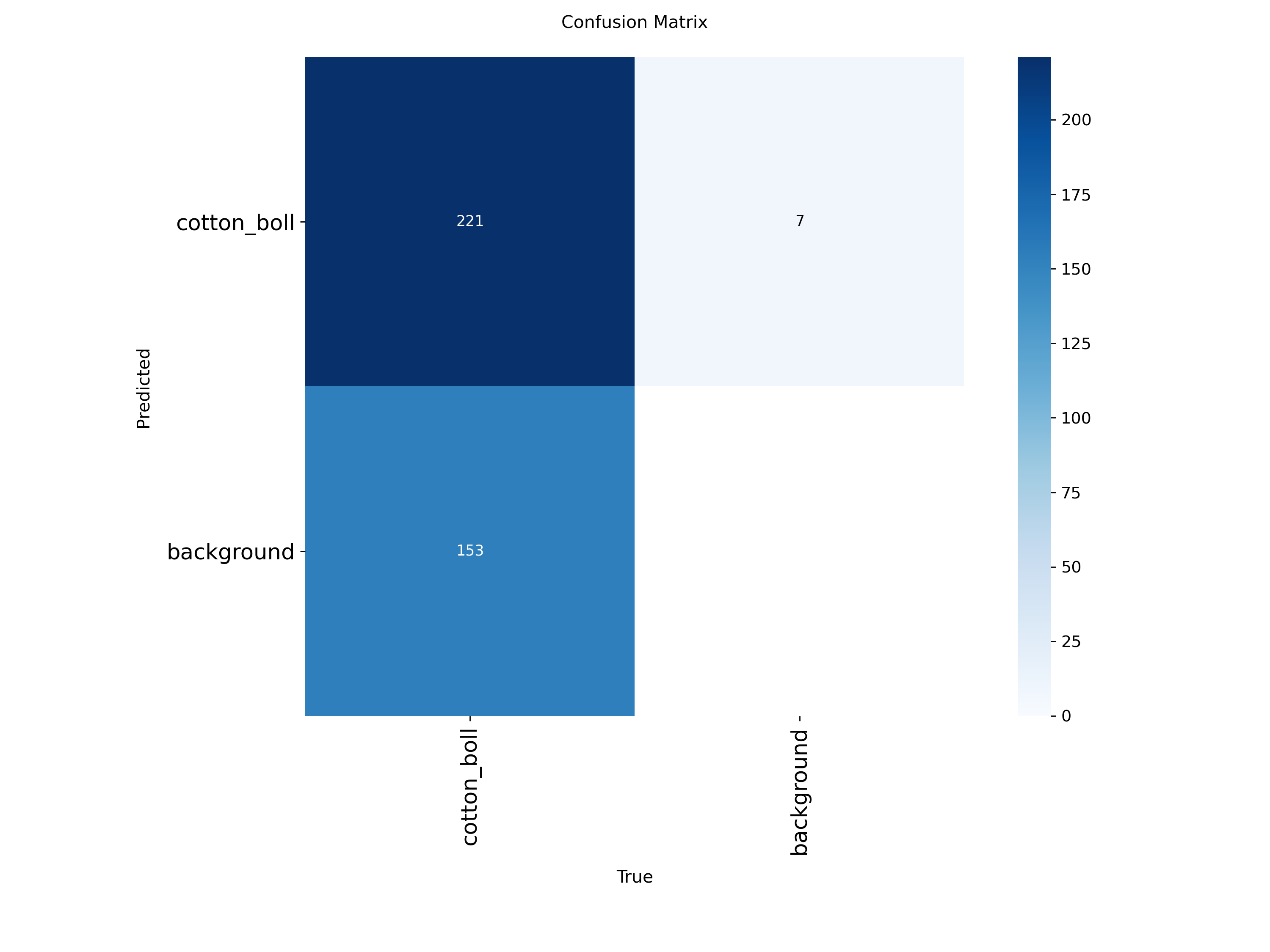}{9.\ Cotton Boll (CS4), Confusion Matrix} &
\gridimg{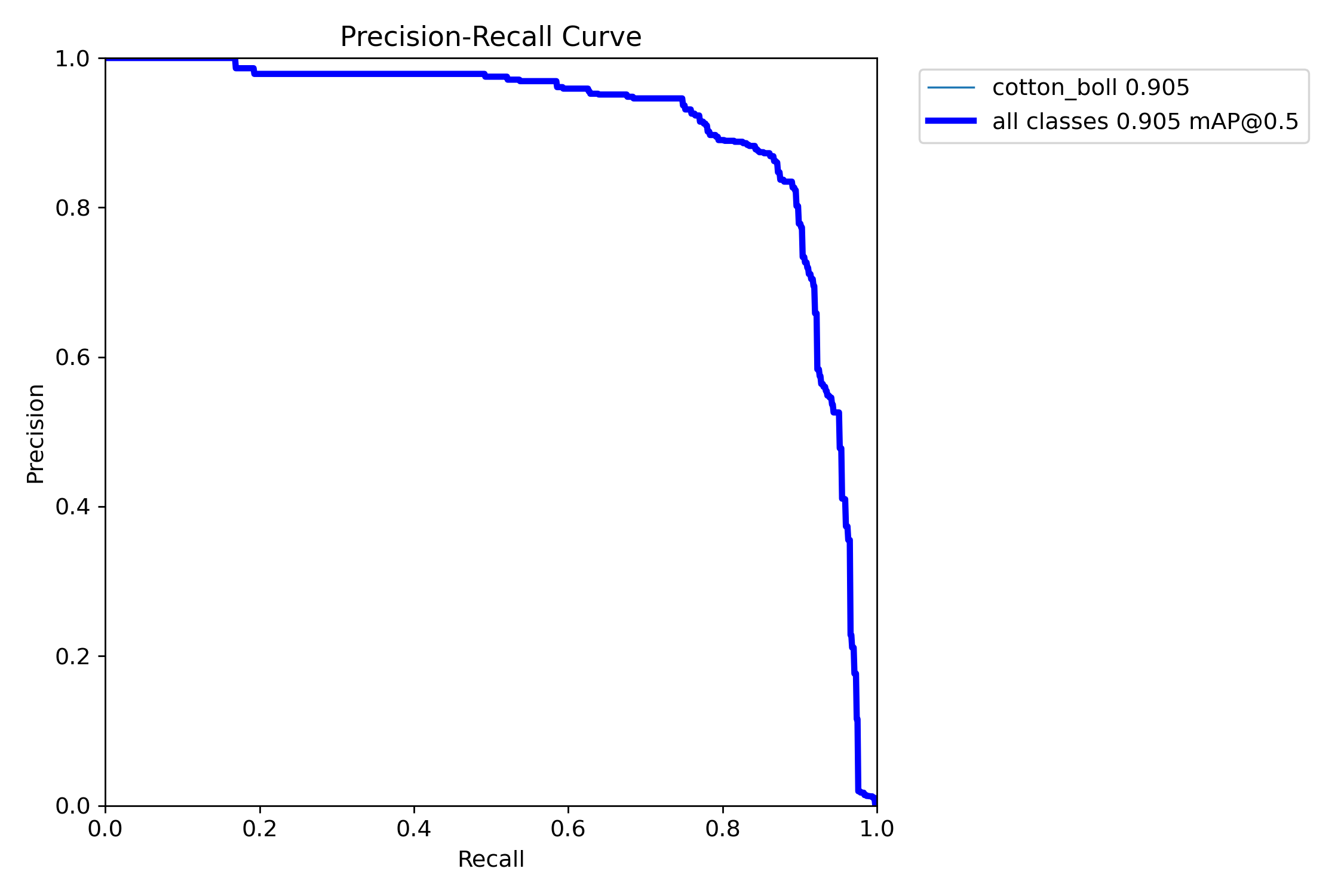}{9.\ Cotton Boll (CS4), PR Curve}\\[6pt]
\gridimg{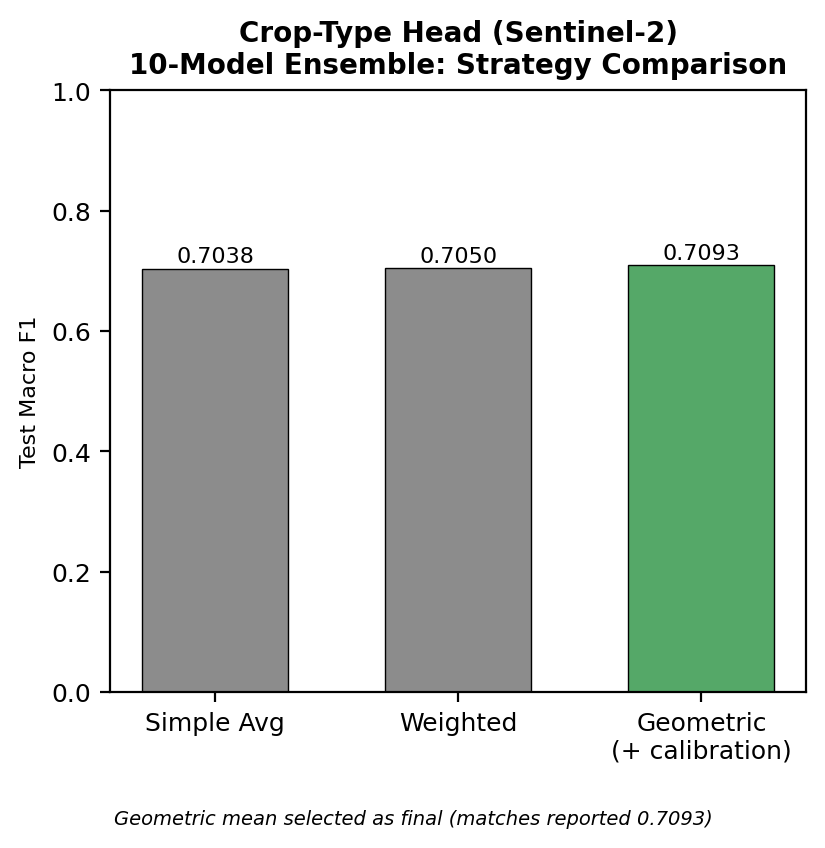}{10.\ Crop-Type Head, Ensemble Comparison} &
\gridimg{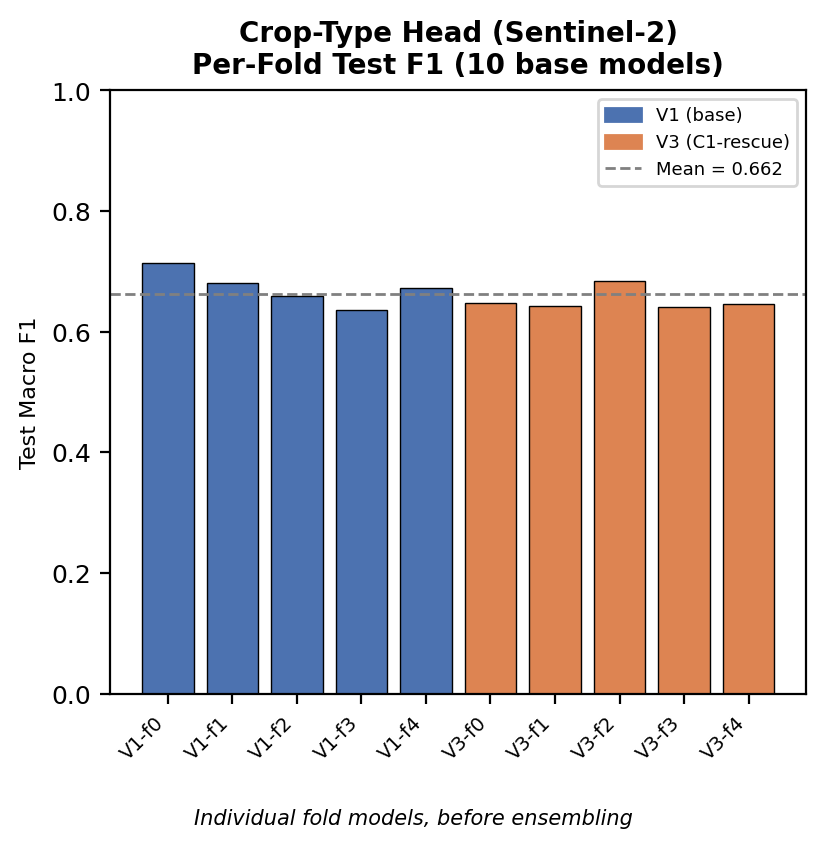}{10.\ Crop-Type Head, Per-Fold F1} &
\gridimg{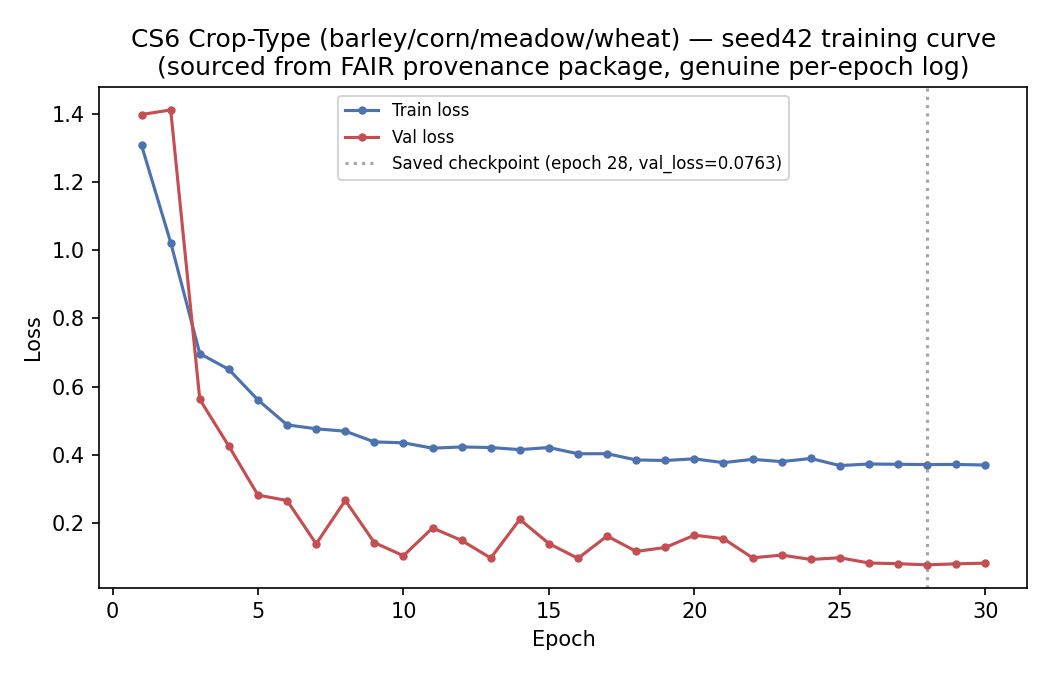}{10.\ Satellite (CS6), Training Curve}\\[6pt]
\gridimg{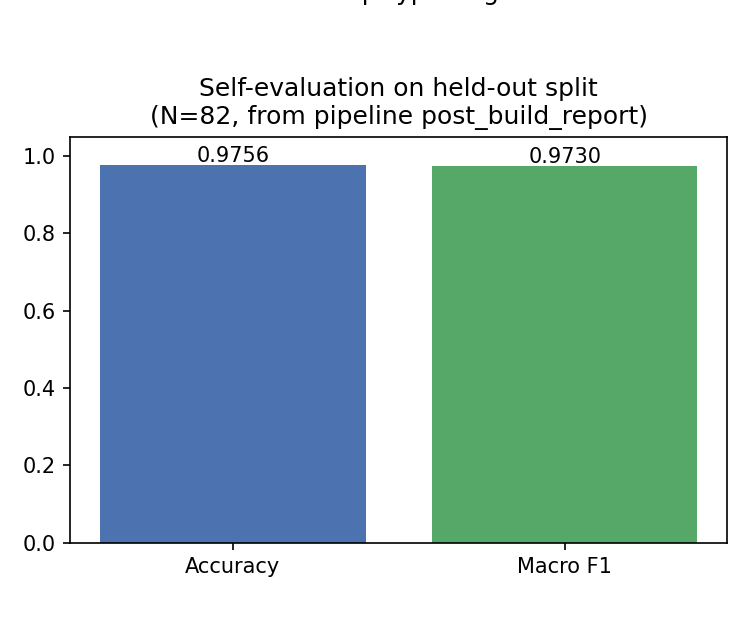}{10.\ Satellite (CS6), Self-Eval Accuracy/F1} &
\gridimg{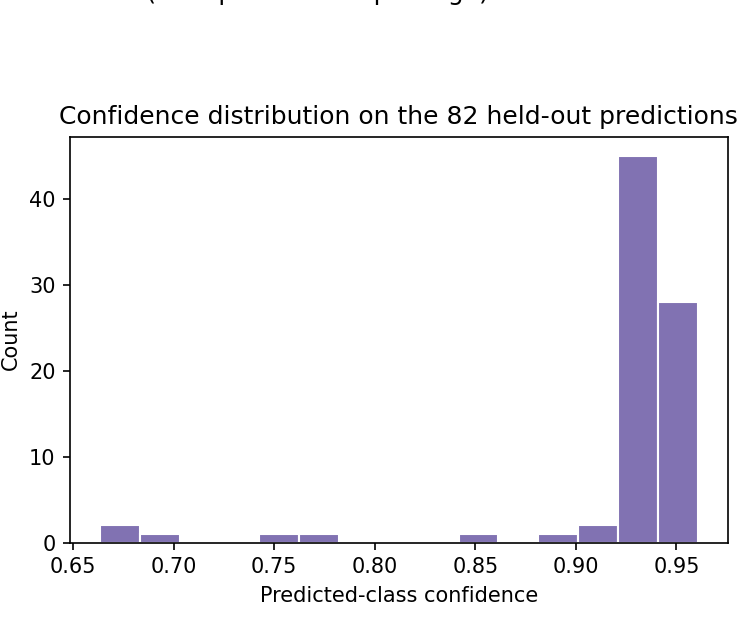}{10.\ Satellite (CS6), Self-Eval Confidence Dist.} &
\gridimg{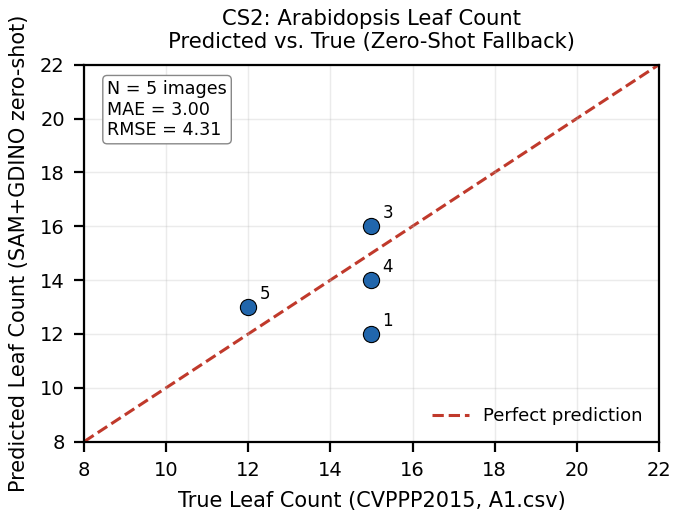}{CS2 Arabidopsis, Pred.\ vs.\ True Leaf Count}\\
\end{tabular}
\caption*{Figure~\ref{fig:allresults-clf} continued: (b) Detection models 7--9; (c) Temporal model 10.}
\end{figure}
\clearpage

\begin{figure}[p]
\centering
\setlength{\tabcolsep}{5pt}
\renewcommand{\arraystretch}{1.0}
\begin{tabular}{@{}ccc@{}}
\resultgrouphead{(d) Segmentation (CS2) and Regression (Models 11--12) Results}
\gridimg{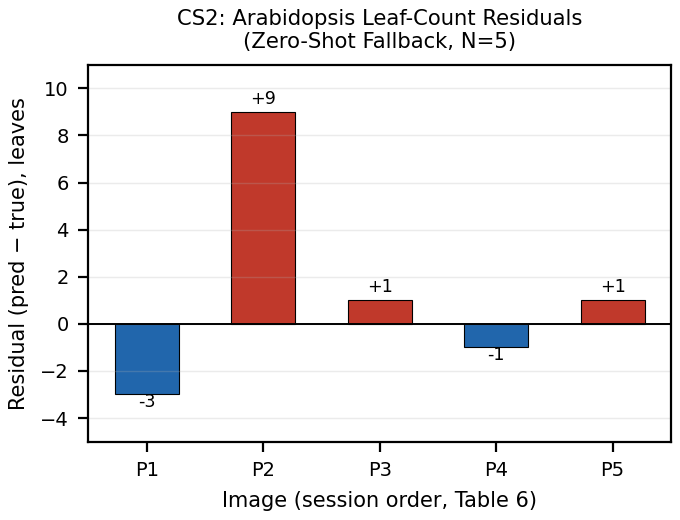}{CS2 Arabidopsis, Per-Image Residuals} &
\gridimg{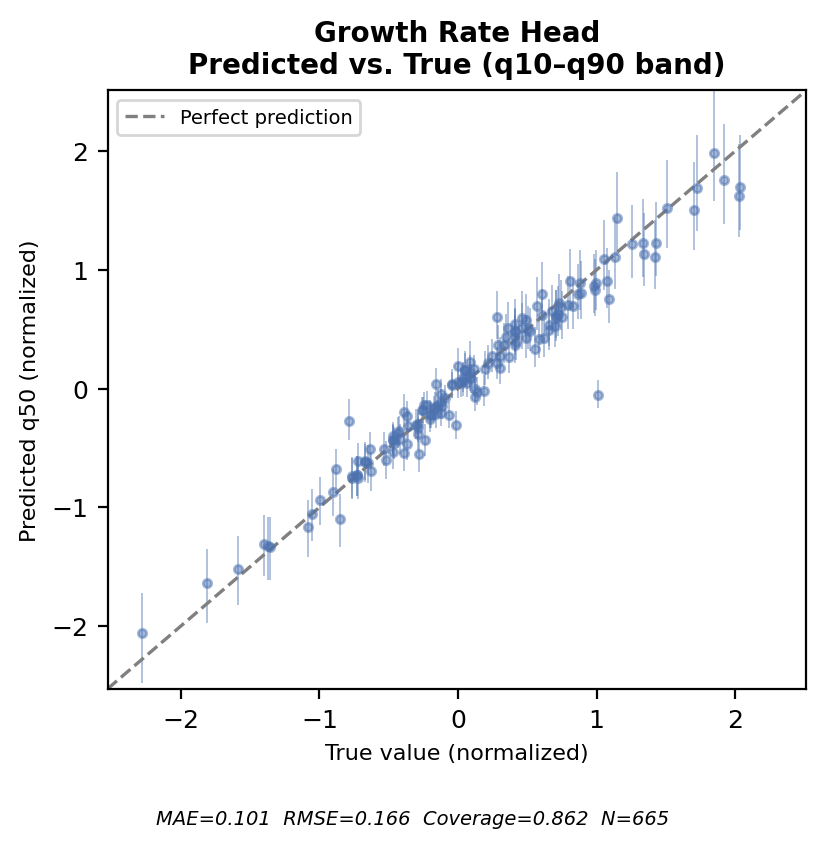}{11.\ Growth-Rate Head, Pred.\ vs.\ True} &
\gridimg{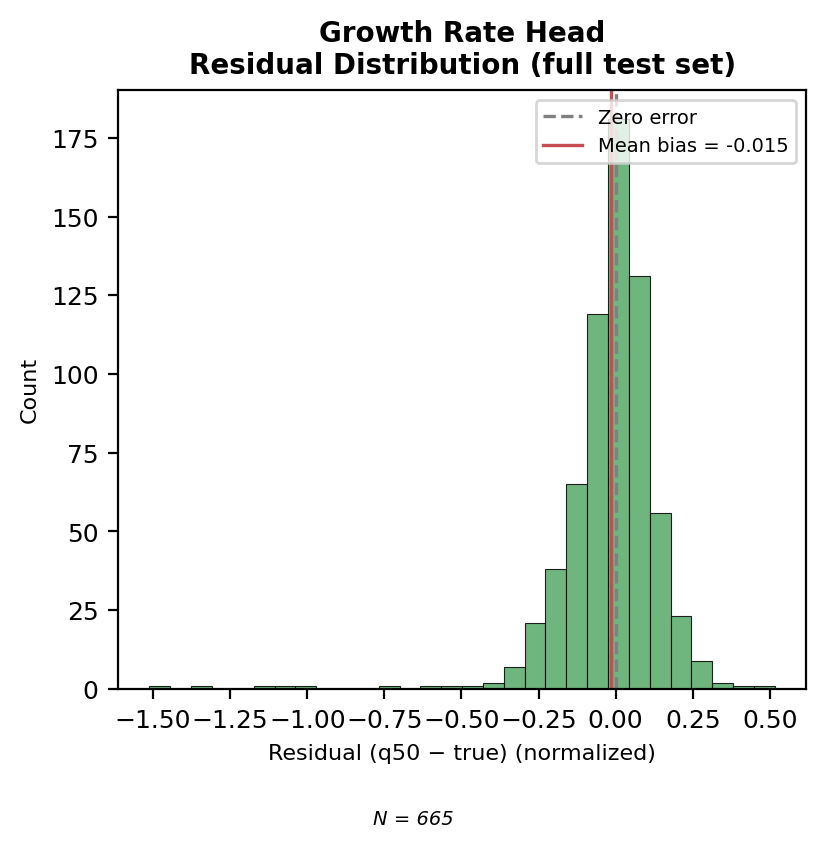}{11.\ Growth-Rate Head, Residuals}\\[6pt]
\gridimg{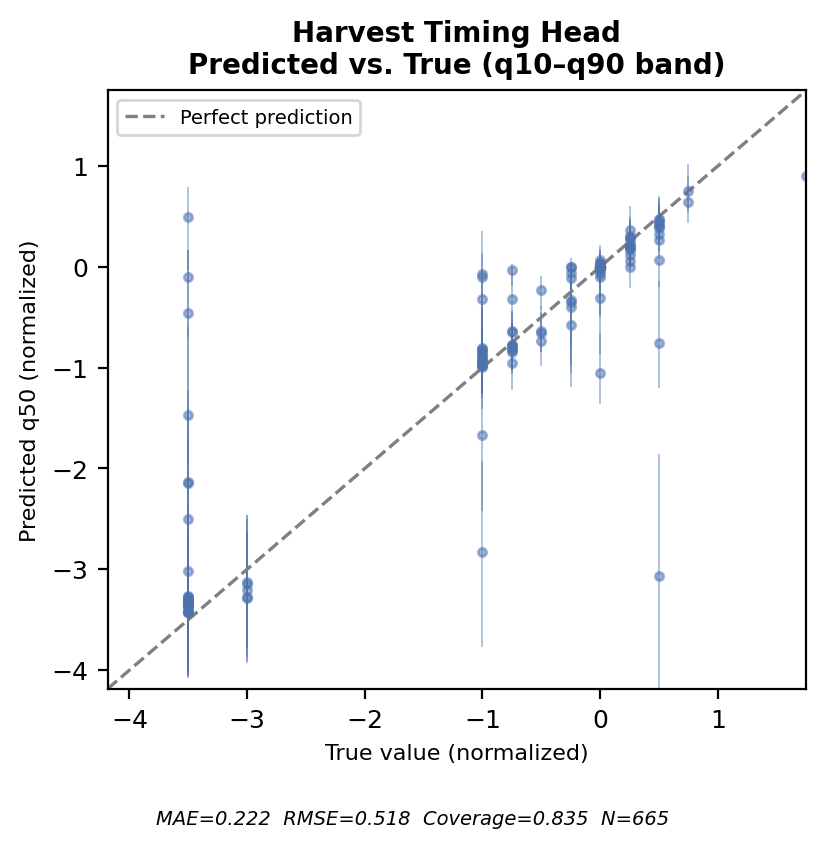}{12.\ Harvest-Timing Head, Pred.\ vs.\ True} &
\gridimg{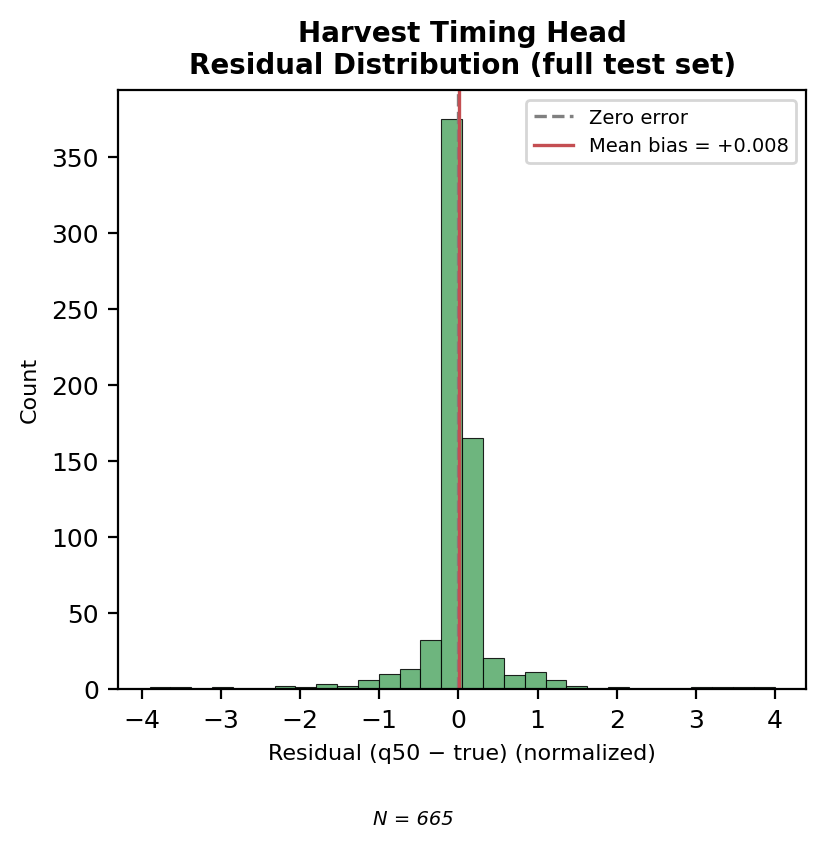}{12.\ Harvest-Timing Head, Residuals} &
\\[8pt]
\resultgrouphead{(e) Multi-Trait Regression (Lettuce, CS5): Predicted vs.\ True}
\gridimg{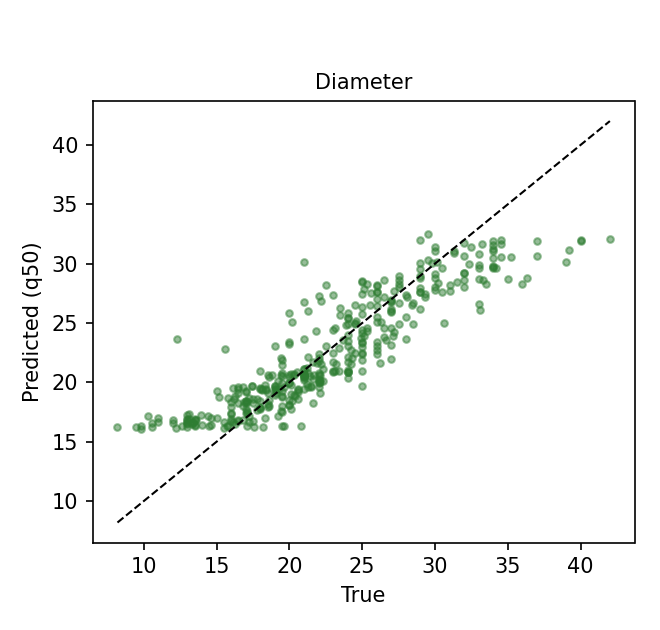}{13.\ Lettuce, Diameter, Pred.\ vs.\ True} &
\gridimg{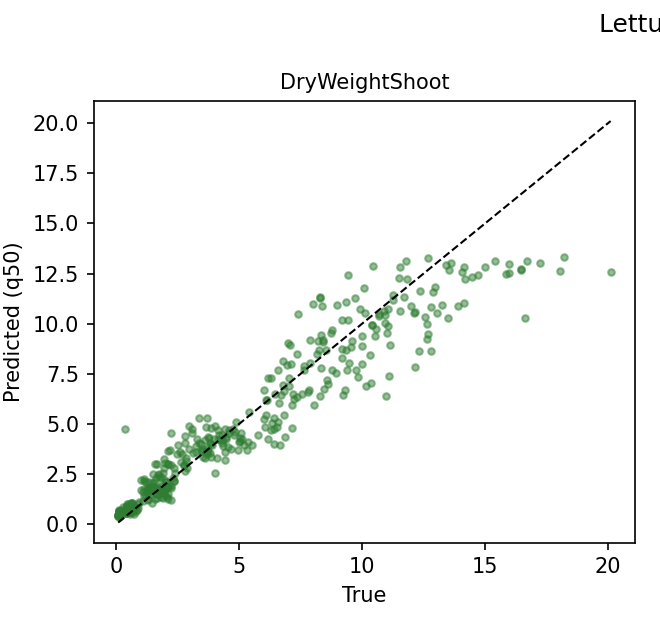}{13.\ Lettuce, Dry Weight (Shoot), Pred.\ vs.\ True} &
\gridimg{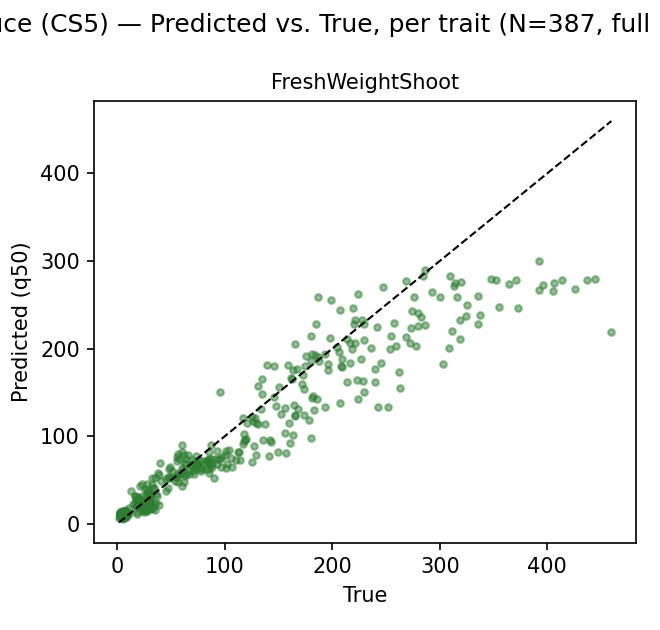}{13.\ Lettuce, Fresh Weight (Shoot), Pred.\ vs.\ True}\\[6pt]
\gridimg{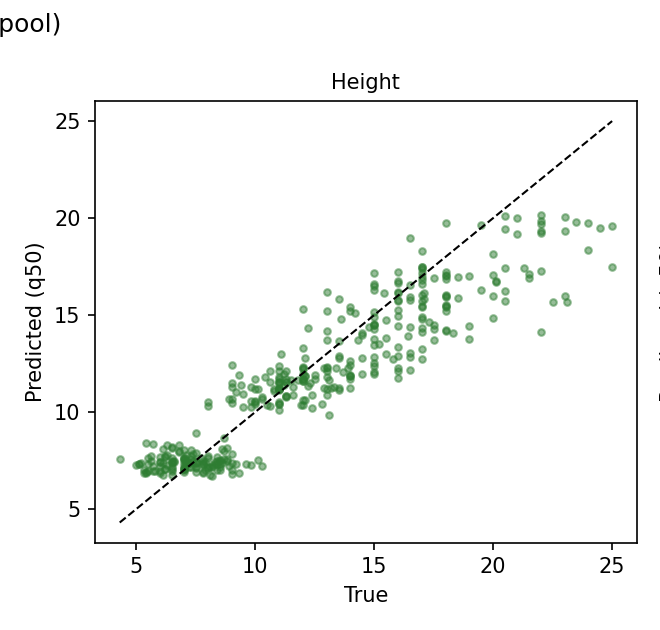}{13.\ Lettuce, Height, Pred.\ vs.\ True} &
\gridimg{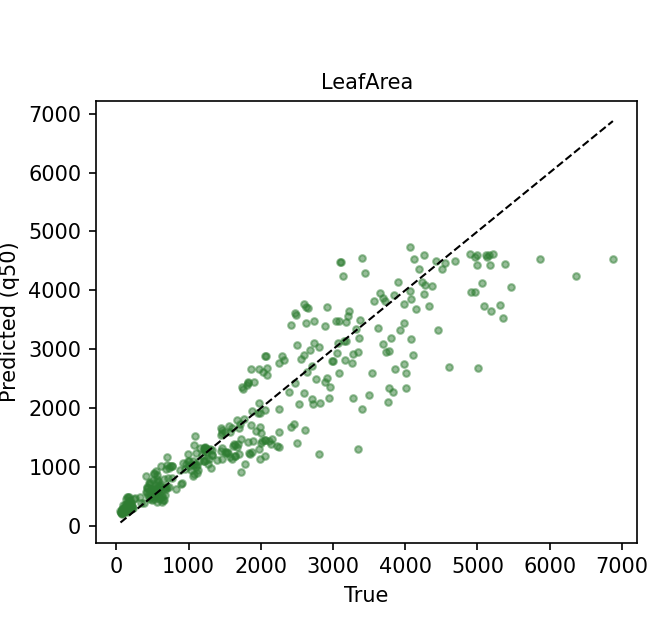}{13.\ Lettuce, Leaf Area, Pred.\ vs.\ True} &
\gridimg{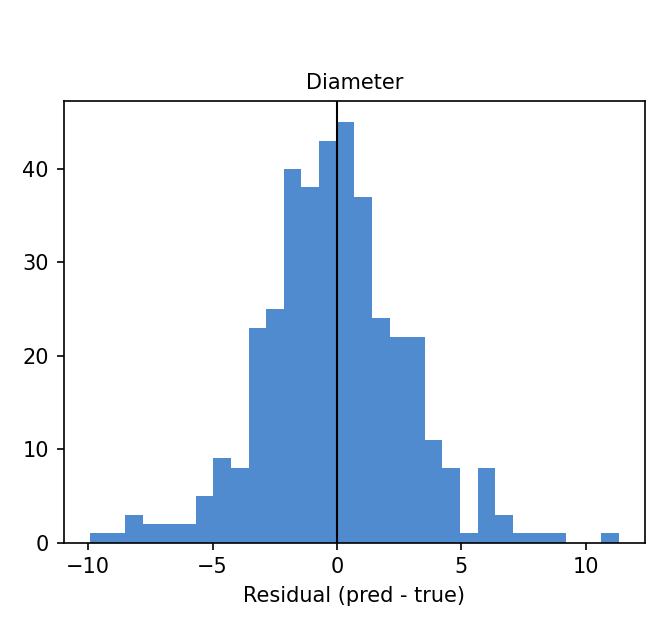}{13.\ Lettuce, Diameter, Residuals}\\
\end{tabular}
\caption*{Figure~\ref{fig:allresults-clf} continued: (d) Segmentation (CS2) and Regression models 11--12; (e) Lettuce (CS5) predicted vs.\ true, per trait.}
\end{figure}
\clearpage

\noindent
\begin{minipage}{\linewidth}
\centering
\setlength{\tabcolsep}{5pt}
\renewcommand{\arraystretch}{1.0}
\begin{tabular}{@{}ccc@{}}
\resultgrouphead{(e) Multi-Trait Regression (Lettuce, CS5): Residual Distributions (cont.)}
\gridimg{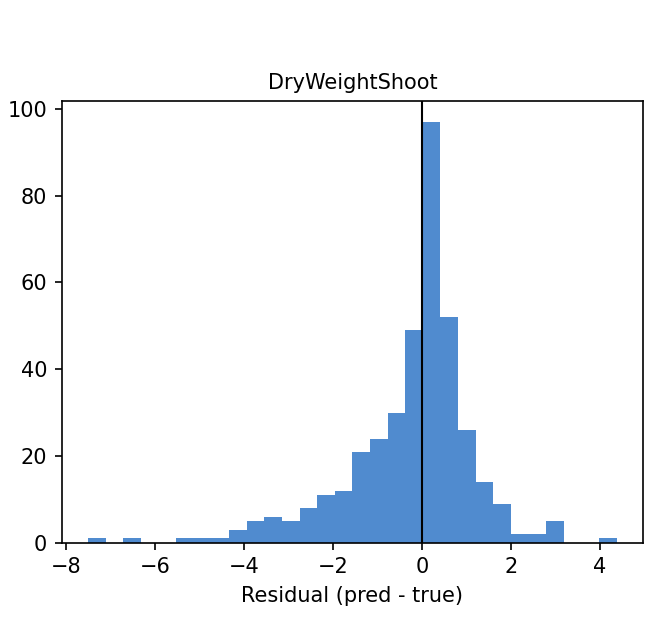}{13.\ Lettuce, Dry Weight (Shoot), Residuals} &
\gridimg{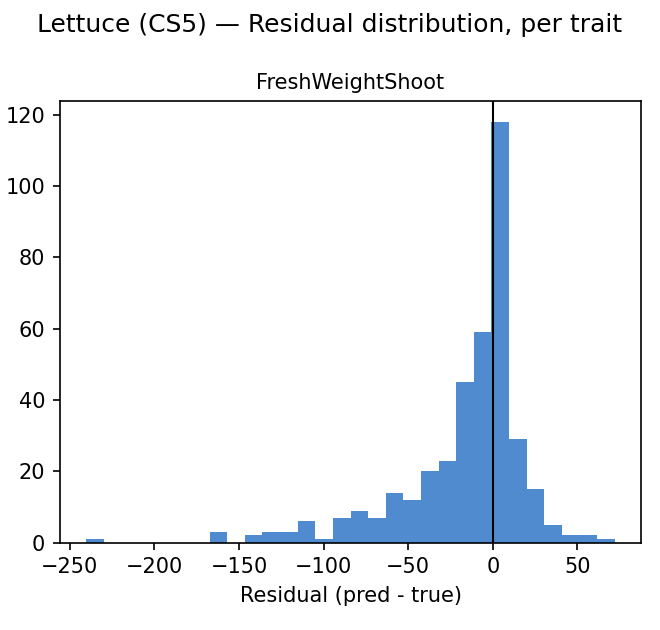}{13.\ Lettuce, Fresh Weight (Shoot), Residuals} &
\gridimg{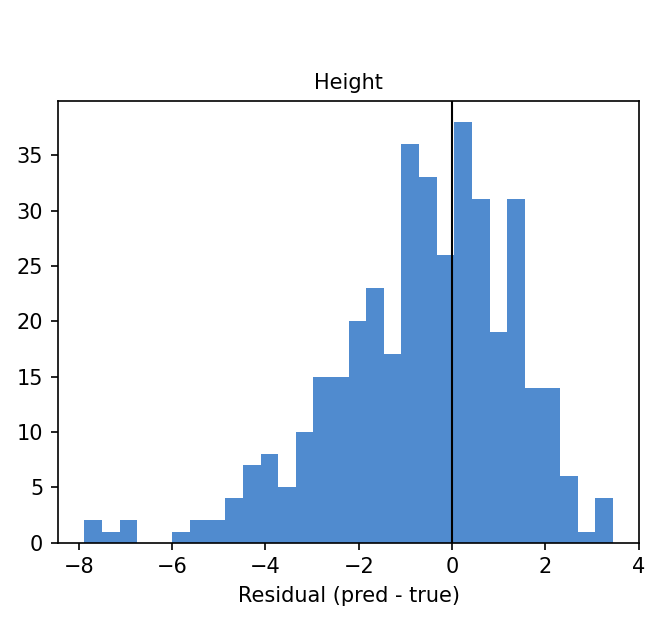}{13.\ Lettuce, Height, Residuals}\\[6pt]
\gridimg{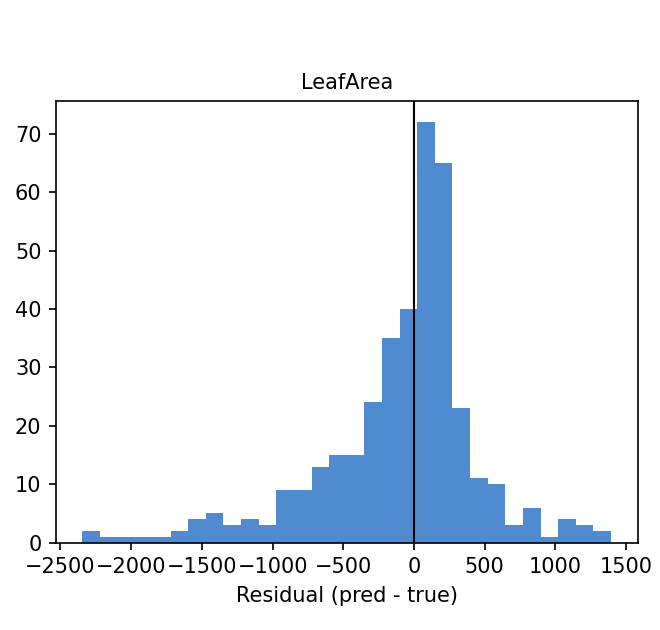}{13.\ Lettuce, Leaf Area, Residuals} &
&
\\
\end{tabular}
\captionof{figure}{Result figures for the ten model-zoo checkpoints and case
studies, by task family. \textbf{(a) Classification}: confusion matrix and
reliability diagram per classification model. \textbf{(b) Detection}:
detections and PR curve per detection model. \textbf{(c) Temporal}:
ensemble comparison, per-fold F1, and the CS6 training/self-evaluation
summary (accuracy/F1 and confidence-distribution panels shown
separately). \textbf{(d) Segmentation \& Regression}: predicted leaf count against
per-image CVPPP2015 ground truth and the corresponding per-image
residuals for the CS2 zero-shot pathway (Section~\ref{sec:711}, matching
the reported MAE\,=\,3.00, RMSE\,=\,4.31, Table~\ref{tab:mastersummary}),
alongside prediction-vs-true and residual plots for the growth-rate and
harvest-timing heads (Models 11--12). \textbf{(e) Multi-Trait Regression
(Lettuce, CS5)}: prediction-vs-true and residual-distribution plots for
all five lettuce traits, each shown as its own panel, concluding with the
residual distributions above. Headline numbers throughout are in
Table~\ref{tab:mastersummary}.}\label{fig:allresults-clf}
\end{minipage}
\FloatBarrier

\subsection{FAIR Principles and AI Governance}\label{sec:7}

\begin{figure}[htbp]
\centering
\resizebox{0.50\linewidth}{!}{%
\begin{tikzpicture}[
  font=\footnotesize,
  root/.style={draw, fill=teallink!22, rounded corners=3pt,
    text width=4.4cm, minimum height=0.60cm, align=center,
    font=\small\bfseries},
  file/.style={draw, fill=white, rounded corners=2pt,
    text width=4.0cm, minimum height=0.55cm,
    align=left, inner sep=4pt},
  F/.style={font=\footnotesize\bfseries,
    fill=blue!15,text=blue!70!black,rounded corners=1.5pt,inner sep=2pt},
  A/.style={font=\footnotesize\bfseries,
    fill=green!15,text=green!60!black,rounded corners=1.5pt,inner sep=2pt},
  I/.style={font=\footnotesize\bfseries,
    fill=orange!18,text=orange!70!black,rounded corners=1.5pt,inner sep=2pt},
  R/.style={font=\footnotesize\bfseries,
    fill=purple!15,text=purple!60!black,rounded corners=1.5pt,inner sep=2pt},
  arr/.style={-{Stealth[length=3.2pt]}, line width=0.5pt, gray!60},
  every node/.append style={inner sep=2.5pt}]
\node[root] (r) at (0,0) {FAIR provenance package (per session)};
\node[file, below left=0.55cm and 1.9cm of r] (mn)
  {Manifest \tikz[baseline]\node[F]{F};\tikz[baseline]\node[A]{A}; file inventory with checksums};
\node[file, below=0.30cm of mn] (rd)
  {Reproduction guide \tikz[baseline]\node[R]{R}; step-by-step instructions};
\node[file, below=0.30cm of rd] (pr)
  {Pipeline run record \tikz[baseline]\node[F]{F}; sequence of stages executed};
\node[file, below=0.30cm of pr] (mp)
  {Model provenance \tikz[baseline]\node[R]{R}; which trained model was used};
\node[file, below=0.30cm of mp] (im)
  {Original images \tikz[baseline]\node[F]{F};\tikz[baseline]\node[A]{A}; if uploaded by the user};
\node[file, below right=0.55cm and 0.05cm of r] (cal)
  {Calibration record \tikz[baseline]\node[I]{I};\tikz[baseline]\node[R]{R}; pixel-to-cm conversion};
\node[file, below=0.30cm of cal] (ac)
  {Assumption checks \tikz[baseline]\node[I]{I}; statistical test results};
\node[file, below=0.30cm of ac] (vr)
  {Verification log \tikz[baseline]\node[I]{I}; record of all checkpoint outcomes};
\node[file, below=0.30cm of vr] (ph)
  {Measurement table \tikz[baseline]\node[I]{I};\tikz[baseline]\node[R]{R}; per-image phenotype values};
\node[file, below=0.30cm of ph] (sr)
  {Statistical report \tikz[baseline]\node[I]{I};\tikz[baseline]\node[R]{R}; APA-formatted results};
\node[file, below=0.30cm of sr] (we)
  {Warnings/errors log \tikz[baseline]\node[R]{R}; flags raised during the run};
\node[file, below=0.30cm of we] (cs)
  {Calibration scores \tikz[baseline]\node[R]{R}; conformal scores (if available)};
\foreach \f in {mn,rd,pr,mp,im}{\draw[arr](r.south)--(\f.north);}
\foreach \f in {cal,ac,vr,ph,sr,we,cs}{\draw[arr](r.south)--(\f.north);}
\node[F] at (-3.4,-7.6){F};\node[font=\footnotesize]at(-2.9,-7.6){Findable};
\node[A] at (-1.7,-7.6){A};\node[font=\footnotesize]at(-1.2,-7.6){Accessible};
\node[I] at (0.3,-7.6){I};\node[font=\footnotesize]at(0.85,-7.6){Interoperable};
\node[R] at (2.5,-7.6) {R};\node[font=\footnotesize]at(3.0,-7.6){Reusable};
\end{tikzpicture}%
}
\caption{FAIR provenance package: 12 always-present files, checksum-indexed
in the package manifest. Badges indicate which FAIR principle each file
satisfies.}\label{fig:fair}
\end{figure}
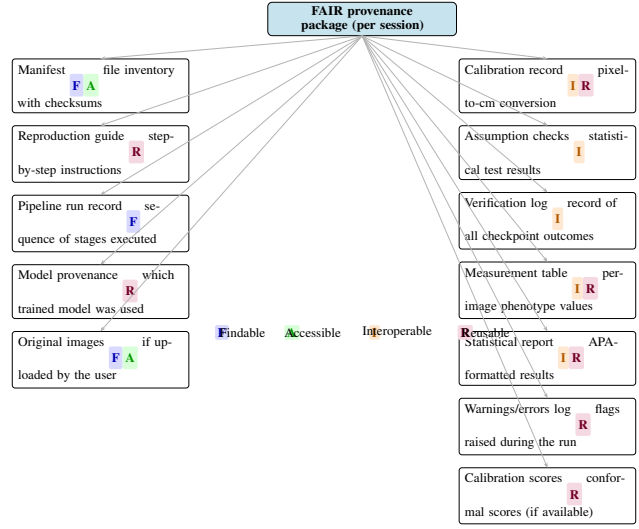

The FAIR data principles, Findable, Accessible, Interoperable,
Reusable~\citep{wilkinson2016}, are now used by major funding
agencies and journals as a governance framework for research data, and
AI-generated outputs are increasingly held to the same standard.
PhenoIntel's FAIR provenance package applies these at the pipeline
level: every session leaves behind a record that another researcher
can locate, open, understand, and use to reproduce the result without
access to PhenoIntel's own servers or code. \textbf{Findability}
comes from a unique session ID; \textbf{Accessibility} from a
dedicated provenance-download endpoint; \textbf{Interoperability}
from open, schema-documented formats throughout (missing values as
empty strings, not a literal zero as in prior systems; APA-compliant
statistics naming; ISO~8601 timestamps); \textbf{Reusability} from a
README with step-by-step reproduction instructions, including the
blueprint ID for any ADAPT-generated architecture. Every session
produces exactly 12 always-present files, checksum-indexed in a
package manifest, shown in Figure~\ref{fig:fair}.

This also operationalises the kind of containment emerging AI
governance frameworks (the EU AI Act, the NIST AI RMF) expect of
consequential AI applications. PhenoIntel's statistical-code sandbox
gives a full audit trail, strong containment (no network access,
read-only filesystem, hard memory/time limits), and execution-level
reproducibility. This directly resolves two confirmed cases of
ungoverned execution in the audited system: model-generated Python run
on the host with isolation disabled, and an unrecoverable runtime
failure mode in the data-query pathway. The same
verify-before-trust principle extends to the Pipeline Adaptation
Agent: a generated architecture never executes until compiled and
formally verified.

\FloatBarrier
\subsection{Internal Validation and Data Integrity}\label{sec:8}

PhenoIntel is covered by a 1,200-test automated suite (1,128 unit
tests across 61 files, 72 integration tests across 3 files), confirmed
1200/1200 passing on a full local run (240.5~s wall-clock; 4
non-blocking warnings, zero failures). This suite tests internal
consistency, that the typed state contract is enforced, that
checkpoints fire on the conditions they claim, that the union-mask
correction produces a larger area than the exclusive-or mask, that the
sandbox allowlist blocks disallowed imports, and is not a
substitute for the external accuracy evaluation above.

\textbf{Data leakage.}\label{sec:82} The most significant data-quality incident
encountered during development was accidental leakage between
training and validation splits for the rice nitrogen classifier:
initial validation Macro~F1 scores above 0.97 were traced to duplicate
images, captured at slightly different focal lengths but perceptually
identical, distributed across both splits by naive random splitting.
The fix, exact-hash deduplication followed by perceptual
near-duplicate detection, with explicit leakage verification, was
implemented in the dataset partitioning stage; after the fix, rice
nitrogen validation F1 stabilised at 0.9649 on the true held-out
split. A secondary pattern appeared in satellite crop-type
experiments, where temporal windows overlapping the test season
inflated F1; the fix was a strict temporal split (2016--2018
train/cross-validation, 2019 held-out test).

\textbf{Evaluation design standards.} Four choices shape how this
paper's own results should be read. Every accuracy figure is measured
against benchmarks this paper's authors did not design (CVPPP~2017,
GWHD). Every inference pathway is evaluated, not only the strongest
one; the weak zero-shot segmentation result is reported in full rather
than omitted. Accuracy is checked against ground truth rather than
only whether the pipeline finished or the right tool was invoked. And
every failure encountered during evaluation, including CS5's recovery
path, is reported rather than set aside.

\section{Discussion}\label{sec:9}

The results reported above extend beyond a head-to-head comparison
with prior monolithic systems~\citep{xu2024}: they show what a
lifecycle-aligned multi-agent architecture, with a verified mechanism
for extending its own capability, makes possible for plant
phenotyping specifically.

Each architectural response targets one agentic-AI failure mode and
is checked against its own metric rather than a single aggregate
score: failure propagation by the typed shared state and
no-zero-from-failure validator; non-deterministic tool selection by
the structured registry and four-tier fallback; unstructured
inter-tool communication by routing all messages through shared
session state; unbounded recovery loops by the one-replan limit and
bounded adaptation; absent uncertainty quantification by per-model UQ
dispatch; statistical invalidity by assumption checks and sandboxed
execution; inaccessibility by the browser deployment requiring no
local GPU; the FAIR deficit by the 12-file provenance package; and
the fixed task/model surface by the task ontology and pipeline
adaptation agents, each evaluated directly in
Sections~\ref{sec:5}--\ref{sec:8} rather than asserted here.

\subsection{Significance and Broader Impact}

\textbf{Scientific rigour.} Published phenotyping studies using
AI-generated measurements face an under-acknowledged reproducibility
problem: a tool with silent failure modes, no uncertainty
quantification, and unchecked statistical assumptions can
systematically bias published data in ways reviewers cannot detect.
The distribution-shift finding for foliar disease
classification~\citep{mohanty2016} shows that
a high same-distribution accuracy figure can misrepresent
field-readiness unless reported alongside its evaluation protocol,
precisely the distinction PhenoIntel's reliability-tier system
(Section~\ref{sec:64}) preserves rather than collapses into a single
headline number. A phenotype table with missing-value semantics,
Shapiro--Wilk results before every ANOVA, and RAPS conformal
prediction sets~\citep{vovk2005,angelopoulos2021} are reviewable
artefacts that make the process auditable without access to the AI
system itself.

\textbf{Agricultural applicability.} The five classification models,
covering rice, wheat, maize, banana, and coffee, expand accessible
crop-AI coverage across primary food and cash crops for hundreds of
millions of smallholder farmers where professional agronomy services
are sparse, reaching Macro~F1 of 0.78--0.996 across these five crops
despite the low-data regime typical of smallholder-relevant species
(coffee, $N=800$; banana, $N=1{,}500$; Section~\ref{sec:62},
Table~\ref{tab:mastersummary}), consistent with reports that
rice-disease classifiers reach comparable accuracy with modest,
field-collected training sets~\citep{sethy2020rice}. A
browser-accessible tool needing no local hardware lowers
the barrier from ``requires a university computing lab'' to ``requires
a smartphone with internet access.'' The temporal-modelling pipeline extends
this to field-scale monitoring using freely available satellite data
alone, in the spirit of prior uncertainty-aware temporal modelling for
remote sensing and forecasting~\citep{wang2020deeppipe,kar2024xwavenet}.

\textbf{Architectural transferability.} The lifecycle-aligned
multi-agent pattern is domain-agnostic: any workflow where a single
LLM orchestrates multiple domain-specific tools risks the same failure
modes. ChemCrow~\citep{bran2023} and GeneGPT~\citep{jin2023} show
subsets of these; the two published multi-agent phenotyping systems
closest to PhenoIntel, Chat Demeter~\citep{zhang2026chatdemeter} and
PestMA~\citep{shi2025pestma}, each add a checking role at only one
pipeline stage (Section~\ref{sec:relwork4}) rather than at every stage
boundary, the specific gap PhenoIntel's uniform V0--V7 checkpoints
close. Four parts of the design are not plant-specific and could be
reused as-is elsewhere: the shared typed record, the inter-stage
verification checkpoints, matching uncertainty method to model, and
the propose-check-verify-promote pattern for new capability, mirroring
the staged discipline MLflow~\citep{zaharia2018} and Kubeflow
Pipelines~\citep{bisong2019} established for training pipelines. Only
the model library and plant-specific plausibility rules would need
swapping out.

\textbf{Open-science reproducibility.} The FAIR provenance
package~\citep{wilkinson2016} operationalises open-science
requirements at a granularity prior phenotyping tools have not
reached: every session's manifest, calibration record, and
verification log are Findable, Accessible, Interoperable, and
Reusable by construction, not asserted after the fact. Measured
directly rather than claimed, PhenoIntel returns 12/12 required
provenance files across all sixteen sessions and full session
resumability (Table~\ref{tab:mastersummary}), giving reviewers a
machine-verifiable audit trail sufficient to reproduce the pipeline
from scratch, including, for ADAPT sessions, the generated
architecture via its blueprint ID. This closes a gap the
reproducibility literature has flagged for uncertainty-aware crop-model
and phenotyping pipelines more broadly~\citep{confalonieri2016quantifying,giacomini2022framework}:
a result's statistical validity is only checkable if the underlying
computation itself is inspectable, the same requirement APA-style
statistical reporting~\citep{APA2020} imposes at the level of a single
test rather than a whole pipeline.

\subsection{Limitations}\label{sec:92}

The evaluation improves on prior system evaluations by measuring
ground-truth phenotype accuracy, uncertainty calibration, and
statistical validity against external benchmarks rather than
self-designed tasks. Several limitations remain, stated plainly rather
than downplayed.

\textbf{Modelling.} No fine-tuned checkpoint yet exists for structural
phenotyping, so the zero-shot segmentation fallback is the weakest
part of the system and the top priority for a fine-tuned replacement
(Section~\ref{sec:93}); a broader CVPPP~2017 evaluation sweep was
planned but not run, for the hardware reason given below. The RAPS
conformal-prediction pathway~\citep{angelopoulos2021} is calibrated
and ready for live use, but its own distribution-free coverage
guarantee~\citep{vovk2005} has not yet been benchmarked offline. The
crop-type head's recorded Test Macro~F1 of 0.7050 could not be
independently regenerated, and a later full-pool, uncalibrated
re-evaluation of the same 10-checkpoint ensemble returned
Macro~F1\,=\,0.2649; part of this gap is expected, since the
re-evaluation skips the report's per-class calibration/threshold step,
but whether that fully accounts for it is unconfirmed and reported as
an open discrepancy rather than resolved away. (The CS6 satellite
model is a separate checkpoint with its own, larger discrepancy --
0.973 self-reported vs.\ 0.3728 reproduced -- discussed in
Table~\ref{tab:allmodels}, footnote~e.) This is exactly the kind of split-integrity
failure the strict temporal-split discipline used elsewhere in the
temporal (satellite) pipeline (Section~\ref{sec:curmethod}) is meant to
prevent~\citep{kar2020spatemhtp,kar2020automated}. A few
uncertainty-propagation constants were also tuned empirically rather
than derived analytically, similar in spirit to empirically calibrated
uncertainty models in canopy-flow and crop-model
literature~\citep{giacomini2022framework,confalonieri2016quantifying}.

\textbf{Evaluation.} Three case studies (CS1 $n{=}10$, CS2 $n{=}5$,
CS4's counting split $n{=}7$) are single runs without bootstrap
confidence intervals, and the 1,200-test suite checks each agent and
checkpoint individually rather than providing a system-level ablation
on a live session. Only one of the Pipeline Adaptation Agent's two
triggers, an inference failure, has been exercised end-to-end (CS5);
the other, a quality-gate rejection on a real dataset, remains
untested. The comparison with PhenoAssistant~\citep{xu2024} covers
only the pathways where it has a complete implementation, and
PhenoIntel's agents use Llama~3 (70B) via Groq while the prior system
uses GPT-4o, a model-family difference worth keeping in mind. The
calibration-propagation step uses a linear approximation valid for
small calibration errors, and may understate uncertainty when
calibration confidence is low. The ten model-zoo checkpoints were also
trained across independent Kaggle sessions rather than one unified
pipeline, so the evaluation protocol is not uniform: some figures are
held-out, some full-pool, and one (cotton, CS4) mixes an internal split
with a later full-pool re-evaluation on a larger pool
(Table~\ref{tab:allmodels}, footnote~g). The reliability-tier system
(Section~\ref{sec:64}) exists to surface this heterogeneity rather than
hide it, but it is a weaker evidentiary foundation than one stratified
split applied identically to every checkpoint; a unified re-training
pass under one held-out protocol remains future work.

\textbf{Hardware and scale.} Every result here was produced on a
single CPU-only, unfunded machine (Section~\ref{sec:accessibility}),
which explains several other limitations rather than sitting apart
from them. Training cost scales with dataset size on this hardware,
which is why the Build workflow caps fine-tuning at 70 minutes and why
the six case studies deliberately use small datasets
(Table~\ref{tab:casestudies}) rather than the larger corpora behind
the ten offline checkpoints: each case study must complete a live
Build session within a practical time budget. A migration to
GPU-backed cloud infrastructure (Section~\ref{sec:93}) would remove
this ceiling; it is specified as infrastructure-as-code but not yet
deployed, for lack of external funding for cloud GPU time. The dataset
sizes and training times reported here should be read as bounded by
what one unfunded machine could evaluate in the time available, not as
a limit on the architecture itself.

None of these limitations affect the paper's structural claims: that a
failure is caught before it reaches a reported measurement, and that
every output is auditable. They bound how far the current numeric
results should be generalised, and are reported here for that reason.

\subsection{Future Directions}\label{sec:93}

The top priority is a fine-tuned segmentation checkpoint (e.g.\ on
CVPPP~2017) for high-overlap crops, followed by the hardware-adequate
CVPPP~2017 A1--A3 evaluation the current machine could not run. The
second priority is exercising the Pipeline Adaptation Agent's
remaining trigger, a quality-gate rejection on a real dataset, rather
than an inference failure as in CS5. Moving training to GPU-backed
cloud infrastructure would remove the hardware ceiling behind both and
allow larger case-study datasets; the migration is already specified
as infrastructure-as-code and is the concrete next step once funding
allows it.

Beyond these, promising directions include active-learning routing of
high-uncertainty images to human review, model-zoo expansion to more
crops and imaging modalities, conformalised quantile regression for
regression tasks~\citep{romano2019}, a larger vision-transformer
backbone for the smallest datasets (banana, coffee), federated
learning across institutions without pooling raw images, longitudinal
multi-season field validation of the growth-rate and harvest-timing
targets, and extending the plausibility-checking approach here toward
process-level uncertainty modelling explored in recent plant-process
and graph-based
work~\citep{srivastava2026seagan,cherif2026uncertainty,faisal2025uncertainty}.

PhenoIntel's design sits between two extremes: a fully free-form
assistant, where the AI decides what to do at every turn, and a fully
hard-coded pipeline with no AI involvement at all. The eight-stage ML
lifecycle supplies the fixed structure, with the AI's role checked at
each point set out in Section~\ref{sec:4overview}, which is what keeps
the system reproducible across runs.

The limitations and future directions above are summarised briefly in
Section~\ref{sec:10}, alongside the paper's concluding remarks.

\section{Conclusion}\label{sec:10}

We presented PhenoIntel, a web application that divides plant
phenotype analysis into separately checked stages rather than
assigning the task to a single AI manager, directly addressing four
recurring problems identified by a source-code audit of a prior
monolithic orchestrator: silent zero-from-failure,
unchecked statistics, no confidence reporting, and inaccessible setup.
Nine agents read from and write to one shared, typed record, checked
at every stage boundary (V0--V7) with bounded recovery; uncertainty is
matched to each model's behaviour; and a self-extending layer adds
new task types and architectures under the same checks as any other
output. It runs in a browser with no GPU, is validated by a 1,200-test
suite, and produces a FAIR-compliant provenance record per session.
The evaluation remains bounded by a single unfunded, CPU-only machine
and by several results resting on a single run or an unconfirmed
split rather than a reproduced one; priority
future work is a fine-tuned segmentation checkpoint, GPU-backed
training infrastructure, and broader model-zoo and multi-season
validation. This work does not claim
PhenoIntel's numbers are the highest achievable; it demonstrates that
its failure modes are structurally bounded and its outputs
independently auditable by design.

\begingroup
\makeatletter
\renewcommand{\section}{%
\@startsection{section}{1}{\z@}%
{-1.1ex \@plus -0.3ex \@minus -0.1ex}%
{ 0.5ex \@plus 0.15ex}%
{\large\bf\raggedright}%
}
\makeatother

\section*{Data Availability Statement}

All datasets used to fine-tune or evaluate a model in this paper are
either publicly available third-party benchmark releases or
author-processed datasets deposited on Kaggle, cited by name and
access point in Table~\ref{tab:datasetsources}. The PhenoIntel source
code, model checkpoints, and the 1,200-test automated suite referenced
throughout this paper are maintained in a version-controlled
repository, available at
\url{https://github.com/Naren1704/PhenoIntel-Internship}. The CVPPP2015 Leaf Counting Challenge per-plant
annotations used for Case Study~2 are distributed by the challenge
organisers and were matched by plant ID to a 5-image evaluation
subset drawn from the A1 (Arabidopsis) release of that 2015 challenge;
this A1 is the 2015 Leaf Counting Challenge's species subset and is
unrelated to the like-named A1--A3 split of the 2017 challenge below.
The CVPPP~2017 Leaf Segmentation Challenge A1--A3
releases~\citep{minervini2016} were not evaluated here (hardware
reasons, Sections~\ref{sec:71}, \ref{sec:93}), so no A1--A3 ground
truth is referenced by this paper's results.

\section*{Declaration of Generative AI and AI-assisted Technologies}

During the preparation of this work, the author(s) used Claude
(Anthropic) to assist with drafting, restructuring, and verifying
consistency of numeric and structural claims against the underlying
source-code repository and test-suite output. After using this
tool/service, the author(s) reviewed and edited the content as needed
and take full responsibility for the content of the published article.

\section*{CRediT Authorship Contribution Statement}

\textbf{Narendren~S~V}: Conceptualization, Methodology, Software,
Data curation, Visualization, Validation, Writing -- original draft,
Writing -- review \& editing. \textbf{Soumyashree~Kar}:
Conceptualization, Methodology, Supervision, Investigation,
Validation, Visualization, Formal analysis, Writing -- review \&
editing, Project administration.

\section*{Declaration of Competing Interest}

The authors declare that they have no known competing financial
interests or personal relationships that could have appeared to
influence the work reported in this paper.

\section*{Funding}

This research did not receive any specific grant from funding
agencies in the public, commercial, or not-for-profit sectors.
\endgroup

\end{document}